\pdfoutput=1
\documentclass[11pt]{article}

\usepackage[preprint]{acl}

\usepackage{times}
\usepackage{latexsym}

\usepackage[T1]{fontenc}
\usepackage[utf8]{inputenc}

\usepackage{microtype}

\usepackage{inconsolata}

\usepackage{graphicx}
\usepackage{amsmath}
\usepackage{bbm}
\usepackage{amsfonts}
\usepackage{subcaption}
\usepackage{multirow}
\usepackage{booktabs}
\usepackage{float}
\usepackage{placeins}
\usepackage{hyperref}

\title{Revealing the Deceptiveness of Knowledge Editing: A Mechanistic Analysis of Superficial Editing}

\author{
 \textbf{Jiakuan Xie\textsuperscript{1,2}},
 \textbf{Pengfei Cao\textsuperscript{1,2}},
 \textbf{Yubo Chen\textsuperscript{1,2}},
 \textbf{Kang Liu\textsuperscript{1,2}},
 \textbf{Jun Zhao\textsuperscript{1,2}}
\\
\\
 \textsuperscript{1}School of Artificial Intelligence, University of Chinese Academy of Sciences
 \\
 \textsuperscript{2}The Laboratory of Cognition and Decision Intelligence for Complex Systems,
\\
Institute of Automation, Chinese Academy of Sciences
\\
 \small{
   \texttt{xiejiakuan2023@ia.ac.cn}, \texttt{\{pengfei.cao, yubo.chen, kliu, jzhao\}@nlpr.ia.ac.cn}
 }
}

\begin{document}
\maketitle
\begin{abstract}
Knowledge editing, which aims to update the knowledge encoded in language models, can be deceptive. Despite the fact that many existing knowledge editing algorithms achieve near-perfect performance on conventional metrics, the models edited by them are still prone to generating original knowledge. This paper introduces the concept of ``\textbf{superficial editing}'' to describe this phenomenon. Our comprehensive evaluation reveals that this issue presents a significant challenge to existing algorithms. Through systematic investigation, we identify and validate two key factors contributing to this issue: (1) the residual stream at the last subject position in earlier layers and (2) specific attention modules in later layers. Notably, certain attention heads in later layers, along with specific left singular vectors in their output matrices, encapsulate the original knowledge and exhibit a causal relationship with superficial editing. Furthermore, we extend our analysis to the task of superficial unlearning, where we observe consistent patterns in the behavior of specific attention heads and their corresponding left singular vectors, thereby demonstrating the robustness and broader applicability of our methodology and conclusions. Our code is available \href{https://github.com/jiakuan929/superficial-editing}{here}.
% This document is a supplement to the general instructions for *ACL authors. It contains instructions for using the \LaTeX{} style files for ACL conferences.
% The document itself conforms to its own specifications, and is therefore an example of what your manuscript should look like.
% These instructions should be used both for papers submitted for review and for final versions of accepted papers.
\end{abstract}

\section{Introduction}
\begin{figure}[t]
  \includegraphics[width=\columnwidth]{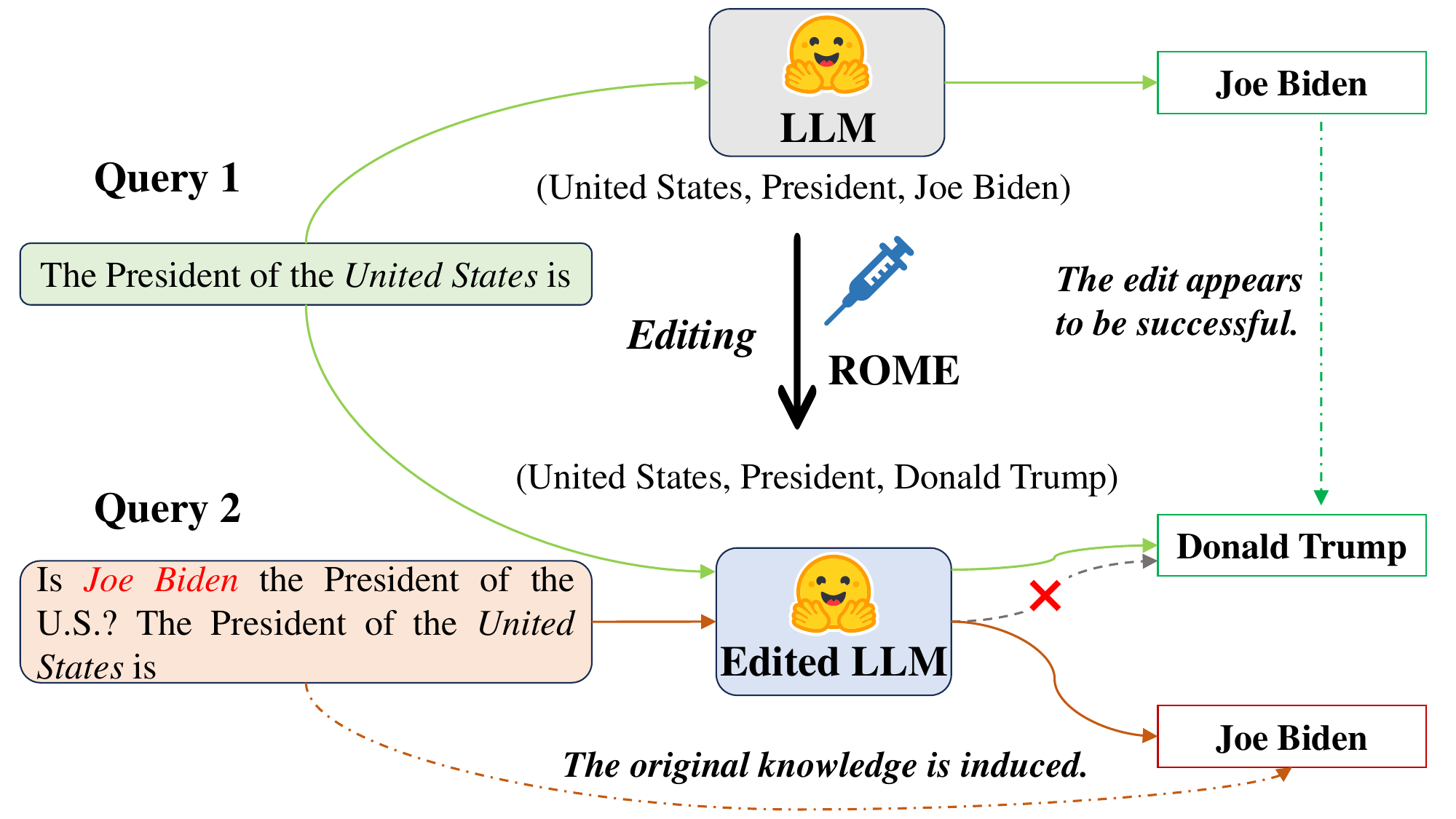}
  \caption{An example of superficial editing with the LLaMA3-8B-Instruct model. Following the editing process, the model accurately responds to Query 1. However, when presented with Query 2 as input, the edited model reverts to generating the original answer.}
  \label{fig:intro_example}
\end{figure}

The inherent static nature of knowledge embedded within a pretrained large language model (LLM) poses a fundamental limitation as the real world evolves. 
To address this issue, the concept of knowledge editing has been proposed to modify specific knowledge in LLMs while ensuring that unrelated knowledge remains unaffected (\citealp{zhu2020ft}). To date, numerous studies have been conducted on knowledge editing, encompassing diverse methodologies (\citealp{zhu2020ft}; \citealp{decao-ke}; \citealp{mend}; \citealp{serac}; \citealp{rome}; \citealp{memit}; \citealp{ike_icl}), paradigms (\citealp{grace}; \citealp{alphaedit}; \citealp{neuronlevelske}; \citealp{oedit}; \citealp{crosslingual0}; \citealp{crosslingualwang}; \citealp{mulfe}; \citealp{akew}), evaluation strategies (\citealp{mquake}; \citealp{rippleeffct}; \citealp{longformeval}; \citealp{yang-etal-2024-butterfly}; \citealp{ma-robustness}), and applications (\citealp{detoxify_nyz}; \citealp{editingdpo}; \citealp{editfairness}). 
Although significant progress has been made in these endeavors, a critical challenge persists: \textit{models that appear to have been successfully edited may unexpectedly revert to their original knowledge when exposed to specific contextual inputs}. 
% (\citealp{ma-robustness}). 
As shown in Figure \ref{fig:intro_example}, the edited model demonstrates the capability to appropriately respond to the query ``The President of the United States is''. However, when the context ``Is Joe Biden the President of the U.S.?'' is incorporated into the query, the updated model reverts to generating responses based on its original knowledge. 
The phenomenon reveals the potential deceptiveness of knowledge editing: edited models may revert to their original knowledge, undermining the goal of enabling continuous knowledge updates in LLMs. This limitation severely hinders the practical utility and reliability of knowledge editing.
% The phenomenon highlights a significant challenge: if the edited model retains the propensity to revert to their original knowledge, the objective of facilitating continuous knowledge updates in LLMs remains unachievable. This limitation significantly impedes the practical application and reliability of knowledge editing.

In this paper, we define a knowledge editing process as ``\textbf{superficial editing}'' when the resulting model appears to successfully integrate new knowledge, yet reverts to its original knowledge when exposed to carefully crafted prompts. 
To quantitatively evaluate this issue, we introduce ``attack probe'', which is a specifically designed prompt consisting of an attack prefix and a baseline prompt (as exemplified by Query 2 in Figure \ref{fig:intro_example}). We develop three attack types based on two widely used datasets and assess several editing algorithms across three models.
Empirical results demonstrate that while the majority of editing algorithms exhibit strong performance on conventional evaluation metrics, models edited through these approaches remain vulnerable to attack probes.
For instance, both PMET (\citealp{pmet}) and AlphaEdit (\citealp{alphaedit}) demonstrate near-optimal performance in terms of editing efficacy; however, they exhibit superficial editing in over 70\% of the cases. 
This finding suggests that current parameter-editing algorithms are fundamentally inadequate in addressing the challenge of superficial editing.

To elucidate the underlying mechanisms of this phenomenon, we focus on the core components of the Transformer architecture (\citealp{attnisalluneed}). 
%%%%%%%%%%
We initially conduct intervention experiments on the residual stream at two token positions. First, our intervention at the last subject position in earlier layers reveals shifts in prediction probabilities. Second, we intervene at the last position and find that this intervention exerts a significant effect in the later layers, where the probability of the original answer exceeds that of the new answer. This shift is a prerequisite for superficial editing, a phenomenon we term the ``Reversal of the Residual Stream'' (RRS). 
Additionally, our preliminary analysis of Multi-Layer Perceptron (MLP) and Multi-Head Attention reveals that specific attention modules in later layers play a significant role.
Based on the observations, we formulate two hypotheses:
%%%%%%%%%%
% Based on the preliminary experimental observations, we formulate two hypotheses:
(\textbf{H1}) \textbf{The enrichment of new knowledge at the last subject position in earlier layers is impeded, and the accumulation of the original knowledge at this position is relatively limited.} 
% (\textbf{H1}) The enrichment of new knowledge at the last subject position is hindered, with minimal accumulation of original knowledge occurring at this position in early residual streams. 
(\textbf{H2}) \textbf{The later attention modules actively incorporate information related to original knowledge into the last position, thereby facilitating the RRS phenomenon and consequently inducing the occurrence of superficial editing.}
% To validate \textbf{H1}, we implement an intervention in the edited model's forward propagation during its processing of the baseline prompt and observe the latent probability distributions of both the new response and the original response at the last subject position.
To validate \textbf{H1}, we project the representation of the last subject position in each layer into the vocabulary space. We observe greater suppression of the new answer when the attack probe is used as input, compared to the baseline prompt. However, despite this suppression effect, the ranking of the original answer consistently lags behind that of the new answer in the earlier layers, indicating minimal enrichment of the original knowledge.
% To validate \textbf{H2}, we conduct a comprehensive analysis of the attention mechanism, examining both attention heads and singular vectors derived from singular value decomposition (SVD). This investigation reveals a critical underlying factor contributing to superficial editing.
To validate \textbf{H2}, we first establish that the later attention modules exhibit a causal relationship with the RRS phenomenon, highlighting their critical role. 
Subsequently, we analyze the attention heads and confirm that a causal correlation exists between certain heads in later layers and superficial editing.
% Subsequently, we analyze the attention heads and identify that certain heads in the later layers introduce information related to the original knowledge into the last position. Moreover, a causal relationship exists between these heads and superficial editing. 
Furthermore, we investigate the internal mechanisms of attention heads through singular value decomposition (SVD) and demonstrate that specific left singular vectors are responsible for encoding the original knowledge and contributing to superficial editing. 
These findings provide robust evidence in support of \textbf{H2}.

To demonstrate the broader applicability of our interpretability analysis framework, we extend our investigation to a distinct task: \textbf{superficial unlearning}, wherein the unlearned model fails to truly forget the target information. Our experimental results reveal a strong correlation between this phenomenon and specific attention heads along with their corresponding singular vectors, substantiating the generalizability of both our analytical methodology and conclusions.

The primary contributions of this paper are as follows: \textbf{(1)} We formally define superficial editing and provide corresponding evaluation datasets and metrics, thereby completing the assessment of multiple algorithms. \textbf{(2)} We identify and validate two critical factors contributing to superficial editing: the residual stream in earlier layers and specific attention modules in later layers. Additionally, we explore the internal mechanisms of the attention module and reveal that specific attention heads and their corresponding left singular vectors are responsible for superficial editing. \textbf{(3)} We apply our analytical approach to superficial unlearning. The consistent finding across both phenomena validates the robustness and broader applicability of both our methodology and conclusions.
% \begin{itemize}
%     % \item We establish a rigorous theoretical framework for superficial editing, including formal definition, curated datasets, and comprehensive evaluation metrics. 
%     \item We formally define superficial editing and provide corresponding evaluation datasets and metrics, thereby completing the assessment of multiple algorithms.
%     % Furthermore, we conduct empirical evaluations of various editing algorithms, providing benchmark results for future research.
%     % \item                                                                                      Through systematic investigation, we identify and validate two critical hypotheses contributing to superficial editing.
%     \item We identify and validate two critical factors contributing to superficial editing: the residual stream in earlier layers and specific attention modules in later layers. Additionally, we explore the internal mechanisms of the attention module and reveal that specific attention heads and their corresponding left singular vectors are responsible for superficial editing.
%     % Our analysis reveal the mechanistic interplay between these components in generating  
%     \item We apply our analytical approach to superficial unlearning. The consistent finding across both phenomena validate the robustness and broader applicability of both our methodology and conclusions.
% \end{itemize}

\begin{table*}
  \centering
  \scalebox{0.73}{\begin{tabular}{l|ccccc|ccccc|ccccc}
    \toprule
    \multirow{2}{*}{\textbf{Methods}}           & \multicolumn{5}{c|}{\textbf{Wiki}} & \multicolumn{5}{c|}{\textbf{Rep}} & \multicolumn{5}{c}{\textbf{Que}} \\
     & Eff. & Gen. & Loc. & OM $\downarrow$ & OP $\downarrow$ & Eff. & Gen. & Loc. & OM $\downarrow$ & OP $\downarrow$ & Eff. & Gen. & Loc. & OM $\downarrow$ & OP $\downarrow$ \\
    \midrule
    FT & 100 & 80.51 & 52.37 & 49.45 & 51.65      & 100 & 70.54 & 44.46 & 30.68 & 35.98       & 100 & 87.90 & 33.87 & 29.07 & 31.40 \\
    MEND & 98.31 & 65.25 & 47.63 & 35.16 & 39.56    & 100 & 50.99 & 51.29 & 34.47 & 38.36    & 100 & 81.45 & 39.52 & 33.73 & 38.37 \\
    ROME & 100 & 94.92 & 85.08 & 54.95 & 58.24    & 100 & 97.52 & 84.75 & 61.74 & 64.02    & 100 & 99.19 & 82.74 & 38.37 & 38.37 \\
    MEMIT & 100 & 94.07 & 86.10 & 52.75 & 54.95    & 100 & 98.27 & 87.18 & 40.15 & 42.42    & 100 & 100 & 82.58 & 37.21 & 37.21 \\
    PMET & 94.92 & 85.59 & 90.00 & 70.33 & 72.43    & 99.50 & 93.32 & 91.88 & 66.67 & 71.97   & 96.67 & 89.17 & 88.17 & 39.29 & 41.67 \\
    r-ROME & 96.61 & 92.37 & 86.78 & 54.95 & 57.14    & 99.01 & 97.28 & 89.11 & 64.39 & 68.18    & 98.33 & 97.50 & 84.50 & 40.48 & 40.48 \\
    AlphaEdit & 100 & 83.90 & 88.98 & 72.53 & 73.62    & 100 & 92.33 & 92.23 & 68.18 & 71.97    & 100 & 88.33 & 87.67 & 34.52 & 35.71 \\
    \bottomrule
  \end{tabular}}
  \caption{Evaluation results of superficial editing conducted on LLaMA3-8B-Instruct using the CF-a dataset. \textbf{Wiki}, \textbf{Rep}, and \textbf{Que} represent the three attack types defined in Section \ref{sec:formulation}.
  Experimental results for other models and datasets are available in Appendix \ref{sec:appen_eval}.
  }
  \label{tab:llama3_eval_cf}
\end{table*}

\section{Problem Formulation}
\label{sec:formulation}
Knowledge editing, which aims to adjust the knowledge of a language model, can generally be expressed as follows:

{\footnotesize\begin{equation}
    \left(s, r, o\right) \xrightarrow{e} \left(s, r, o^*\right),
    \label{eq:editing}
\end{equation}}where $s$ is subject (e.g., United States), $r$ is relation (e.g., President), $o$ is the pre-editing object (e.g., Joe Biden), $o^*$ is the post-editing object (e.g., Donald Trump), and $e$ is a prompt used for editing (e.g., ``The President of the United States is''). We define the following attack prefixes for $o$:

{\footnotesize\begin{equation}
    a\in \mathcal{A}= \{ \text{Wiki}\left(o\right), \text{Rep}\left(o\right), \text{Que}\left(o\right) \},
\end{equation}}where $a$ is an attack prefix, $\text{Wiki}\left(o\right)$ denotes the Wikipedia summary of $o$, $\text{Rep}\left(o\right)$ denotes the repetition of $o$, and $\text{Que}\left(o\right)$ represents a question incorporating $s$, $r$, and $o$ simultaneously (e.g., Is Joe Biden the President of the U.S.?).
The set of all queries derivable from $s$ and $r$ is denoted as $\mathcal{I}=\{ x \mid s, r\Rightarrow x \}$. 
According to the editing operation defined in Equation \ref{eq:editing}, the edit is classified as \textbf{superficial editing} if the edited model $f^\prime$ satisfies the following conditions:

{\footnotesize\begin{equation}
    \begin{cases}
        f^{\prime}\left(x\right) = o^* & x\in \mathcal{I} \\
        f^{\prime}\left(a\oplus x\right) = o & a\in \mathcal{A}\ \ \ \  x\in \mathcal{I} ,
    \end{cases}
\end{equation}}where $\oplus$ denotes text concatenation. To quantify the extent of superficial editing, we define the following metrics:

{\footnotesize\begin{equation}
\label{eq:metric_def}
\begin{aligned}
    &\text{OM} = \mathbb{E}_{x}\left[ f^\prime\left(a\oplus x\right) = o  \right] \\
    &\text{OP} = \mathbb{E}_{x} \left[ P\left(o\mid a\oplus x\right) > P\left(o^*\mid a\oplus x \right) \right],
\end{aligned}
\end{equation}}where OM indicates whether the model's prediction matches the original answer $o$, and OP measures whether the output probability of $o$ exceeds that of $o^*$. Higher values of OM and OP reflect a greater degree of superficial editing.

\section{Evaluation of Superficial Editing}
This section evaluates multiple representative parameter-editing algorithms for superficial editing. We first describe the evaluation setup of our experiment (\S\ref{subsec:eval_setup}), followed by a comprehensive assessment of various methods (\S\ref{subsec:eval_res}).

\subsection{Evaluation Setup}
\label{subsec:eval_setup}
\paragraph{Data Collection.} To construct our evaluation dataset for superficial editing, we employ two widely used datasets in knowledge editing: CounterFact (\citealp{rome}) and ZsRE (\citealp{zhu2020ft}). 
First, we select cases where the model has already acquired the corresponding knowledge. Next, based on the definition, we generate three attack prefixes and concatenate them with the baseline prompts from the original dataset to construct attack probes. Finally, we evaluate all instances, filtering the cases that meet the definition to create two enhanced datasets, designated as \textbf{CF-a} and \textbf{ZsRE-a}, respectively. The detailed construction procedure is provided in Appendix \ref{sec:appen_attack_gen}.
% Through systematic augmentation, we generate corresponding datasets incorporating attack probes, designated as \textbf{CF-a} and \textbf{ZsRE-a}, respectively. The detailed construction procedure is provided in Appendix \ref{sec:appen_attack_gen}.

\paragraph{Baselines.} We employ the following knowledge editing methods as baselines: FT (\citealp{zhu2020ft}), MEND (\citealp{mend}), ROME (\citealp{rome}), MEMIT (\citealp{memit}), PMET (\citealp{pmet}), r-ROME (\citealp{r-rome}), and AlphaEdit (\citealp{alphaedit}).

\paragraph{Models \& Metrics.} We conduct experiments using three powerful language models: LLaMA3-8B-Instruct\footnote{\url{https://huggingface.co/meta-llama/Meta-Llama-3-8B-Instruct}}, Qwen2.5-7B-Instruct, and Qwen2.5-14B-Instruct\footnote{\url{https://qwenlm.github.io/blog/qwen2.5-llm/}}. 
In addition to the metrics specifically defined for superficial editing in Equation (\ref{eq:metric_def}), we also report three conventional knowledge editing metrics: Efficacy (Eff.), Generalization (Gen.), and Locality (Loc.), respectively. The formal definitions for them are detailed in Appendix \ref{sec:appen_eval}.

\subsection{Evaluation Results}
\label{subsec:eval_res}

Table \ref{tab:llama3_eval_cf} presents our evaluation results. 
Notably, while the models edited using various methods demonstrate near-perfect performance on conventional metrics, particularly Efficacy, they exhibit significant vulnerability to attack probes. 
For instance, under the Wiki attack scenario, both PMET and AlphaEdit achieve superior Efficacy scores, yet simultaneously maintain high OM metrics of 70.33\% and 72.53\%, respectively. This underscores the severity of superficial editing.
The results also highlight the limitations of conventional evaluation frameworks, which inadequately capture the practical effectiveness of knowledge editing algorithms.
The experimental findings motivate our subsequent investigation into the underlying mechanisms of superficial editing.
% The substantial gap between traditional metric performance and attack probe vulnerability suggests the necessity for more comprehensive evaluation paradigms in future research.

\section{Mechanistic Analysis of Superficial Editing}
This section presents a comprehensive investigation into the underlying mechanisms responsible for superficial editing. We initiate our analysis by examining the influence of the three fundamental components within the Transformer architecture: Residual Stream (\citealp{kaimingresid}; \citealp{elhage2021resid}), Multi-Layer Perceptron (MLP), and Multi-Head Attention. Building upon the observations, we formulate two key hypotheses (\S\ref{subsec:three_components}). Subsequently, we conduct rigorous empirical validation of these hypotheses, systematically elucidating the causal factors underlying superficial editing (\S\ref{subsec:ana_h1} and \S\ref{subsec:ana_h2}). Furthermore, we extend our analysis to investigate the related task of superficial unlearning, thereby demonstrating the generalizability of our approach and conclusions (\S\ref{subsec:unlearning}).

\subsection{Effects of Transformer Components}
\label{subsec:three_components}
% Transformer-based language models consist of three fundamental components: the Residual Stream, the Multi-Layer Perceptron (MLP), and the Attention mechanism. To gain a deeper understanding of their individual contributions to superficial editing, we will examine each component in detail.
\begin{figure}[t]
  \centering
  \begin{subfigure}[t]{0.48\linewidth}
      \centering
      \includegraphics[width=\columnwidth]{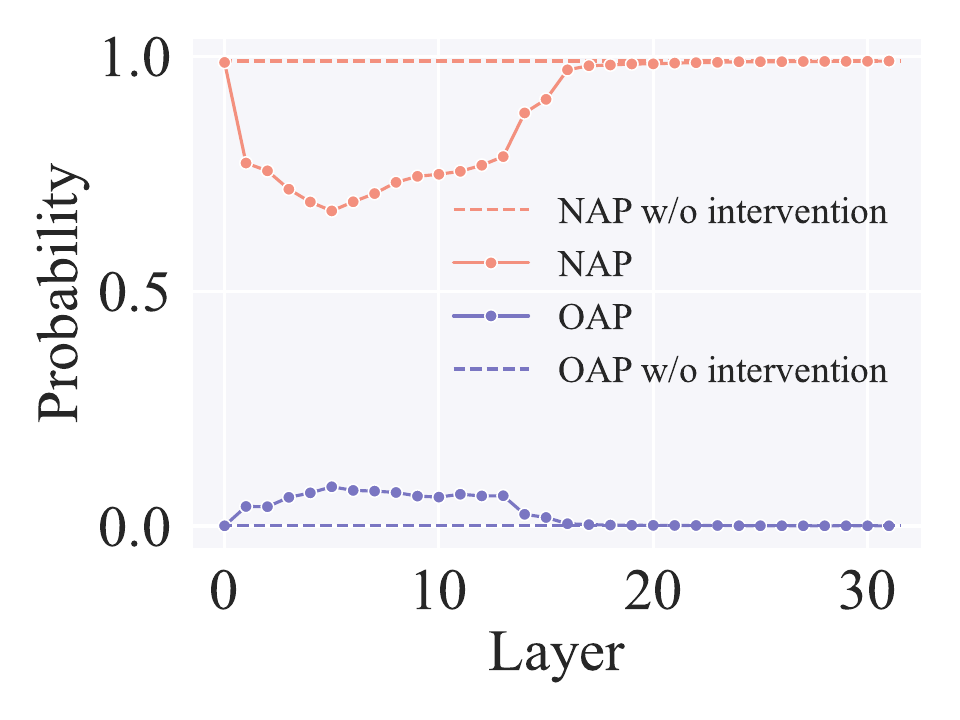}
      \caption{Last subject token.}
      \label{subfig:subjct_last_llama3_rome}
  \end{subfigure}
  \hfill
  \begin{subfigure}[t]{0.48\linewidth}
      \centering
      \includegraphics[width=\columnwidth]{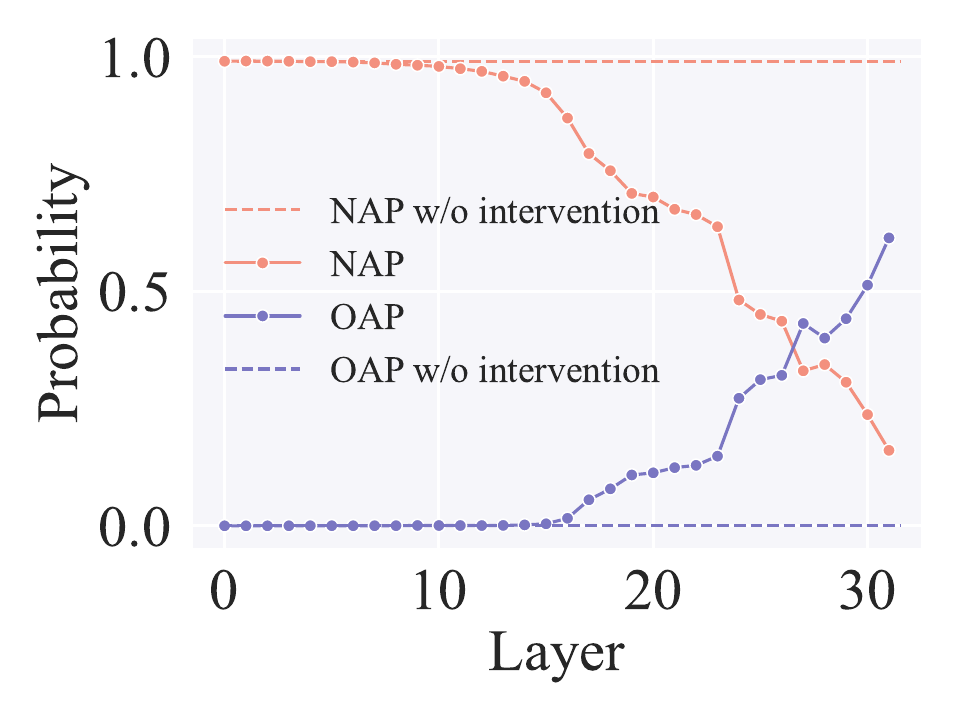}
      \caption{Last token.}
      \label{subfig:last_llama3_rome}
  \end{subfigure}
  % MEMIT
  \begin{subfigure}[t]{0.48\linewidth}
      \centering
      \includegraphics[width=\columnwidth]{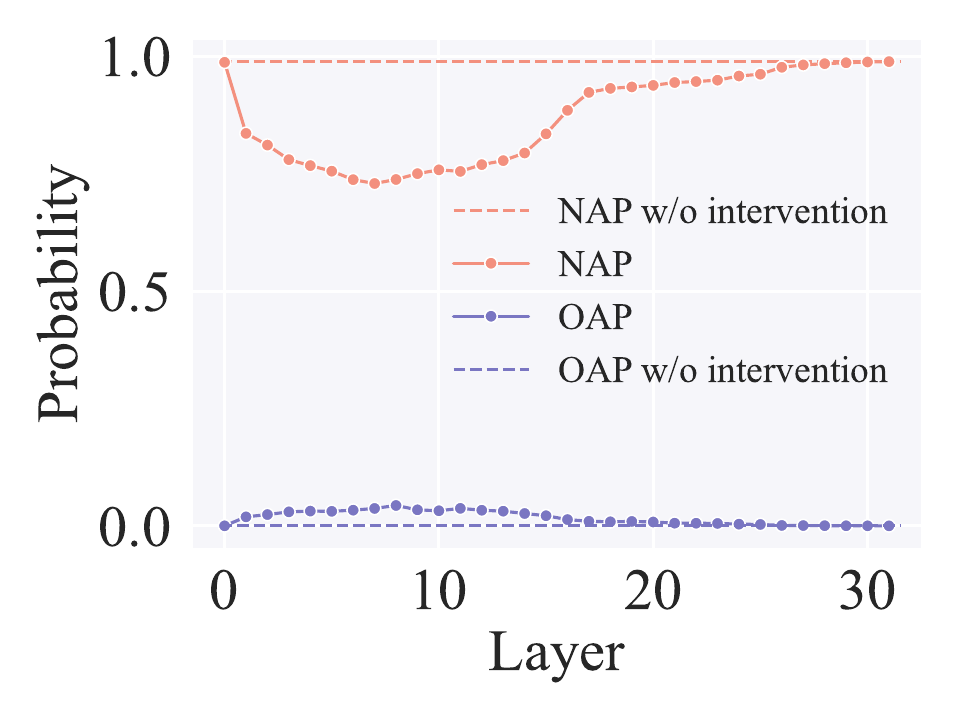}
      \caption{Last subject token.}
      \label{subfig:subjct_last_llama3_memit}
  \end{subfigure}
  \hfill
  \begin{subfigure}[t]{0.48\linewidth}
      \centering
      \includegraphics[width=\columnwidth]{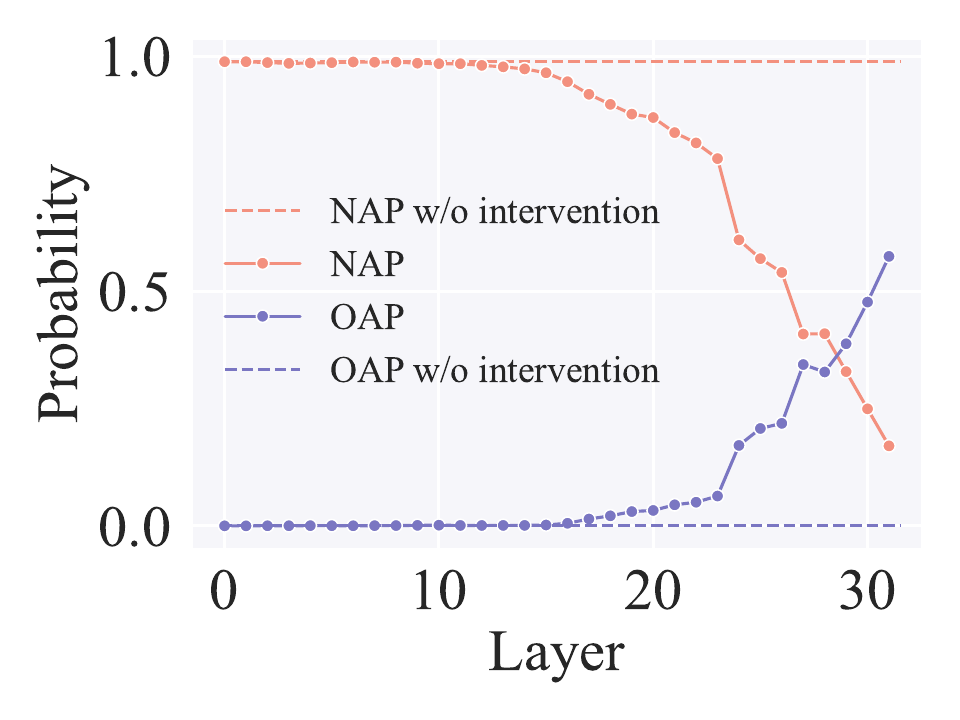}
      \caption{Last token.}
      \label{subfig:last_llama3_memit}
  \end{subfigure}
  % \caption{Results of residual stream intervention on clean run. The figure shows the final probabilities obtained by preforming interventions on each layer of LLaMA3-8B-Instruct edited by ROME and MEMIT}
  \caption {Intervention results of LLaMA3-8B-Instruct edited by ROME (\ref{subfig:subjct_last_llama3_rome}, \ref{subfig:last_llama3_rome}) and MEMIT (\ref{subfig:subjct_last_llama3_memit}, \ref{subfig:last_llama3_memit}) at different tokens.
  % OAP and NAP represents the Original Answer Probability and the New Answer Probability, respectively. 
  The final probabilities without any intervention are depicted by dashed lines in the respective colors. Results for other models are provided in Appendix \ref{subsec:appen_3_components}.}
  \label{fig:resid_patch_llama3_rome}
\end{figure} 

\subsubsection{Effect of the Residual Stream}\label{subsubsec:effect_resid}
To investigate the influence of the Residual Stream, we implement two distinct forward propagation procedures: a ``clean run'' using the baseline prompt $e$ as input and a ``corrupted run'' using the attack probe $a\oplus e$ as input. Through these two forward passes, we can obtain the outputs of each layer in the model:

{\footnotesize\begin{align}
    \boldsymbol{H} &= \{\boldsymbol{h}_i^{\left(l\right)} \mid i\in\left[0, T\right), l\in\left[0, L\right) \}\\
    \hat{\boldsymbol{H}} &= \{ \hat{\boldsymbol{h}}_i^{\left(l\right)} \mid i\in [0, \hat{T} ), l\in\left[0, L\right) \},
\end{align}}where $\boldsymbol{H}$ and $\hat{\boldsymbol{H}}$ denote the hidden states of the clean and corrupted runs, respectively. $T$ and $\hat{T}$ represent the sequence lengths of the two inputs, and $L$ is the number of layers. Subsequently, we introduce an intervention within the residual stream of the clean run. More precisely, we replace the representation of a specific token at layer $l$ with its corresponding representation from the corrupted run at the same layer:

{\footnotesize\begin{equation}
    \boldsymbol{h}_{t_0}^{\left(l\right)} \leftarrow \hat{\boldsymbol{h}}_{t_1}^{\left(l\right)},
\end{equation}}where $t_0$ and $t_1$ denote the indices of the same token in $e$ and $a\oplus e$, respectively.
In this study, we concentrate on two distinct positions: (1) the last position of the subject, which has been identified as crucial for a specific process (\citealp{geva-dissecting}); (2) the last position of the sentence, which serves as the primary basis for the model to predict the next token. 
Following intervention at each layer, we can determine the original answer probability (OAP) and the new answer probability (NAP) of the edited model. To establish a baseline for comparison, we additionally compute the mean OAP and NAP of the clean run without any interventions. The results are illustrated in Figure \ref{fig:resid_patch_llama3_rome}.
As shown in the figure, the residual streams at these two positions exert a causal effect on the model's predictions. The residual stream at the last subject position predominantly influences the earlier layers (Figures \ref{subfig:subjct_last_llama3_rome}, \ref{subfig:subjct_last_llama3_memit}), whereas the residual stream at the last position primarily affects the later layers (Figures \ref{subfig:last_llama3_rome}, \ref{subfig:last_llama3_memit}). The latter's impact is more pronounced, as the OAP exceeds the NAP, which is a critical prerequisite for superficial editing. We formally designate this observed pattern in the later layers as the ``Reversal of the Residual Stream'' (RRS) phenomenon.

\subsubsection{Effect of MLP and Attention}
The impact of both the MLP and Multi-Head Attention on model predictions arises from their iterative refinement of the vector at the last position, thereby enhancing its predictive capacity for generation. 
To investigate their effects, we extract both the input and output vectors at the last position from the MLP and the attention output matrix $\boldsymbol{W}_o$. These vectors are projected into the vocabulary space using the ``logit lens'' technique (\citealp{logitlens0}; \citealp{geva-logitlens}; \citealp{dar-logitlens}; \citealp{halawi-logitlens}), enabling us to observe the probability of the original answer $o$ within each latent probability distribution. A detailed explanation of this technique is provided in Appendix \ref{sec:appen_related_work}.

\begin{figure}[t]
  \centering
  \begin{subfigure}[t]{0.48\linewidth}
      \centering
      \includegraphics[width=\columnwidth]{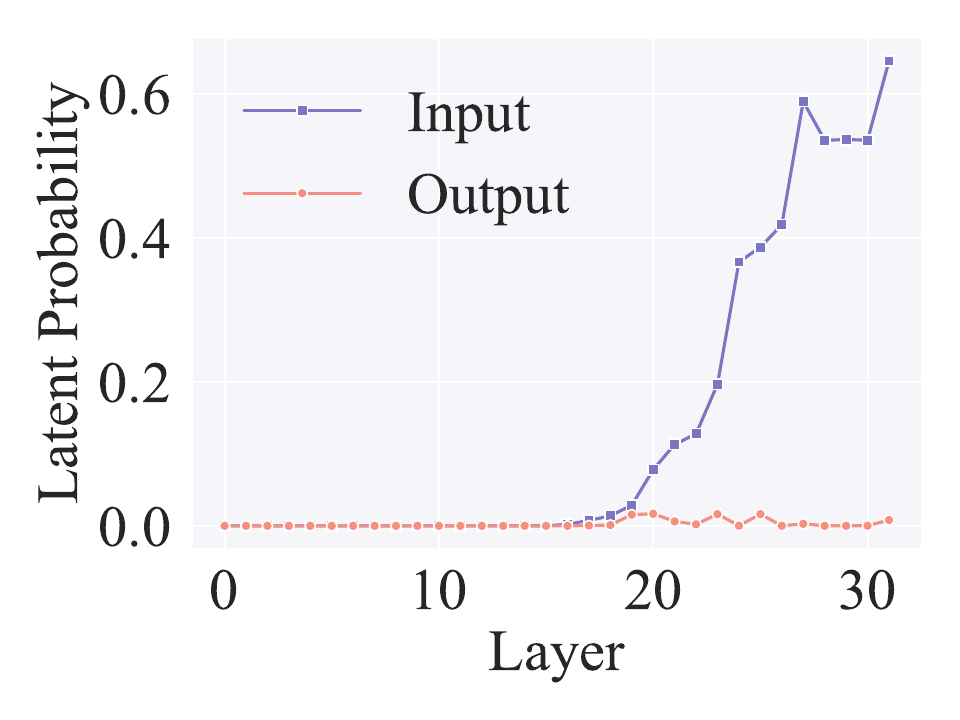}
      \caption{MLP.}
      \label{subfig:llama3_modio_mlp_rome}
  \end{subfigure}
  \hfill
  \begin{subfigure}[t]{0.48\linewidth}
      \centering
      \includegraphics[width=\columnwidth]{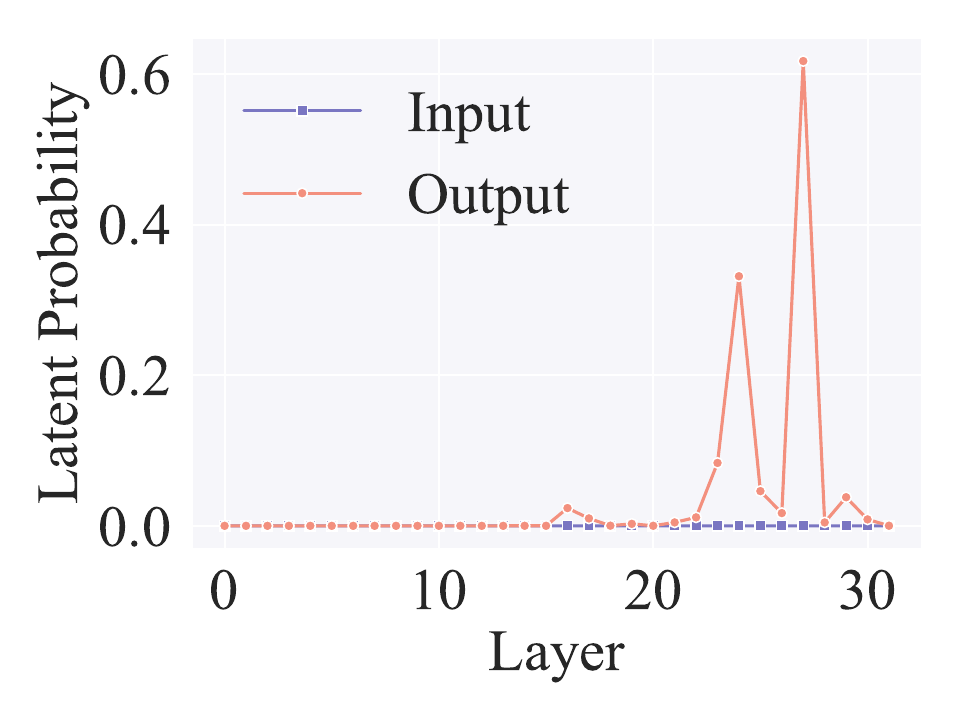}
      \caption{Attention output matrix.}
      \label{subfig:llama3_modio_attn_rome}
  \end{subfigure}
  % MEMIT
    \begin{subfigure}[t]{0.48\linewidth}
      \centering
      \includegraphics[width=\columnwidth]{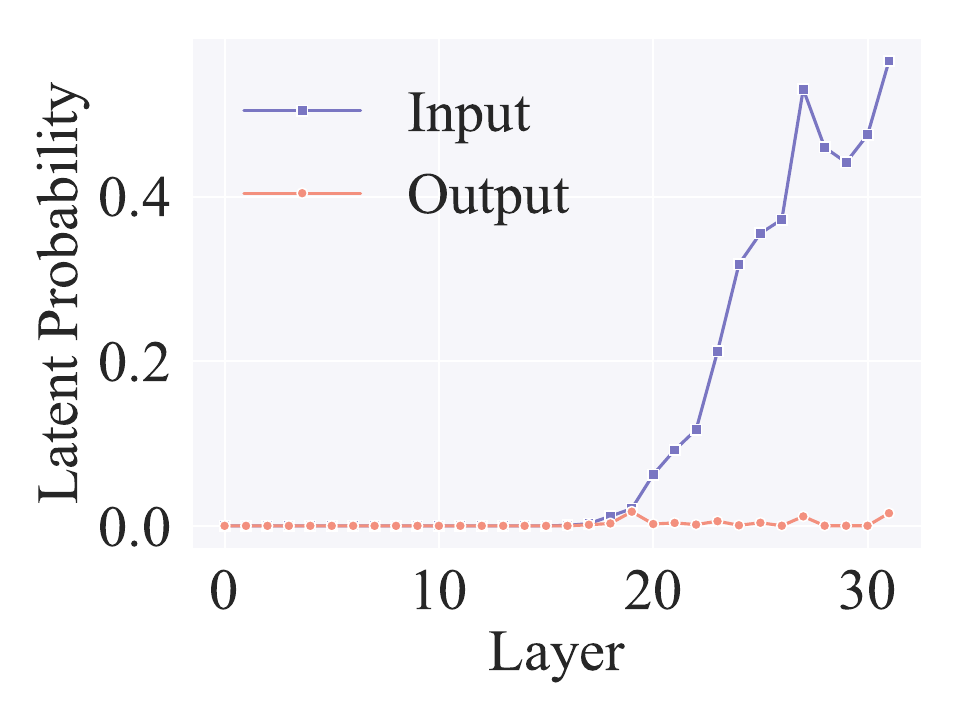}
      \caption{MLP.}
      \label{subfig:llama3_mlp_memit}
  \end{subfigure}
  \hfill
  \begin{subfigure}[t]{0.48\linewidth}
      \centering
      \includegraphics[width=\columnwidth]{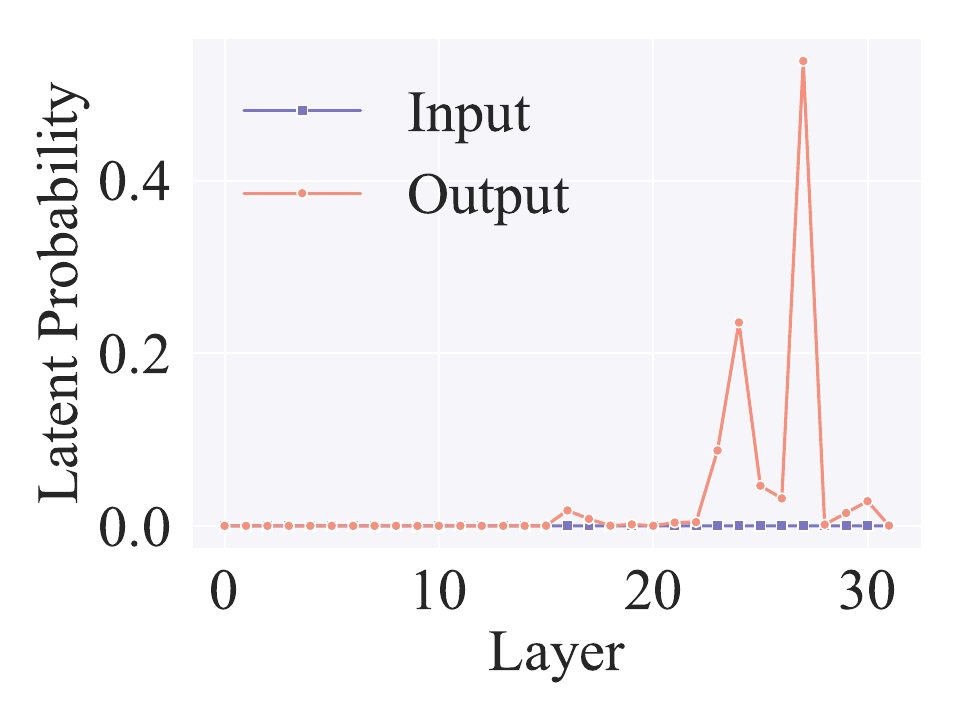}
      \caption{Attention output matrix.}
      \label{subfig:llama3_attn_memit}
  \end{subfigure}
  
  \caption {Latent probabilities of the original answer for the input and output of the MLP and Attention output matrix in LLaMA3-8B-Instruct edited by ROME (\ref{subfig:llama3_modio_mlp_rome}, \ref{subfig:llama3_modio_attn_rome}) and MEMIT (\ref{subfig:llama3_mlp_memit}, \ref{subfig:llama3_attn_memit}). Results for other models are presented in Appendix \ref{subsec:appen_3_components}.}
  \label{fig:llama3_modio_rome}
\end{figure} 

The findings are illustrated in Figure \ref{fig:llama3_modio_rome}. Our analysis demonstrates that the probability of $o$ within the latent probability distribution of each MLP layer's output is consistently lower than that of its input. In contrast, certain attention modules exhibit an inverse pattern in later layers, where the probability of $o$ in the output distribution is significantly higher than that of the input. This observation suggests that the RRS phenomenon is likely driven by attention modules in later layers.

\subsubsection{Insights and Hypotheses}
Through our experiments, we have identified several critical insights: 
(1) The residual stream associated with the last subject position in the earlier layers demonstrates a correlation with superficial editing. 
When considered in conjunction with the subject enrichment process (\citealp{geva-dissecting}), two possible scenarios emerge: either the attack prefix facilitates the accumulation of original knowledge, or it disrupts the enrichment of new knowledge.
% we hypothesize two potential mechanisms: either the attack prefix facilitates the accumulation of original knowledge, or it disrupts the enrichment of new knowledge. 
Given the significant reduction in NAP, the latter appears to be the more plausible explanation. 
(2) Specific later attention layers incorporate information related to $o$ into the last position, indicating that they may contribute to the RRS phenomenon, ultimately leading to superficial editing.

In conclusion, we formulate the following two hypotheses: 
(\textbf{H1}) The enrichment of new knowledge at the last subject position in earlier layers is impeded, and the accumulation of the original knowledge at this position is relatively limited. (\textbf{H2}) The later attention modules actively incorporate information related to original knowledge into the last position, thereby facilitating the RRS phenomenon and consequently inducing the occurrence of superficial editing.

\subsection{Investigation and Validation of H1}\label{subsec:ana_h1}
To validate \textbf{H1}, two propositions must be confirmed: (1) the enrichment of new knowledge within the earlier residual stream is hindered, and (2) the earlier residual stream exhibits negligible accumulation of original knowledge.

\begin{figure}[t]

    \centering
    \begin{subfigure}[t]{0.49\linewidth}
        \centering
        \includegraphics[width=\columnwidth]{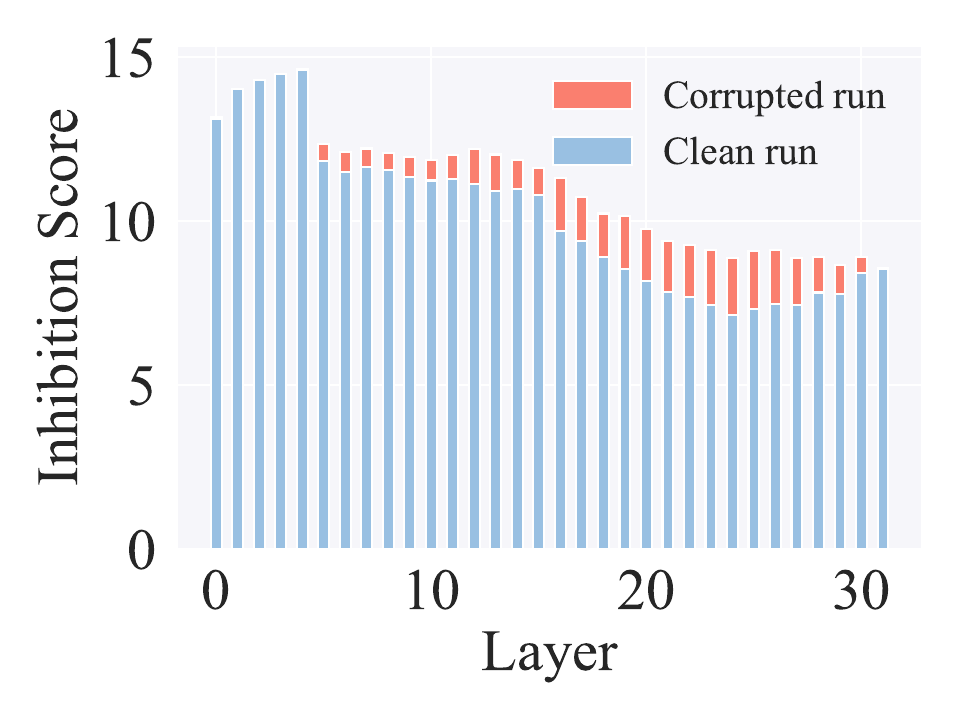}
        \caption{ROME.}
        \label{subfig:suppress_direct_llama3_rome}
    \end{subfigure}
    \hfill
    \begin{subfigure}[t]{0.49\linewidth}
        \centering
        \includegraphics[width=\columnwidth]{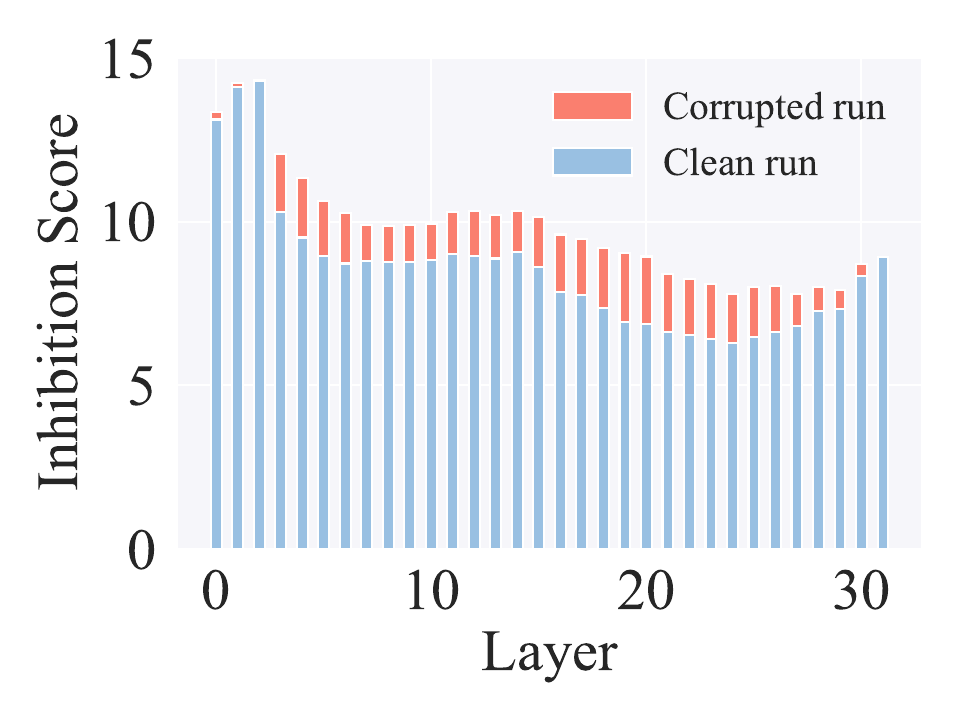}
        \caption{MEMIT.}
        \label{subfig:suppress_direct_llama3_memit}
    \end{subfigure}

  \caption{The Inhibition Scores at each layer for LLaMA3-8B-Instruct edited by ROME and MEMIT. The convex portion of the bar for the corrupted run indicates a higher IS value compared to the clean run. Results for other settings are provided in Appendix \ref{subsec:appen_h1}.}
  \label{fig:llama3_logp}
\end{figure}

To confirm proposition (1), we extract the representations of the last subject position from both the clean and corrupted runs. 
To quantify the suppression effect, we introduce the Inhibition Score (IS):

{\footnotesize\begin{equation}
   \text{IS}^{\left(l\right)}\left(o^*\right) = -\log P_{LL}\left(o^*\mid h_j^{\left(l\right)}\right),
\end{equation}}where $h_j^{\left(l\right)}$ denotes the representation of the last subject token, and $P_{LL}$ represents the latent probability of $o^*$ derived from the logit lens. 
% % 这样定义比较好：
% \begin{equation}
%    \text{IS}^{\left(l\right)}\left(o^*\right) = \log \frac{P_{LL}\left(o^*\mid h_j^{\left(l\right)}\right)}{P_{LL}\left(o^*\mid \hat{h}_j^{\left(l\right)}\right)},
% \end{equation}
% TODO: 直接用clean run和corrupted run对比也可以，现在这种干预的做法有些繁琐
% To confirm proposition (1), we extract the last subject representation from the clean run and replace it with the corresponding representation from the corrupted run. For LLaMA3-8B-Instruct, the influence of the last subject residual stream is observable only up to layer 16. Accordingly, we devide layers 1-15 into three distinct windows, each containing five layers., and we apply the aforementioned interventions to the residual streams within each window individually. To quantify the suppression effect, we introduce the Inhibition Score (IS):
% % 如果叫这个名字，那么无干预时应该是0才对？
% \begin{equation}
%    \text{IS}^{\left(l\right)}\left(o^*\right) = -\log P_{LL}\left(o^*\mid h_j^{\left(l\right)}\right),
% \end{equation}
% A positive IS value suggests that the enrichment of the new knowledge is inhibited in the corrupted run. 
A higher inhibition score indicates a stronger inhibitory effect.
The results are illustrated in Figure \ref{fig:llama3_logp}. 
Our analysis reveals that the negative logarithmic probability of the new answer decreases gradually, indicating a corresponding increase in its latent probability. Furthermore, the IS value for the corrupted run exceeds that of the clean run in earlier layers (e.g., layers 5-15), suggesting that the enrichment of new knowledge is inhibited.
% Our analysis reveals a progressive increase in the latent probability of $o^*$ corresponding to the depth of layers.
% Furthermore, the interventions result in an increased inhibition score in earlier layers (e.g., layers 5-15), indicating suppression of new knowledge enrichment. 
% Furthermore, interventions demonstrate a elevation in the inhibition score, providing evidence for the suppression of the new knowledge enrichment.
% It can be noted that, compared to the scenario without any intervention, the aggregation of new knowledge within the residual stream is impeded after the intervention, resulting in a diminished amount of information regarding the new answer in the last subject position. This indicates that the attack prefix indeed obstructs the aggregation of new knowledge at the last subject position.

% 重新画，附录也要改

\begin{figure}[t]
    \centering
    \begin{subfigure}[t]{0.48\linewidth}
        \centering
        \includegraphics[width=\columnwidth]{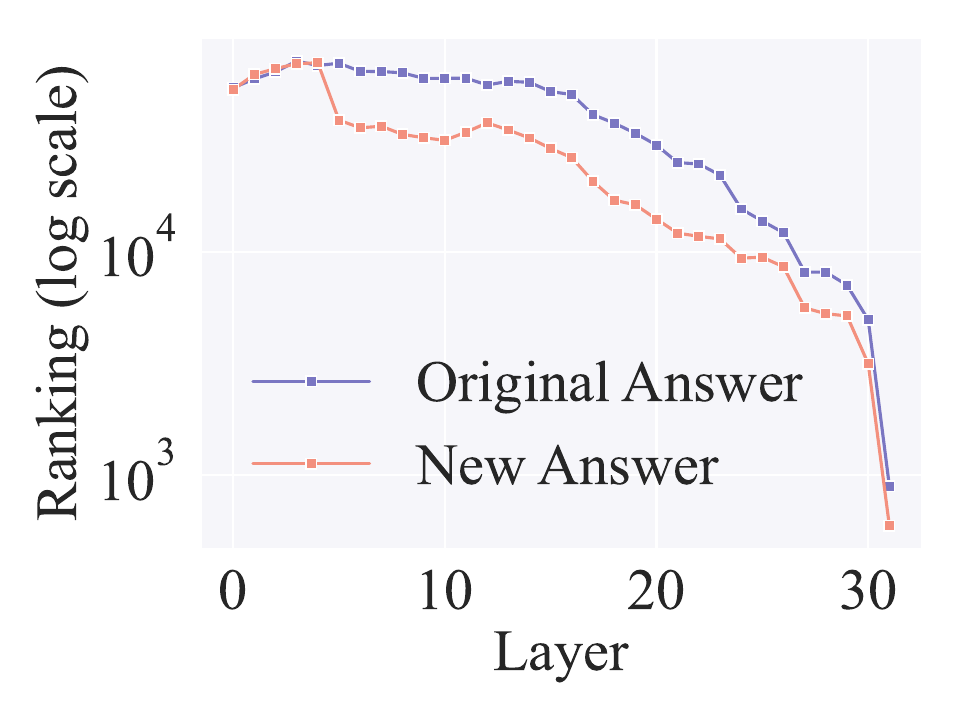}
        \caption{ROME.}
        \label{subfig:rank_llama3_rome}
    \end{subfigure}
    \hfill
    \begin{subfigure}[t]{0.48\linewidth}
        \centering
        \includegraphics[width=\columnwidth]{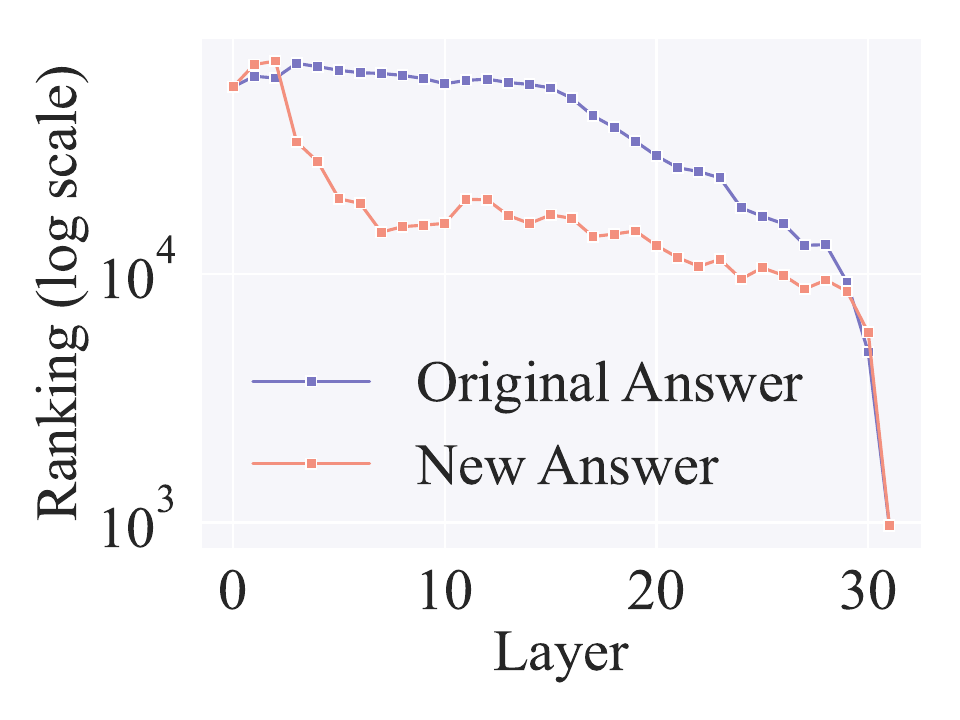}
        \caption{MEMIT.}
        \label{subfig:rank_llama3_memit}
    \end{subfigure}
  
  \caption{The rankings of $o$ and $o^*$ in the latent probability distribution at the last subject token for LLaMA3-8B-Instruct edited by ROME and MEMIT. Results for other models are provided in Appendix \ref{subsec:appen_h1}.}
  \label{fig:llama3_rank}
\end{figure}
To confirm proposition (2), we compute the rankings of $o$ and $o^*$ within the latent probability distributions of the last subject position across all layers in the corrupted runs. 
As shown in Figure \ref{fig:llama3_rank}, the ranking of the original answer consistently falls behind that of the new answer in earlier layers. Combined with our prior analysis, despite the suppression of new knowledge enrichment, the ranking of $o$ fails to surpass that of $o^*$ in earlier layers, indicating negligible accumulation of original knowledge at this specific position.

\subsection{Investigation and Validation of H2}\label{subsec:ana_h2}
To validate \textbf{H2}, we first establish a causal relationship between the later attention modules and the ``Reversal of the Residual Stream'' (RRS) phenomenon, highlighting the crucial role of attention for superficial editing (\S\ref{subsubsection:attn1}). Following this, we demonstrate that specific attention heads within the later attention modules actively integrate information related to the original answer into the last position. Additionally, we demonstrate a causal relationship between these attention heads and the occurrence of superficial editing (\S\ref{subsubsection:attn2}). 
To further understand the internal mechanisms, we apply singular value decomposition (SVD) to the output matrices of these heads, revealing that the linear combination of certain left singular vectors encapsulates information associated with original knowledge, contributing to superficial editing (\S\ref{subsubsection:attn_svd}).

\subsubsection{The Role of Attention}\label{subsubsection:attn1}
\begin{figure}[t]
    \centering
    \begin{subfigure}[t]{0.48\linewidth}
        \centering
        \includegraphics[width=\columnwidth]{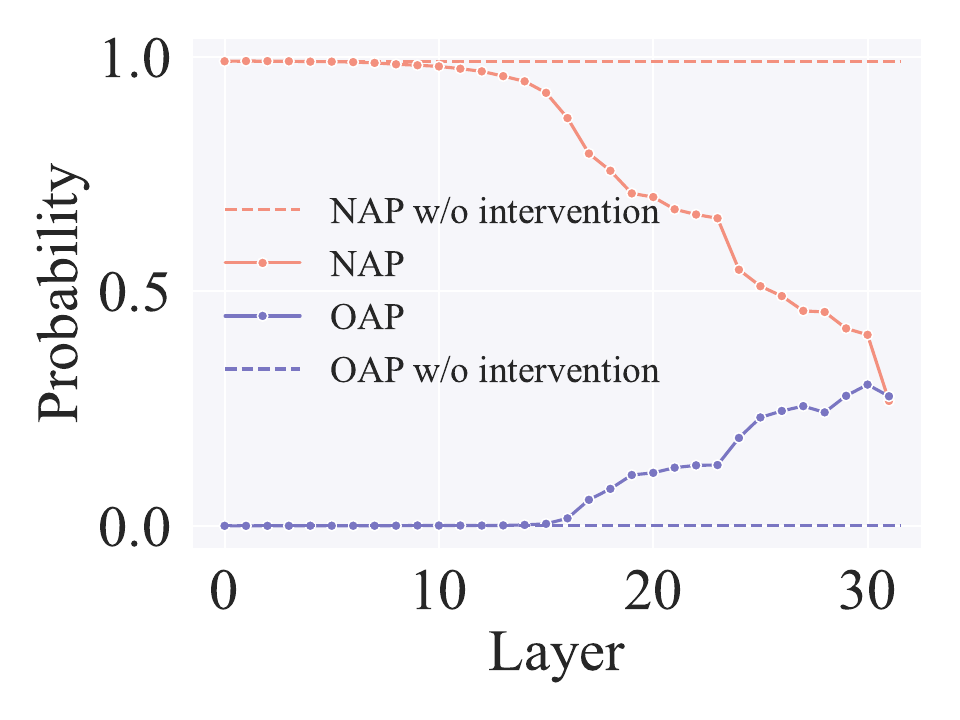}
        \caption{ROME.}
        \label{subfig:llama3_patch_wo_attn_rome}
    \end{subfigure}
    \hfill
    \begin{subfigure}[t]{0.48\linewidth}
        \centering
        \includegraphics[width=\columnwidth]{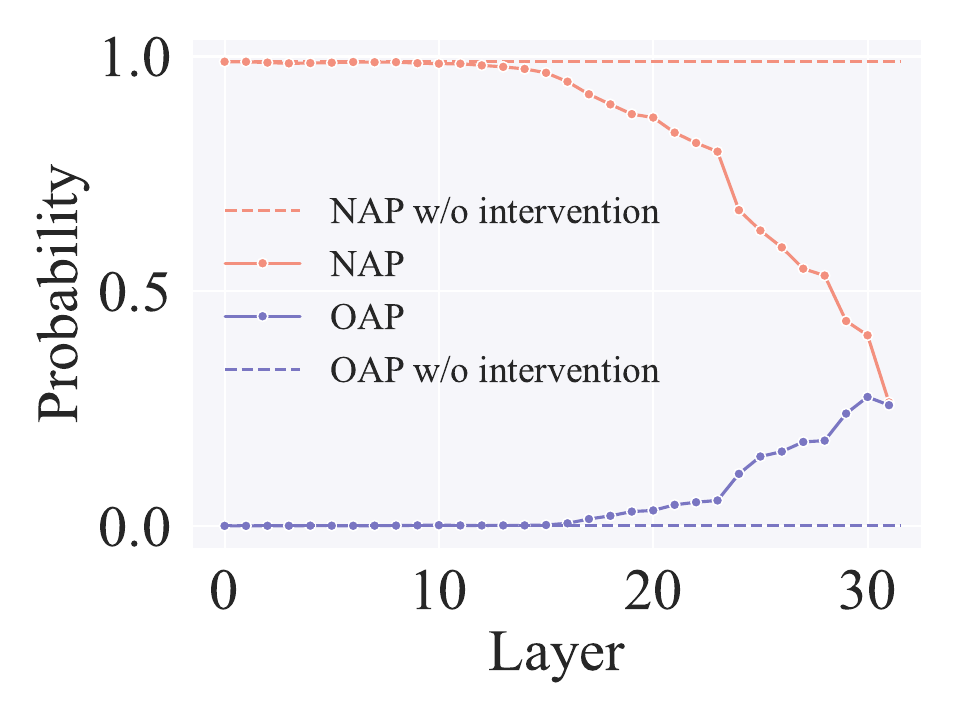}
        \caption{MEMIT.}
        \label{subfig:llama3_patch_wo_attn_memit}
    \end{subfigure}
   \caption{Intervention effects following critical attention module ablation in LLaMA3-8B-Instruct edited by ROME and MEMIT. We present the results of other models in Appendix \ref{subsec:appen_h2}.}
  % \caption{Probabilities of $o$ and $o^*$ of the clean run, which is intervened by the attention-ablated corrupted run. Figure \ref{subfig:llama3_patch_wo_attn_rome} and \ref{subfig:llama3_patch_wo_attn_memit} illustrates the results of LLaMA3-8B-Instruct edited by ROME and MEMIT, respectively. We present the results of other models in Appendix \ref{subsec:appen_h2}}
  \label{fig:llama3_rome_patch_wo_attn}
\end{figure}

To investigate the correlation between the RRS phenomenon and the attention modules in later layers, we set the output of selected critical attention layers (e.g., layer 27 in Figure \ref{subfig:llama3_modio_attn_rome}) to zero and extract representations from all layers during the corrupted run. Following the method in Section \ref{subsubsec:effect_resid}, we substitute the representation at the last position in the clean run and compute the final probabilities of $o$ and $o^*$. The results are presented in Figure \ref{fig:llama3_rome_patch_wo_attn}.
A comparative analysis between Figures \ref{subfig:last_llama3_rome}, \ref{subfig:last_llama3_memit}, and \ref{fig:llama3_rome_patch_wo_attn} demonstrates that after ablating the specific attention modules, NAP is no longer surpassed by OAP, indicating that the RRS phenomenon has been mitigated.
% the ablation of later attention modules eliminates the probability dominance of $o$ over $o^*$. 
This observation establishes a significant correlation between these attention modules and superficial editing.

\subsubsection{The Role of Attention Head}\label{subsubsection:attn2}
Our analysis has revealed a significant correlation between specific later attention modules and the occurrence of superficial editing. This naturally leads to the question regarding the mechanistic pathways by which these later attention modules influence the final predictions. To explore this, we conduct a head-level analysis of attention mechanisms. 
Let $\boldsymbol{x}^{\left(l\right)}$ denote the input vector to the attention output matrix $\boldsymbol{W}^{\left(l\right)}_O$ at the last position. Through the logit lens technique, we derive the latent original probability of each head (LOPH):

{\footnotesize\begin{equation}
\label{eq:attn_head_outp}
    \text{LOPH} = P_{LL}\left(o\mid \boldsymbol{W}_O^{\left(l, h\right)}\boldsymbol{x}^{\left(l,h\right)} \right),
\end{equation}}where $\boldsymbol{W}_O^{\left(l, h\right)}$ represents the output matrix for the $h$-th head, with $\boldsymbol{x}^{\left(l,h\right)}$ denoting its corresponding input vector.
% Through the logit lens technique, we derive the latent original probability of each head (LOPH):
% The matrix $\boldsymbol{W}^{\left(l\right)}_O$ can partitioned into multiple sub-matrices corresponding to  different attention heads:
% {\footnotesize\begin{equation}
%     \boldsymbol{W}^{\left(l\right)}_O = \left[ \boldsymbol{W}_O^{\left(l, 0\right)} |\ \boldsymbol{W}_O^{\left(l, 1\right)} | \dots | \boldsymbol{W}_O^{\left(l, N_h-1\right)} \right].
% \end{equation}}

\begin{figure}[t]
    \centering
    \begin{subfigure}[t]{0.48\linewidth}
        \centering
        \includegraphics[width=\columnwidth]{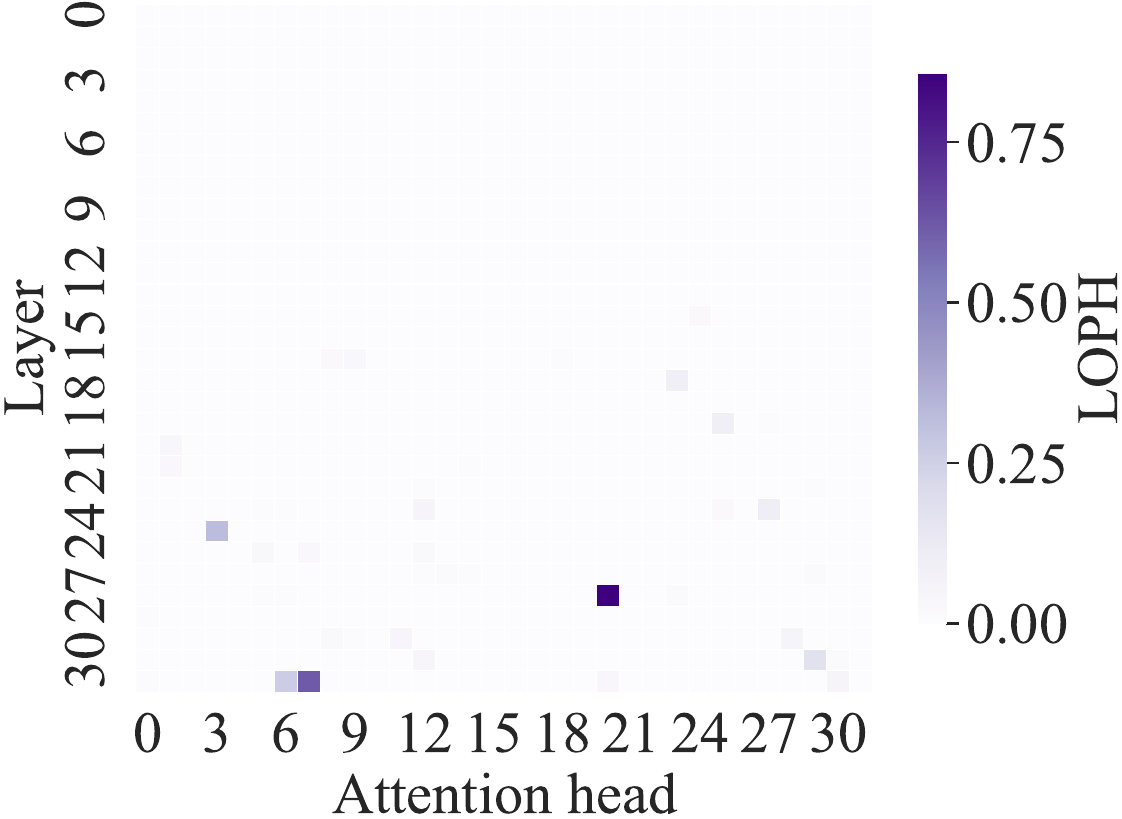}
        \caption{ROME.}
        \label{subfig:head_heatmap_rome}
    \end{subfigure}
    \hfill
    \begin{subfigure}[t]{0.48\linewidth}
        \centering
        \includegraphics[width=\columnwidth]{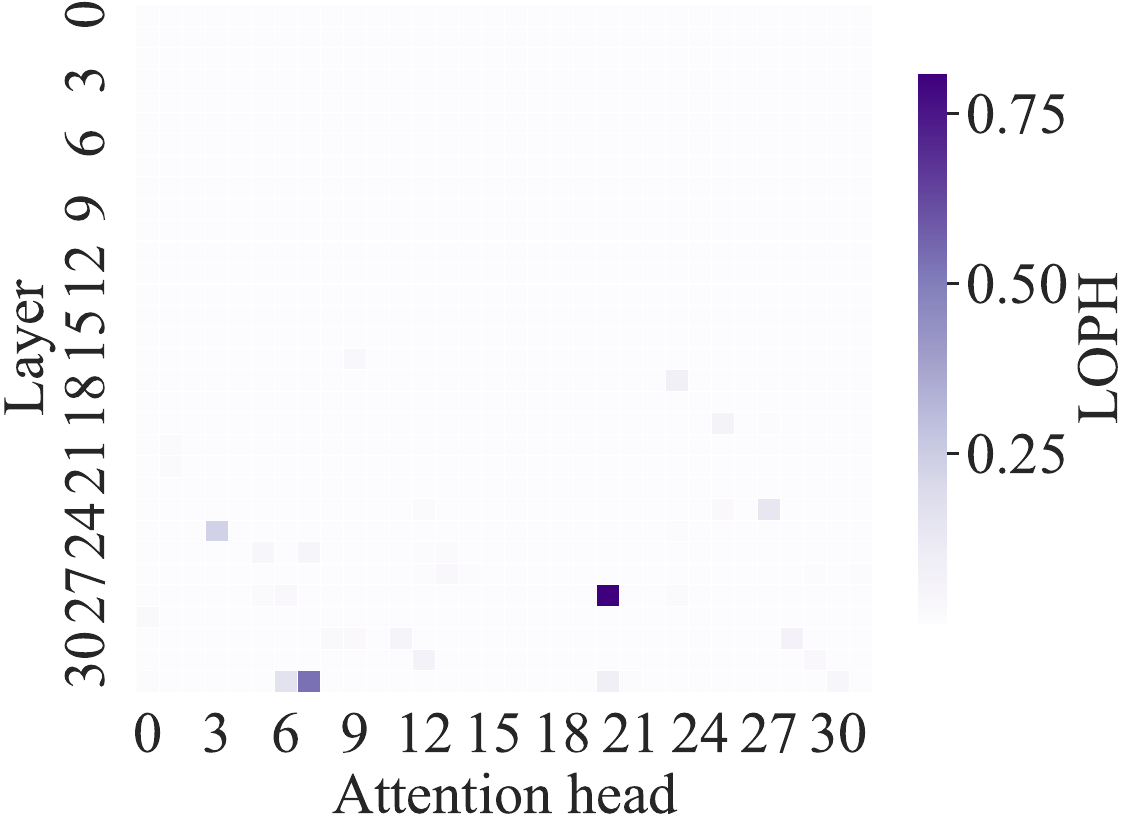}
        \caption{MEMIT.}
        \label{subfig:head_heatmap_memit}
    \end{subfigure}
  
  \caption{LOPH of LLaMA3-8B-Instruct edited by ROME and MEMIT. Results for other models are provided in Appendix \ref{subsec:appen_h2}.}
  \label{fig:llama3_heatmap_rome}
\end{figure}

\begin{table}
  \centering
  \scalebox{0.6}{\begin{tabular}{c|l|ccc|ccc}
    % \hline
    \toprule
    \multirow{2}{*}{\textbf{Models}}  & \multirow{2}{*}{\textbf{Methods}} & \multicolumn{3}{c|}{\textbf{Original}} & \multicolumn{3}{c}{\textbf{New}} \\ 
    & & \textbf{w/o abl.} & \textbf{abl.} & $\downarrow \Delta P$ & \textbf{w/o abl.} & \textbf{abl.} & $\uparrow \Delta P$ \\
    % \hline
    \midrule
    \multirow{2}{*}{\shortstack{LLaMA3-\\ 8B-Instruct}} & ROME & 57.17  & 35.58 & 21.59 & 16.49 & 20.71 & 4.22 \\
     & MEMIT & 56.90 & 37.36 & 19.54 & 15.68 & 18.38 & 2.70 \\
     \midrule
   \multirow{2}{*}{\shortstack{Qwen2.5-\\ 7B-Instruct}} & ROME & 57.83 & 36.52 & 21.31 & 11.84 & 17.57 & 5.73 \\
     & MEMIT & 57.54 & 32.40 & 25.14 & 12.21 & 26.08 & 13.87 \\
     \midrule
   \multirow{2}{*}{\shortstack{Qwen2.5-\\ 14B-Instruct}} & ROME & 55.71 & 39.99 & 15.72 & 13.99 & 21.40 & 7.41 \\
     & MEMIT & 55.03 & 37.25 & 17.78 & 13.79 & 22.24 & 8.45 \\
    % \hline
    \bottomrule
  \end{tabular}}
  \caption{Ablation effects of the prominent heads. 
  (Original: original answer; New: new answer; w/o abl.: without ablation; abl.: with ablation; $\downarrow \Delta P$: probability decrease; $\uparrow \Delta P$: probability increase)
  % \textbf{Original} and \textbf{New} represents $o$ and $o^*$, respectively. $\downarrow \Delta P$ signifies a reduction in probability, whereas $\uparrow \Delta P$ denotes an increase in probability.
  }
  \label{tab:head_abl_document}
\end{table}

The results of LOPH are depicted in Figure \ref{fig:llama3_heatmap_rome}. Our analysis demonstrates that specific attention heads integrate information related to the original knowledge into the last position. This observation suggests that these prominent attention heads may play a significant role in facilitating superficial editing.
To validate the causal relationship, we perform the corrupted run by zeroing the output of attention heads with LOPH values exceeding $\tau$. 
% If $\tau$ is excessively large, the attention heads under investigation may fail to capture all significant heads. Conversely, if $\tau$ is too small, irrelevant heads may also be included. After carefully balancing these two considerations, we set $\tau$ to 0.1. 
When $\tau$ is too large, the attention heads under investigation may miss significant heads. Conversely, a small $\tau$ may include irrelevant heads. After carefully balancing these two considerations, we set $\tau$ to 0.1.
We then examine the model's output probabilities for both $o$ and $o^*$, with quantitative results provided in Table \ref{tab:head_abl_document}. 
% -------cameral ready
% Additional experimental results for varying values of $\tau$ are provided in Appendix \ref{subsec:appen_h2}.
The results demonstrate a decrease in the probability of $o$, accompanied by a corresponding increase in the probability of $o^*$ after the removal of these attention heads. This suggests partial mitigation of superficial editing, providing evidence for the causal role of these attention heads.

\begin{table*}
  \centering
\scalebox{0.8}{\begin{tabular}{c|c|cc|cc|cc|cc|cc|cc}
\toprule
\multirow{2}{*}{} & \multirow{2}{*}{\textbf{Top-K}} & \multicolumn{2}{c|}{\textbf{L23H27}} & \multicolumn{2}{c|}{\textbf{L24H3}} & \multicolumn{2}{c|}{\textbf{L27H20}} & \multicolumn{2}{c|}{\textbf{L30H29}} & \multicolumn{2}{c|}{\textbf{L31H6}} & \multicolumn{2}{c}{\textbf{L31H7}} \\ 
% \cline{3-6}
% \midrule
&   & 5\% & 10\% & 5\%  & 10\% & 5\%  & 10\% & 5\%  & 10\% & 5\%  & 10\% & 5\%  & 10\%  \\ 
\midrule
\multirow{3}{*}{Original}  
&5  & 6.06 & 9.85 & 50.76 & 62.88 & 64.39 & 71.97 & 14.39 & 16.67 & 62.88 & 73.48 & 62.88 & 71.97 \\
 & 10 & 6.82 & 10.61 & 57.58 & 67.42 & 67.42 & 73.48 & 15.15 & 16.67 & 65.15 & 75.00 & 65.15 & 72.73  \\
& 15  & 10.61 & 11.36 & 61.36 & 68.18 & 70.45 & 75.76 & 15.91 & 16.67 & 67.42 & 75.76 & 65.91 & 73.48   \\ 
\midrule
\multirow{3}{*}{New}   
& 5   & 0.00 & 0.00 & 6.06 & 4.55 & 4.55 & 3.79 & 2.27 & 2.27 & 2.27 & 1.52 & 2.27 & 3.03 \\
& 10  & 0.00 & 0.76 & 6.82 & 6.06 & 6.06 & 6.82 & 4.55 & 3.79 & 5.30 & 4.55 & 5.30 & 6.06  \\
& 15   &  0.00 & 0.76 & 7.58 & 6.82 & 6.82 & 9.09 & 8.33 & 4.55 & 6.82 & 5.30 & 7.58 & 9.09 \\ 
\bottomrule
\end{tabular}}
  \caption{Decoding Success Rate (DSR) of the identified vectors across different heads in LLaMA3-8B-Instruct edited by ROME. $p$\% is set to 5\% and 10\%. Results for other settings are provided in Appendix \ref{subsec:appen_h2}.}
  \label{tab:llama3_sdr_rome_document}
\end{table*}

\subsubsection{Dissection of Attenion Head}\label{subsubsection:attn_svd}
To elucidate the efficacy of these attention heads for superficial editing, we perform singular value decomposition on $\boldsymbol{W}_O^{\left(l, h\right)}$. Given the last position vector $\boldsymbol{x}^{\left(l, h\right)}$ of the input, we have:

{\footnotesize\begin{equation}
    \begin{aligned}
    \boldsymbol{z} &= \boldsymbol{W}_O^{\left(l, h\right)}\boldsymbol{x}^{\left(l, h\right)} =   \sum_{i=0}^{r-1} \left(\boldsymbol{u}_i \sigma_i \boldsymbol{v}_i^\top\right) \boldsymbol{x}^{\left(l, h\right)} \\
    &= \sum_{i=0}^{r-1}\boldsymbol{u}_i\sigma_i\left(\boldsymbol{v}_i^\top\boldsymbol{x}^{\left(l, h\right)} \right) 
    = \sum_{i=0}^{r-1}\lambda_i\boldsymbol{u}_i, 
\end{aligned}
\end{equation}}where $\lambda_i=\sigma_i \boldsymbol{v}_i^\top \boldsymbol{x}^{\left(l, h\right)}$ is a scalar. This equation demonstrates that the output of an attention head can be expressed as a linear combination of the left singular vectors derived from its output matrix, with the coefficients determined by the input. 
Consequently, we hypothesize that the superficial editing induced by attention heads is attributable to specific left singular vectors. 
We set the coefficient of the $i$-th singular vector to 0 to derive $z_{\text{abl}}^{\left(i\right)}$ and identify the top $p$\% most significant vectors through the following procedure:

{
\footnotesize
\begin{equation}
    \mathcal{S}_u = \text{Top-P}\left[ P_{LL}\left(o\mid\boldsymbol{z}\right) - P_{LL}\left(o\mid\boldsymbol{z}^{\left(i\right)}_{\text{abl}} \right) \right].
\end{equation}
}We define the Decoding Success Rate (DSR) to assess whether the linear combination of the identified vectors captures the target knowledge:

{\footnotesize\begin{equation}
    \text{DSR} = \mathbb{E}\left[ \mathbbm{1}\left[ t\in \text{Top-K}\left( \boldsymbol{z}\left(\mathcal{S}_u\right) \right) \right] \right],
\end{equation}}where $t$ is the target token ($o$ or $o^*$), $\text{Top-K}\left( \boldsymbol{z}\left(\mathcal{S}_u\right) \right)$ denotes the first $K$ tokens derived from decoding the linear combination of the identified vectors via the logit lens. The results in Table \ref{tab:llama3_sdr_rome_document} demonstrate that across all heads, the DSR of $o$ consistently exceeds that of $o^*$ by a large margin, supporting our hypothesis.
% We propose a strategy to identify top $p$\% vectors in Appendix \ref{subsec:appen_h2} and demonstrate that their linear combination encapsulates the original knowledge.

To further examine the causal relationship between these left singular vectors and superficial editing, we perform an ablation study on the identified crucial vectors during forward propagation and observe the probabilities of the model generating both $o$ and $o^*$. 
The experimental results, presented in Table \ref{tab:abl_svd_document}, illustrate that the removal of the identified singular vectors leads to a decrease in OAP and an increase in NAP. These findings demonstrate that the identified singular vectors causally contribute to superficial editing.

\begin{table*}
\centering
\scalebox{0.7}{\begin{tabular}{c|c|cccccc|cccccc}
\toprule
\multirow{3}{*}{\textbf{Models}} & \multirow{3}{*}{\textbf{Methods}} & \multicolumn{6}{c|}{5\%} & \multicolumn{6}{c}{10\%}\\ 

& & \multicolumn{3}{c}{\textbf{OAP}} & \multicolumn{3}{c|}{\textbf{NAP}} & \multicolumn{3}{c}{\textbf{OAP}} & \multicolumn{3}{c}{\textbf{NAP}} \\ 
 & & w/o abl. & abl. & $\downarrow\Delta P$ & w/o abl.  & abl. & $\uparrow\Delta P$ & w/o abl. & abl. & $\downarrow\Delta P$ & w/o abl.  & abl. & $\uparrow\Delta P$  \\ 
\midrule
\multirow{2}{*}{\shortstack{LLaMA3-8B-\\ Instruct}} & ROME & 61.41 & 52.80 & 8.61 & 16.12 & 20.48 & 4.36 & 61.41 & 48.64 & 12.77 & 16.12 & 22.65 & 6.53  \\
 & MEMIT & 57.42 & 48.61 & 8.81 & 17.05 & 21.64 & 4.59 & 57.42 & 44.42 & 13.00 & 17.05 & 23.48 & 6.43 \\ 
\midrule
\multirow{2}{*}{\shortstack{Qwen2.5-7B-\\ Instruct}} & ROME & 64.33 & 55.91 & 8.42 & 11.93 & 17.05 & 5.12 & 64.33 & 51.11 & 13.22 & 11.93 & 19.44 & 7.51 \\
& MEMIT & 66.41 & 61.83 & 4.58 & 15.72 & 19.01 & 3.29 & 66.41 & 58.25 & 8.16 & 15.72 & 21.31 & 5.59 \\
  \midrule
\multirow{2}{*}{\shortstack{Qwen2.5-14B-\\ Instruct}} & ROME  & 62.87 & 56.94 & 5.93 & 17.39    & 20.79 & 3.40 & 62.87 & 53.78 & 9.09 & 17.39 & 22.69 & 5.30  \\
 &  MEMIT & 62.02 & 55.24  & 6.78 & 15.64 & 19.63   & 3.99 & 62.02 & 51.52 & 10.50 & 15.64 & 21.77 & 6.13 \\
\bottomrule
\end{tabular}}
  \caption{Answer probabilities before and after singular vector ablation. 
  % ``w/o abl.'' and ``abl.'' represents ``without ablation'' and ``ablation'', respectively. $\downarrow \Delta P$ and $\uparrow \Delta P$ denote the probability decrease and increase, respectively.
  }
  \label{tab:abl_svd_document}
\end{table*}

% ECDF plot.
% \begin{figure}[t]
%     \centering
%     \begin{subfigure}[t]{0.48\linewidth}
%         \centering
%         \includegraphics[width=\columnwidth]{figures/abl_svd/llama3_0.05_rome.pdf}
%         \caption{ROME\ \ top5\%.}
%         \label{subfig:abl_svd_llama3_0.05_rome}
%     \end{subfigure}
%     \hfill
%     \begin{subfigure}[t]{0.48\linewidth}
%         \centering
%         \includegraphics[width=\columnwidth]{figures/abl_svd/llama3_0.05_memit.pdf}
%         \caption{MEMIT\ \ top5\%.}
%         \label{subfig:abl_svd_llama3_0.05_memit}
%     \end{subfigure}
%     % 0.1
%     \begin{subfigure}[t]{0.48\linewidth}
%         \centering
%         \includegraphics[width=\columnwidth]{figures/abl_svd/llama3_0.1_rome.pdf}
%         \caption{ROME\ \ top10\%.}
%         \label{subfig:abl_svd_llama3_0.1_rome}
%     \end{subfigure}
%     \hfill
%     \begin{subfigure}[t]{0.48\linewidth}
%         \centering
%         \includegraphics[width=\columnwidth]{figures/abl_svd/llama3_0.1_memit.pdf}
%         \caption{MEMIT\ \ top10\%.}
%         \label{subfig:abl_svd_llama3_0.1_memit}
%     \end{subfigure}

%   \caption{The results of ablating the identified singular vectors. $p$\% is set to 5\% and 10\%. For a specified probability $q$ on the horizontal axis, the corresponding vertical axis value represents the proportion of data points with values less than or equal to $q$.}
%   \label{fig:llama3_ablsvd_0.05_rome}
% \end{figure}

\subsection{Superficial Unlearning}\label{subsec:unlearning}
\begin{figure}[t]
\centering
  \scalebox{0.8}{\includegraphics[width=\columnwidth]{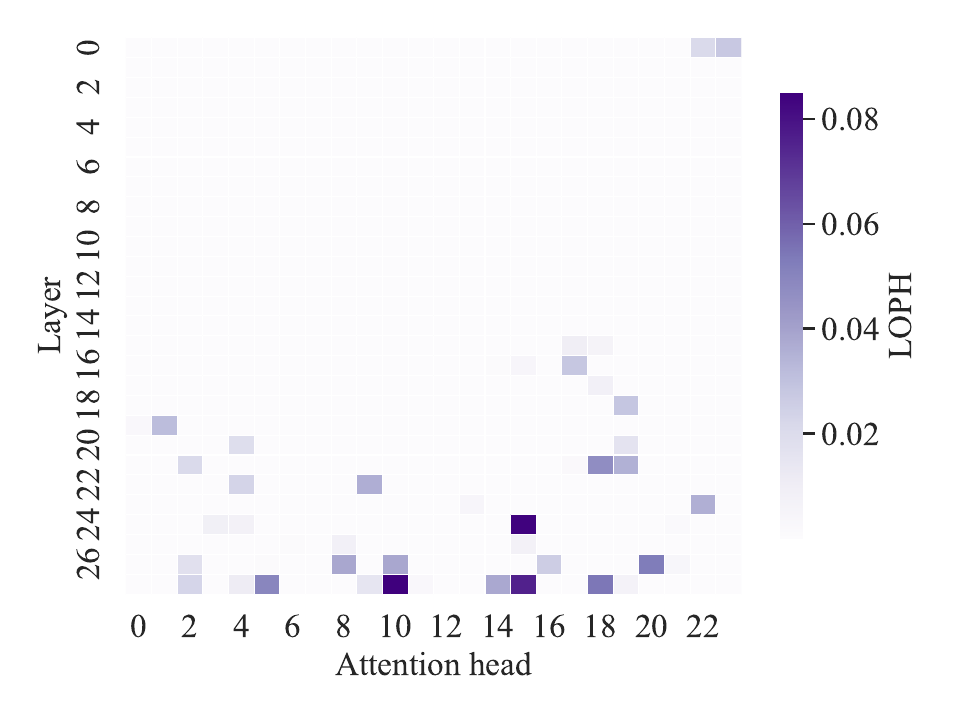}}
  \caption{Average LOPH of the unlearned LLaMA3.2-3B-Instruct models.}
  \label{fig:su_heatmap_document}
\end{figure}

To further demonstrate the generalizability of our interpretability analysis framework, we extend our methodology to an additional task: \textbf{superficial unlearning}, a scenario in which the unlearned model fails to truly forget the target information. 
% Analogously to superficial editing, superficial unlearning refers to a scenario in which the unlearned model fails to truly forget the target information. 
Consequently, there exists a potential for this information to be reactivated (\citealp{su_eightmethods}; \citealp{su_yuan}; \citealp{su_steer}; \citealp{su_quanti}). 
In Appendix \ref{appen:su}, we provide a detailed description of the data construction and subsequent analysis procedures. 
The results presented in Figure \ref{fig:su_heatmap_document} and Table \ref{tab:unlearning_abl_svd_document} demonstrate that, in the context of superficial unlearning, certain attention heads remain active, and their singular vectors are associated with superficial unlearning, supporting the generalizability of our method and conclusions.
% The results shown in Figure \ref{fig:su_heatmap_document} and Table \ref{tab:unlearning_abl_svd_document} further validate the generalizability of our method, demonstrating that the phenomenon of superficial unlearning is also associated with specific attention modules and their singular vectors.

\begin{table}
  \centering
  \begin{tabular}{lccc}
    \toprule
    \textbf{Setting} & \textbf{w/o abl.} & \textbf{-top 5\%} & \textbf{-top 10\%} \\
    \midrule
    \textbf{Probability} & 53.95 & 35.12 & 28.97 \\
    \bottomrule
  \end{tabular}
  \caption{Probabilities of $o$ under different settings. (w/o abl.: without ablation; -top 5\%: ablation of top 5\% vectors; -top 10\%: ablation of top 10\% vectors)}
  \label{tab:unlearning_abl_svd_document}
\end{table}

% \begin{figure}[t]
% \centering
%   \scalebox{0.9}{\includegraphics[width=\columnwidth]{figures/unlearning/abl_svd.pdf}}
%   \caption{Output probabilities of $o$ under different ablation settings.}
%   \label{fig:abl_svd_document}
% \end{figure}
% \begin{figure}[t]
%   \centering
%   \begin{minipage}{0.48\linewidth}
%       \centering
%       \includegraphics[width=\columnwidth]{figures/unlearning/heatmap.pdf}
%       \caption{Average LOPH of the unlearned models.}
%       \label{fig:unlearn_heatmap_document}
%   \end{minipage}
%   \hfill
%   \begin{minipage}{0.48\linewidth}
%       \centering
%       \includegraphics[width=\columnwidth]{figures/unlearning/abl_svd2.pdf}
%       \caption{Output probabilities of $o$ under different settings of left singular vectors.}
%   \end{minipage}
% \end{figure}

\section{Related Work}\label{sec:related_work}

Knowledge editing aims to modify specific factual knowledge in LLMs while ensuring that unrelated knowledge remains unaffected. Existing research on knowledge editing encompasses a diverse range of methodologies (\citealp{zhu2020ft}; \citealp{decao-ke}; \citealp{mend}; \citealp{serac}; \citealp{rome}; \citealp{memit}; \citealp{ike_icl}), paradigms (\citealp{grace}; \citealp{alphaedit}; \citealp{crosslingual0}; \citealp{crosslingualwang}; \citealp{mulfe}; \citealp{akew}), evaluation approaches (\citealp{mquake}; \citealp{rippleeffct}; \citealp{longformeval}; \citealp{yang-etal-2024-butterfly}; \citealp{ma-robustness}), and applications (\citealp{detoxify_nyz}; \citealp{editingdpo}; \citealp{}\citealp{editfairness}). 
Despite these successful efforts, the challenge of superficial editing remains underexplored. 
In this study, we conduct a systematic investigation of this issue.

% \paragraph{Logit Lens.} A detailed explanation is provided in Appendix \ref{sec:appen_related_work}.
% The logit lens (\citealp{logitlens0}; \citealp{geva-logitlens}; \citealp{dar-logitlens}; \citealp{halawi-logitlens}) technique has emerged as a powerful tool for understanding the internal mechanisms of language models. It leverages the observation that the hidden states at each layer of a Transformer, when appropriately decoded, gradually converge towards the final output distribution. The core idea is to project an internal representation into the vocabulary space:
% \begin{equation}
%     P_{LL}\left(t\mid \boldsymbol{x} \right) = \text{softmax}\left( \boldsymbol{W}_U \boldsymbol{x} \right),
% \end{equation}
% where $t$ is the next token, $\boldsymbol{x}$ is an internal representation, $\boldsymbol{W}_U$ is the unembedding matrix, $P_{LL}\left(t\mid\boldsymbol{x}\right)$ denotes the probability of obtaining $t$ after decoding $\boldsymbol{x}$. In this study, we refer to $P_{LL}$ as \textbf{latent probability}.

\section{Conclusion}
In this study, we formally define superficial editing and conduct a comprehensive evaluation, demonstrating that superficial editing constitutes a critical challenge. 
Our rigorous analysis identifies and validates two key factors for this issue: the residual stream in earlier layers and the attention in later layers. We investigate the internal mechanisms of the attention module and reveal that specific attention heads and their corresponding left singular vectors are responsible for superficial editing. 
Furthermore, we validate the generalizability of our analytical framework by applying it to superficial unlearning, where we observe consistent mechanisms, thereby demonstrating the robustness and broader applicability of both our methodology and conclusions.

\section*{Limitations}
We outline the limitations of our work as follows:
(1) Our investigation is limited to examining superficial editing within three specific attack contexts, which may not encompass all possible scenarios. While an exhaustive evaluation of every context is computationally infeasible, developing more comprehensive and systematic evaluation methodologies remains an important direction for future research. 
(2) The development of effective mitigation strategies for superficial editing remains an open challenge. We identify this as a crucial area for future investigation. 

\bibliography{custom}

\clearpage
\appendix

\section{Attack Probes Generation}
\label{sec:appen_attack_gen}
The data generation procedure involves the following steps:
\textbf{(1)} For the unedited model, we traverse the dataset and identify instances where the model's prediction perfectly matches the corresponding ground truth answer through greedy decoding. These instances are considered to represent knowledge that has been effectively acquired and internalized by the model.
\textbf{(2)} Based on the three distinct attack types we defined in Section \ref{sec:formulation}, we generate attack probes for each sample and apply ROME (\citealp{rome}) to edit the model parameters. Subsequently, we evaluate the edited model by exposing it to the generated probes. Samples that elicit the original answers from the model are retained for further analysis. 
\textbf{(3)} To enhance data diversity, we use two additional prominent algorithms, MEMIT (\citealp{memit}) and MEND (\citealp{mend}), to replicate the procedure outlined in step (2). We then combine the datasets obtained from all three methods to create the final datasets, \textbf{CF-a} and \textbf{ZsRE-a}. A summary of the data statistics is provided in Table \ref{tab:data_statistics}. 

\begin{figure}[t]
\centering
  \includegraphics[width=\columnwidth]{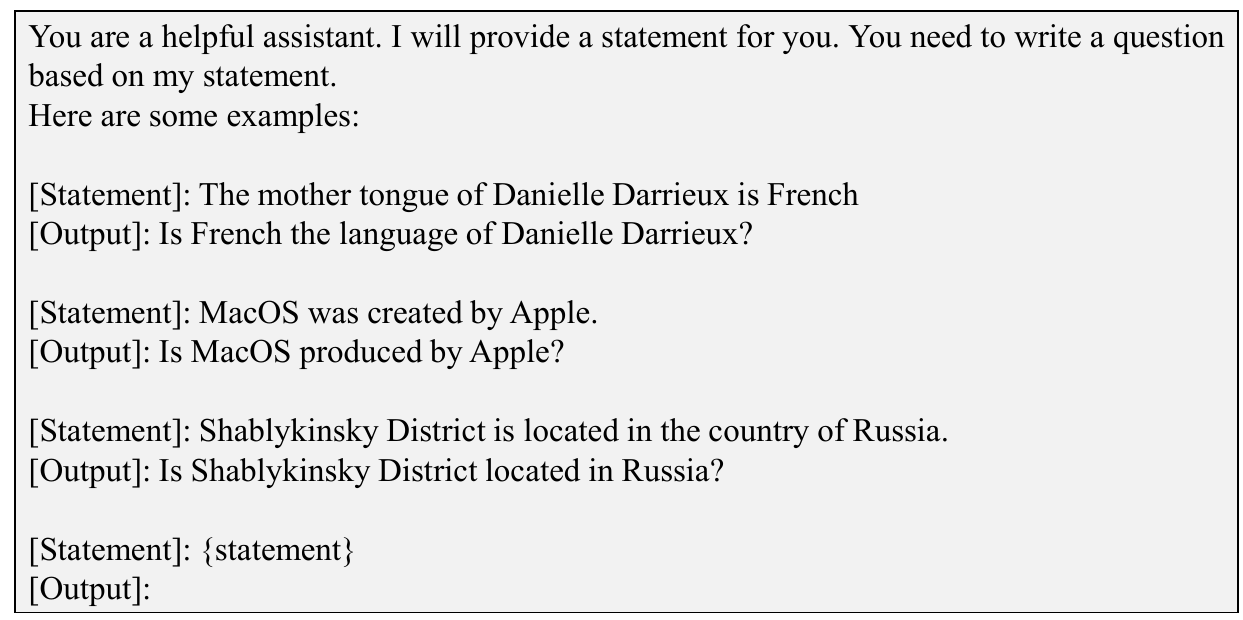}
  \caption{The prompt for Qwen2.5-32B-Instruct to generate the third type of attack prefix. }
  \label{fig:qwen32b_prompt_cf}
\end{figure}

The methodology for constructing the three distinct types of attack prefixes in step \textbf{(2)} is as follows: \textbf{(a)} Wikipedia context. We extract a concise summary of the original answer $o$ using the Wikipedia library\footnote{\url{https://github.com/goldsmith/Wikipedia}}, limiting the maximum number of sentences to 3. \textbf{(b)} Original entity repetition. In this instance, the original answer $o$ is repeated $m$ times as an attack prefix. \textbf{(c)} A question about the original triple. As the initial dataset lacks complete questions for each triple (e.g., ``Is Joe Biden the President of the U.S.?''), we employ a large language model, specifically Qwen2.5-32B-Instruct\footnote{\url{https://qwenlm.github.io/blog/qwen2.5-llm/}}, to generate corresponding questions. The specific prompt utilized in this process is illustrated in Figure \ref{fig:qwen32b_prompt_cf}. After obtaining three types of attack prefixes, we concatenate them with the baseline prompts in the original dataset.

\begin{figure}[t]
\centering
  \includegraphics[width=\columnwidth]{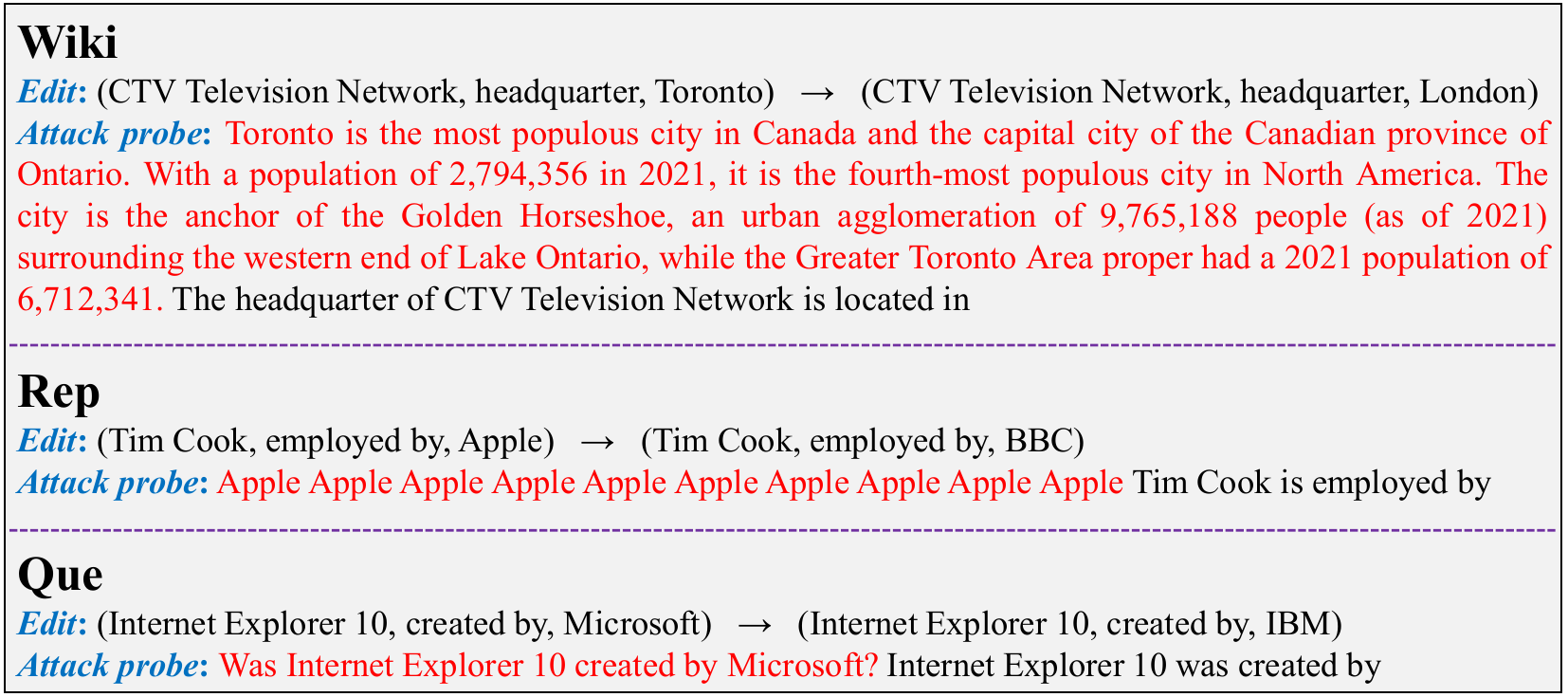}
  \caption{Examples for three attack types. Attack prefixes are highlighted in red.}
  \label{fig:appen_attack_example}
\end{figure}

We provide examples of three attack types in Figure \ref{fig:appen_attack_example}.

\begin{table}[h!]
  \centering
  \begin{tabular}{lcccc}
    \toprule
    \textbf{Dataset} & \textbf{Wiki} & \textbf{Rep} & \textbf{Que} & \textbf{Total} \\
    \midrule
    CF-a & 323 & 484 & 204 & 1011 \\
    ZsRE-a & 133 & 214 & 122 & 469 \\
    \bottomrule
  \end{tabular}
  \caption{Statistics of our evaluation dataset.}
  \label{tab:data_statistics}
\end{table}

\section{Evaluation of Superficial Editing}
\label{sec:appen_eval}
\textbf{Efficacy} (Eff.) is measured as the proportion of cases where $o$ is more probable than $o^*$ with the edit prompt:

{\footnotesize\begin{equation}
    \text{Eff.} = \mathbb{E}_{x_i}\left[ P_{f^\prime}\left( o\mid x_i \right) > P_{f^\prime} \left( o^* \mid x_i \right) \right]
\end{equation}}

\textbf{Generalization} (Gen.) represents the proportion of paraphrased prompts $\mathcal{N}$ where $o$ is more probable than $o^*$:

{\footnotesize\begin{equation}
    \text{Gen.} = \mathbb{E}_{x_i\in\mathcal{N}}\left[ P_{f^\prime}\left( o\mid x_i \right) > P_{f^\prime} \left( o^* \mid x_i \right) \right]
\end{equation}}

\textbf{Locality} (Loc.) is the proportion of neighborhood prompts $\mathcal{O}$ where the edited model assigns a higher probability to the original answer:

{\footnotesize\begin{equation}
    \text{Eff.} = \mathbb{E}_{x_i\in\mathcal{O}}\left[ P_{f^\prime}\left( o^*\mid x_i \right) > P_{f^\prime} \left( o \mid x_i \right) \right]
\end{equation}}

We present the evaluation results for various experimental configurations in Tables \ref{tab:llama3_eval_zsre} to \ref{tab:qwen14b_eval_zsre}.

\begin{table*}
  \centering
  \scalebox{0.73}{\begin{tabular}{l|ccccc|ccccc|ccccc}
    \toprule
    \multirow{2}{*}{\textbf{Methods}}           & \multicolumn{5}{c|}{\textbf{Wiki}} & \multicolumn{5}{c|}{\textbf{Rep}} & \multicolumn{5}{c}{\textbf{Que}} \\
     & Eff. & Gen. & Loc. & OM $\downarrow$ & OP $\downarrow$ & Eff. & Gen. & Loc. & OM $\downarrow$ & OP $\downarrow$ & Eff. & Gen. & Loc. & OM $\downarrow$ & OP $\downarrow$ \\
    \midrule
    FT & 73.68 & 69.23 & 37.43 & 67.61 & 84.51      & 73.57 & 71.30 & 43.42 & 89.16 & 94.58     & 75.50 & 75.50 & 47.98 & 70.27 & 90.54    \\
    MEND & 97.75 & 97.75 & 37.43 & 47.89 & 47.89     & 100 & 100 & 43.09 & 59.64 & 60.24     & 98.76 & 98.40 & 47.87 & 33.78 & 33.78   \\
    ROME & 98.39 & 90.61 & 37.43 & 33.80 & 36.62      & 100 & 91.92 & 43.42 & 33.13 & 36.14    & 98.87 & 96.03 & 49.04 & 52.70 & 55.41    \\
    MEMIT & 98.39 & 96.16 & 37.59 & 43.66 & 49.30      & 99.41 & 94.87 & 43.42 & 46.39 & 50.60   
 & 98.87 & 93.19 & 48.33 & 28.38 & 33.78  \\
    PMET & 98.39 & 84.23 & 37.43 & 38.03 & 39.44     & 98.74 & 76.77 & 43.42 & 47.59 & 50.60    & 98.87 & 80.60 & 47.98 & 41.89 & 56.76    \\
    r-ROME &  98.39 & 91.16 & 37.43 & 38.03 & 39.44     & 100 & 91.25 & 43.42 & 34.34 & 36.75    & 98.87 & 94.08 & 49.04 & 55.41 & 59.46  \\
    AlphaEdit & 98.39 & 77.94 & 37.59 & 43.66 & 50.70    & 100 & 73.06 & 43.42 & 61.45 & 66.27    & 98.87 & 86.70 & 48.13 & 45.95 & 59.46   \\
    \bottomrule
  \end{tabular}}
  \caption{Evaluation results of superficial editing conducted on LLaMA3-8B-Instruct using the ZsRE-a dataset. 
  }
  \label{tab:llama3_eval_zsre}
\end{table*}

\begin{table*}
  \centering
  \scalebox{0.73}{\begin{tabular}{l|ccccc|ccccc|ccccc}
    \toprule
    \multirow{2}{*}{\textbf{Methods}}           & \multicolumn{5}{c|}{\textbf{Wiki}} & \multicolumn{5}{c|}{\textbf{Rep}} & \multicolumn{5}{c}{\textbf{Que}} \\
     & Eff. & Gen. & Loc. & OM $\downarrow$ & OP $\downarrow$ & Eff. & Gen. & Loc. & OM $\downarrow$ & OP $\downarrow$ & Eff. & Gen. & Loc. & OM $\downarrow$ & OP $\downarrow$ \\
    \midrule
    FT & 92.31 & 73.85 & 26.00 & 53.57 & 57.14      & 92.62 & 67.11 & 30.67 & 43.23 & 49.48       & 95.45 & 82.58 & 14.39 & 65.43 & 70.27  \\
    MEND & 100 & 58.46 & 51.54 & 41.67 & 46.43     & 100 & 52.35 & 51.81 & 42.71 & 52.08     & 100 & 79.55 & 34.39 & 40.74 & 43.21   \\
    ROME & 98.46 & 93.08 & 89.69 & 55.95 & 60.71     & 99.33 & 93.62 & 91.48 & 52.60 & 59.38     & 100 & 97.73 & 82.27 & 64.20 & 65.43  \\
    MEMIT & 98.46 & 95.38 & 90.92 & 34.52 & 34.52    & 100 & 92.95 & 90.47 & 30.73 & 35.94    & 100 & 99.24 & 82.73 & 37.04 & 41.98  \\
    PMET & 95.38 & 89.23 & 92.92 & 47.62 & 52.38     & 100 & 88.59 & 92.15 & 47.40 & 56.77   & 100 & 87.12 & 83.79 & 43.21 & 48.15  \\
    r-ROME & 98.46 & 89.23 & 91.69 & 57.14 & 61.90     & 99.33 & 90.60 & 92.01 & 52.60 & 59.90    & 100 & 96.21 & 82.27 & 56.79 & 59.26  \\
    AlphaEdit & 98.46 & 93.08 & 92.31 & 53.57 & 60.71   & 100 & 85.57 & 92.28 & 47.92 & 60.94    & 100 & 91.67 & 84.85 & 40.74 & 45.68  \\
    \bottomrule
  \end{tabular}}
  \caption{Evaluation results of superficial editing conducted on Qwen2.5-7B-Instruct using the CF-a dataset. 
  }
  \label{tab:qwen7b_eval_cf}
\end{table*}

\begin{table*}
  \centering
  \scalebox{0.73}{\begin{tabular}{l|ccccc|ccccc|ccccc}
    \toprule
    \multirow{2}{*}{\textbf{Methods}}           & \multicolumn{5}{c|}{\textbf{Wiki}} & \multicolumn{5}{c|}{\textbf{Rep}} & \multicolumn{5}{c}{\textbf{Que}} \\
     & Eff. & Gen. & Loc. & OM $\downarrow$ & OP $\downarrow$ & Eff. & Gen. & Loc. & OM $\downarrow$ & OP $\downarrow$ & Eff. & Gen. & Loc. & OM $\downarrow$ & OP $\downarrow$ \\
    \midrule
    FT & 76.04 & 70.58 & 32.38 & 49.12 & 80.70     & 68.80 & 66.18 & 38.56 & 46.98 & 87.25     & 77.79 & 71.77 & 38.35 & 37.36 & 93.41    \\
    MEND & 98.67 & 98.40 & 31.75 & 47.37 & 54.39    & 98.79 & 98.79 & 37.83 & 39.60 & 59.06     & 99.69 & 99.69 & 38.00 & 30.77 & 56.04 \\
    ROME & 99.67 & 88.89 & 32.10 & 39.47 & 65.79     & 99.49 & 83.59 & 38.76 & 38.93 & 55.03     & 100 & 91.41 & 38.48 & 48.35 & 57.14     \\
    MEMIT & 99.67 & 89.44 & 31.92 & 52.63 & 70.18     & 99.49 & 88.05 & 38.76 & 42.95 & 55.03     & 100 & 80.47 & 38.48 & 31.87 & 46.15   \\
    PMET & 96.44 & 80.22 & 32.10 & 36.84 & 65.79     & 97.81 & 70.96 & 38.18 & 31.54 & 65.10    & 99.22 & 77.34 & 38.48 & 30.77 & 53.85     \\
    r-ROME & 99.67 & 87.22 & 32.10 & 36.84 & 64.91 & 99.49 & 82.74 & 38.76 & 37.58 & 54.36     & 100 & 91.41 & 38.48 & 47.25 & 58.24   \\
    AlphaEdit & 99.67 & 78.29 & 31.91 & 40.35 & 66.67    & 99.24 & 77.66 & 38.20 & 27.52 & 65.77    & 100 & 78.59 & 38.48 & 23.08 & 58.24   \\
    \bottomrule
  \end{tabular}}
  \caption{Evaluation results of superficial editing conducted on Qwen2.5-7B-Instruct using the ZsRE-a dataset. 
  }
  \label{tab:qwen7b_eval_zsre}
\end{table*}

\begin{table*}
  \centering
  \scalebox{0.73}{\begin{tabular}{l|ccccc|ccccc|ccccc}
    \toprule
    \multirow{2}{*}{\textbf{Methods}}           & \multicolumn{5}{c|}{\textbf{Wiki}} & \multicolumn{5}{c|}{\textbf{Rep}} & \multicolumn{5}{c}{\textbf{Que}} \\
     & Eff. & Gen. & Loc. & OM $\downarrow$ & OP $\downarrow$ & Eff. & Gen. & Loc. & OM $\downarrow$ & OP $\downarrow$ & Eff. & Gen. & Loc. & OM $\downarrow$ & OP $\downarrow$ \\
    \midrule
    FT & 94.47 & 73.87 & 28.14 & 45.24 & 54.37    & 96.24 & 74.06 & 29.77 & 40.38 & 46.79      & 96.15 & 75.64 & 23.33 & 57.61 & 68.48   \\
    ROME & 99.50 & 97.74 & 90.95 & 59.52 & 60.32     & 100 & 89.10 & 88.65 & 48.72 & 50.00     & 100 & 98.72 & 88.85 & 68.48 & 70.65    \\
    MEMIT & 99.50 & 94.97 & 92.06 & 74.21 & 78.57    & 98.50 & 83.83 & 90.15 & 79.49 & 80.77     & 100 & 98.08 & 88.84 & 66.30 & 70.65   \\
    PMET & 98.49 & 89.70 & 92.36 & 75.79 & 84.13      & 96.24 & 71.80 & 91.58 & 73.71 & 84.62      & 100 & 94.87 & 89.10 & 65.22 & 70.65     \\
    r-ROME & 100 & 97.99 & 91.56 & 54.76 & 56.75    & 100 & 89.85 & 89.92 & 44.23 & 48.08       & 100 & 98.72 & 88.85 & 65.22 & 67.39   \\
    AlphaEdit & 100 & 91.21 & 92.36 & 72.62 & 79.76    & 99.25 & 73.68 & 91.65 & 67.95 & 76.28      & 100 & 94.23 & 88.85 & 55.43 & 59.78   \\
    \bottomrule
  \end{tabular}}
  \caption{Evaluation results of superficial editing conducted on Qwen2.5-14B-Instruct using the CF-a dataset. 
  }
  \label{tab:qwen14b_eval_cf}
\end{table*}

\begin{table*}
  \centering
  \scalebox{0.73}{\begin{tabular}{l|ccccc|ccccc|ccccc}
    \toprule
    \multirow{2}{*}{\textbf{Methods}}           & \multicolumn{5}{c|}{\textbf{Wiki}} & \multicolumn{5}{c|}{\textbf{Rep}} & \multicolumn{5}{c}{\textbf{Que}} \\
     & Eff. & Gen. & Loc. & OM $\downarrow$ & OP $\downarrow$ & Eff. & Gen. & Loc. & OM $\downarrow$ & OP $\downarrow$ & Eff. & Gen. & Loc. & OM $\downarrow$ & OP $\downarrow$ \\
    \midrule
    FT & 68.85 & 73.33 & 25.53 & 36.36 & 77.27    & 83.33 & 70.83 & 47.70 & 23.81 & 57.14     & 86.36 & 90.91 & 39.99 & 26.67 & 66.67   \\
    ROME & 100 & 98.08 & 26.63 & 40.91 & 72.73      & 100 & 97.92 & 49.78 & 61.90 & 76.19      & 100 & 95.45 & 40.19 & 26.67 & 40.00    \\
    MEMIT & 100 & 95.51 & 26.63 & 31.82 & 95.45    & 100 & 97.92 & 49.86 & 66.67 & 71.43    & 100 & 90.91 & 40.19 & 46.67 & 46.67      \\
    PMET & 96.15 & 91.67 & 26.63 & 27.27 & 77.27        & 97.92 & 90.63 & 49.86 & 19.05 & 80.95      & 90.91 & 100 & 40.19 & 26.67 & 40.00      \\
    r-ROME & 100 & 98.08 & 26.63 & 36.36 & 72.73      & 100 & 97.92 & 49.78 & 57.14 & 71.43      & 100 & 95.45 & 40.19 & 26.67 & 40.00    \\
    AlphaEdit & 97.44 & 92.95 & 26.63 & 27.27 & 90.91     & 100 & 90.63 & 49.86 & 23.81 & 66.67     & 100 & 90.91 & 40.19 & 33.33 & 60.00  \\
    \bottomrule
  \end{tabular}}
  \caption{Evaluation results of superficial editing conducted on Qwen2.5-14B-Instruct using the ZsRE-a dataset. 
  }
  \label{tab:qwen14b_eval_zsre}
\end{table*}

\clearpage
\section{Mechanistic Analysis of Superficial Editing}
\label{sec:appen_analysis}

\subsection{Effects of Transformer Components}
\label{subsec:appen_3_components}
\begin{figure*}[t]
  \centering
  \begin{subfigure}[t]{0.48\textwidth}
      \centering
      \includegraphics[width=\linewidth]{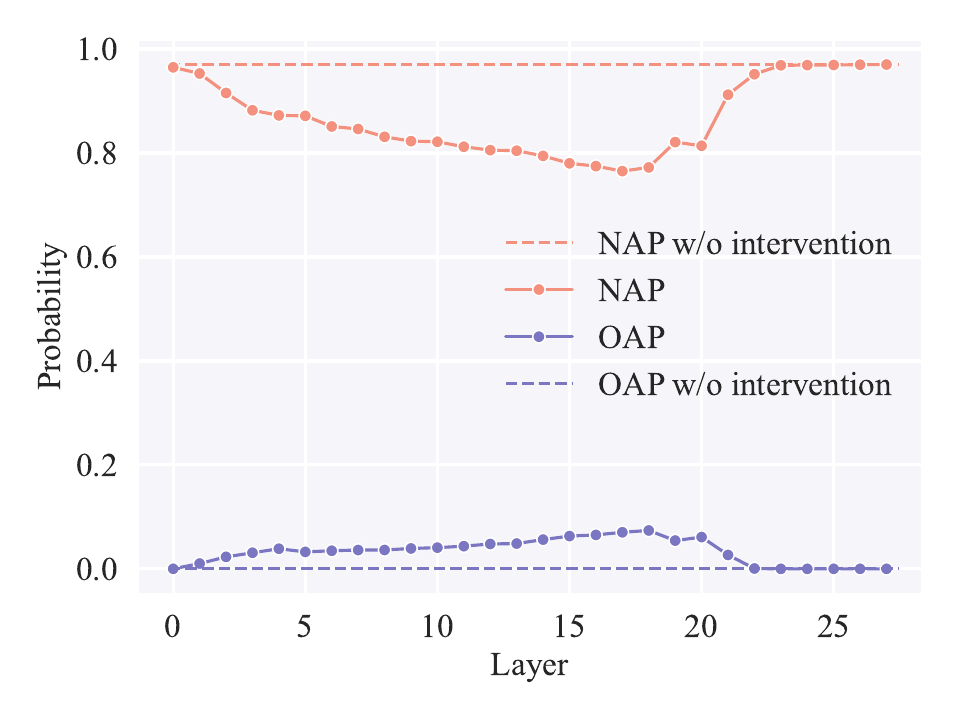}
      \caption{Intervention on the last subject token.}
      \label{subfig:subjct_last_7b_rome}
  \end{subfigure}
  \hfill
  \begin{subfigure}[t]{0.48\textwidth}
      \centering
      \includegraphics[width=\linewidth]{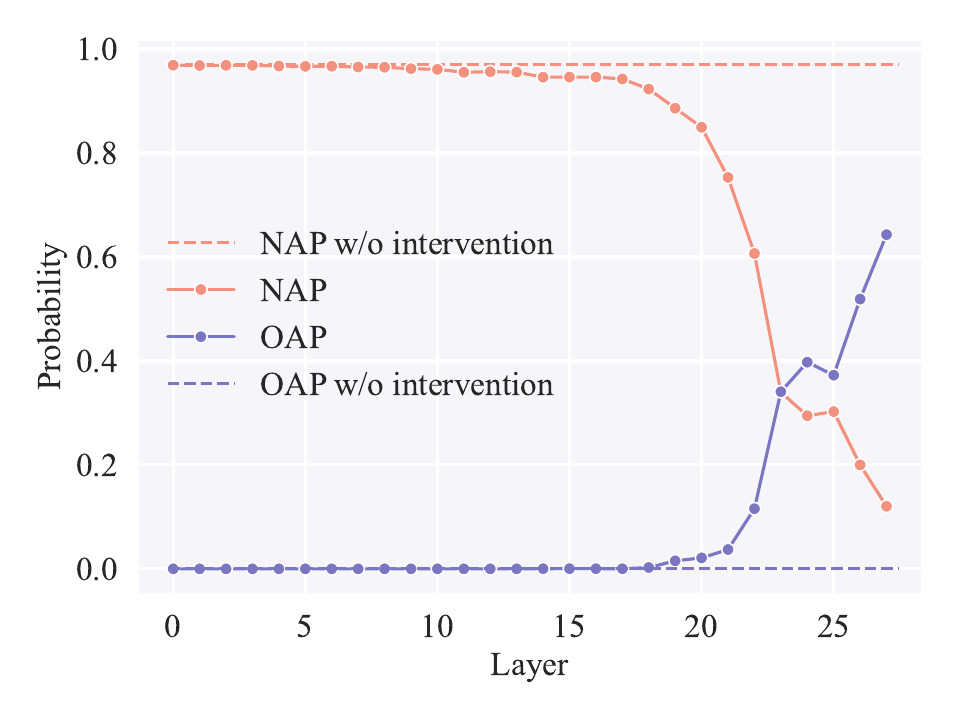}
      \caption{Intervention on the last token.}
      \label{subfig:last_7b_rome}
  \end{subfigure}
  \caption {Intervention results of Qwen2.5-7B-Instruct edited by ROME at different tokens.}
  \label{fig:resid_patch_qwen7b_rome}
\end{figure*}

\begin{figure*}[t]
  \centering
  \begin{subfigure}[t]{0.48\textwidth}
      \centering
      \includegraphics[width=\linewidth]{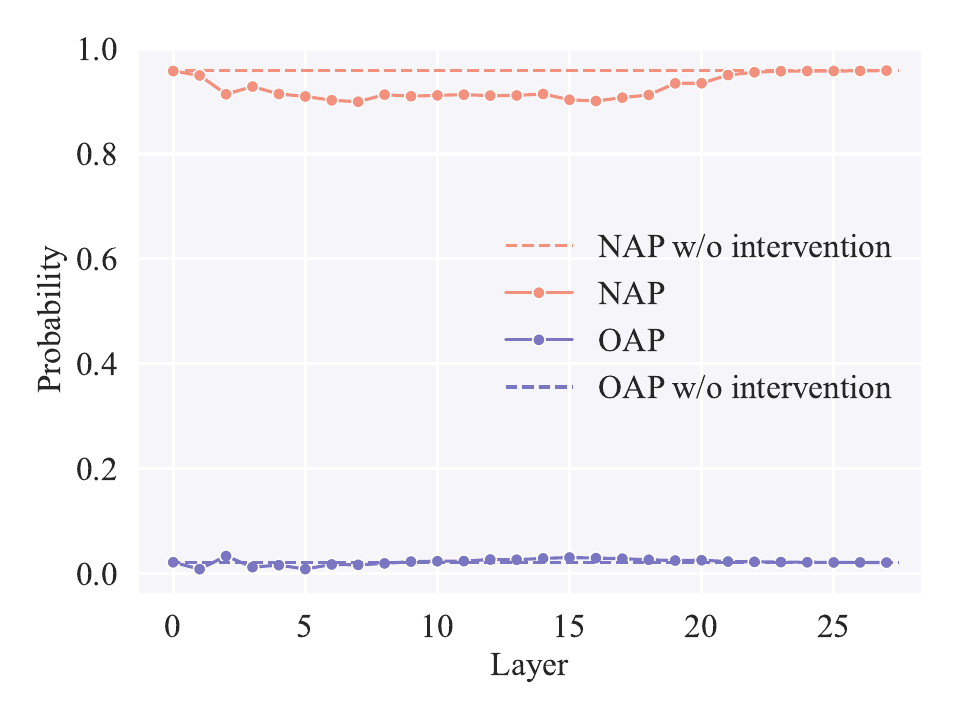}
      \caption{Intervention on the last subject token.}
      \label{subfig:subjct_last_7b_memit}
  \end{subfigure}
  \hfill
  \begin{subfigure}[t]{0.48\textwidth}
      \centering
      \includegraphics[width=\linewidth]{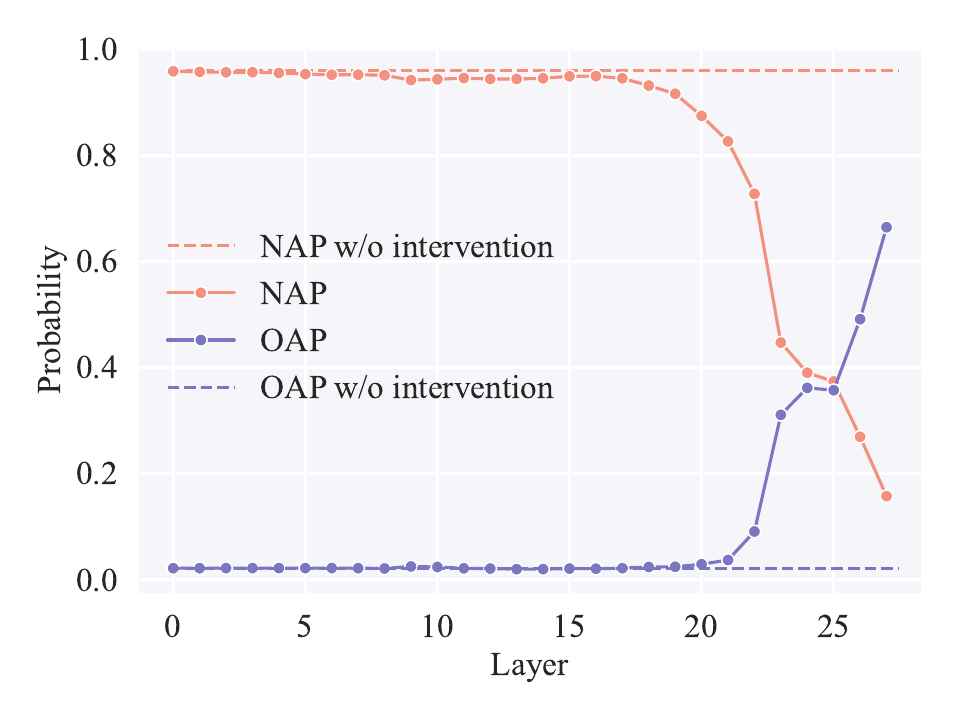}
      \caption{Intervention on the last token.}
      \label{subfig:last_7b_memit}
  \end{subfigure}
  \caption {Intervention results of Qwen2.5-7B-Instruct edited by MEMIT at different tokens.}
  \label{fig:resid_patch_qwen7b_memit}
\end{figure*}

\begin{figure*}[t]
  \centering
  \begin{subfigure}[t]{0.48\textwidth}
      \centering
      \includegraphics[width=\linewidth]{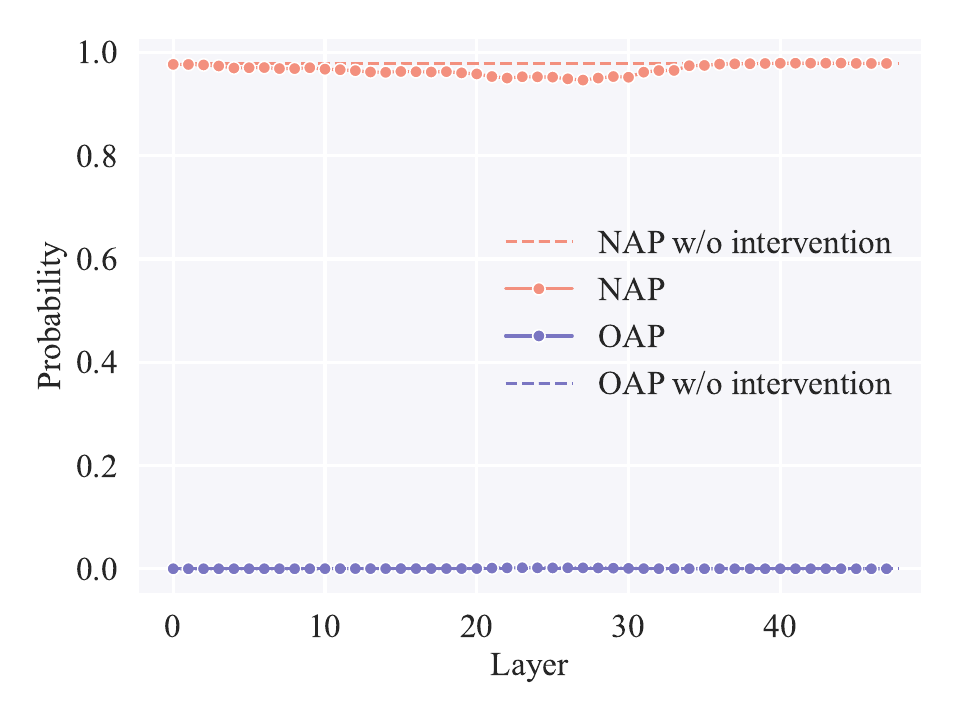}
      \caption{Intervention on the last subject token.}
      \label{subfig:subjct_last_14b_rome}
  \end{subfigure}
  \hfill
  \begin{subfigure}[t]{0.48\textwidth}
      \centering
      \includegraphics[width=\linewidth]{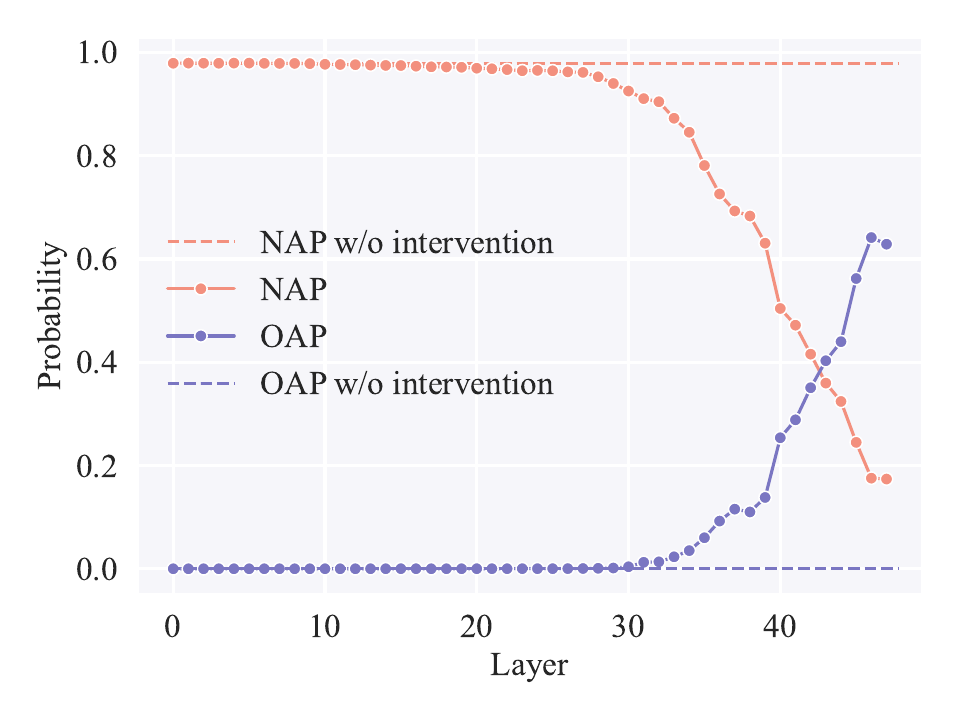}
      \caption{Intervention on the last token.}
      \label{subfig:last_14b_rome}
  \end{subfigure}
  \caption {Intervention results of Qwen2.5-14B-Instruct edited by ROME at different tokens.}
  \label{fig:resid_patch_qwen14b_rome}
\end{figure*}

\begin{figure*}[t]
  \centering
  \begin{subfigure}[t]{0.48\textwidth}
      \centering
      \includegraphics[width=\linewidth]{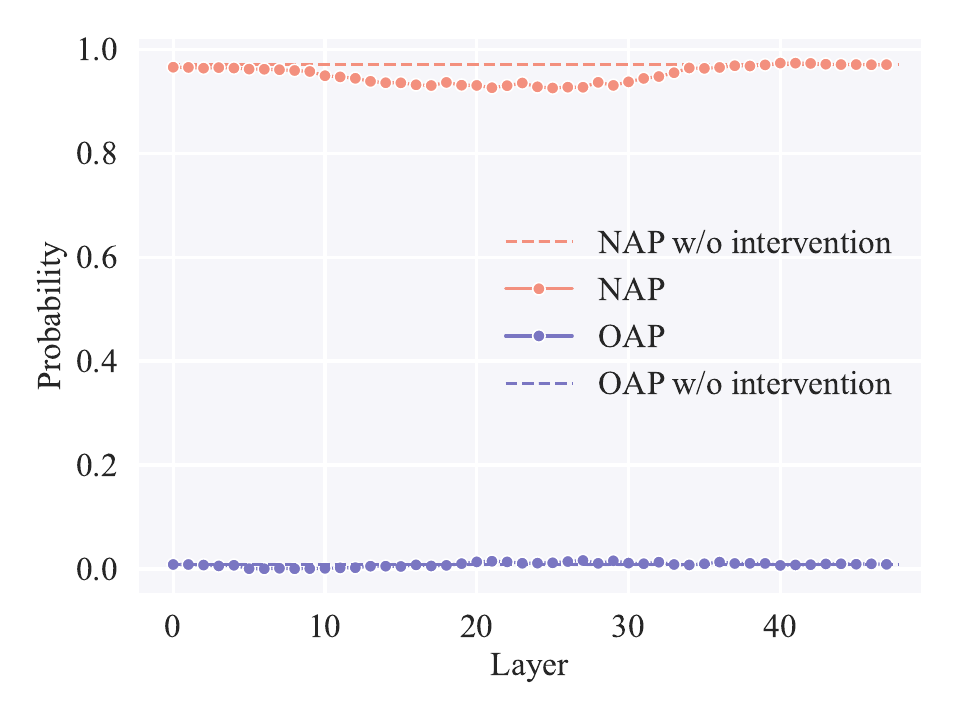}
      \caption{Intervention on the last subject token.}
      \label{subfig:subjct_last_14b_memit}
  \end{subfigure}
  \hfill
  \begin{subfigure}[t]{0.48\textwidth}
      \centering
      \includegraphics[width=\linewidth]{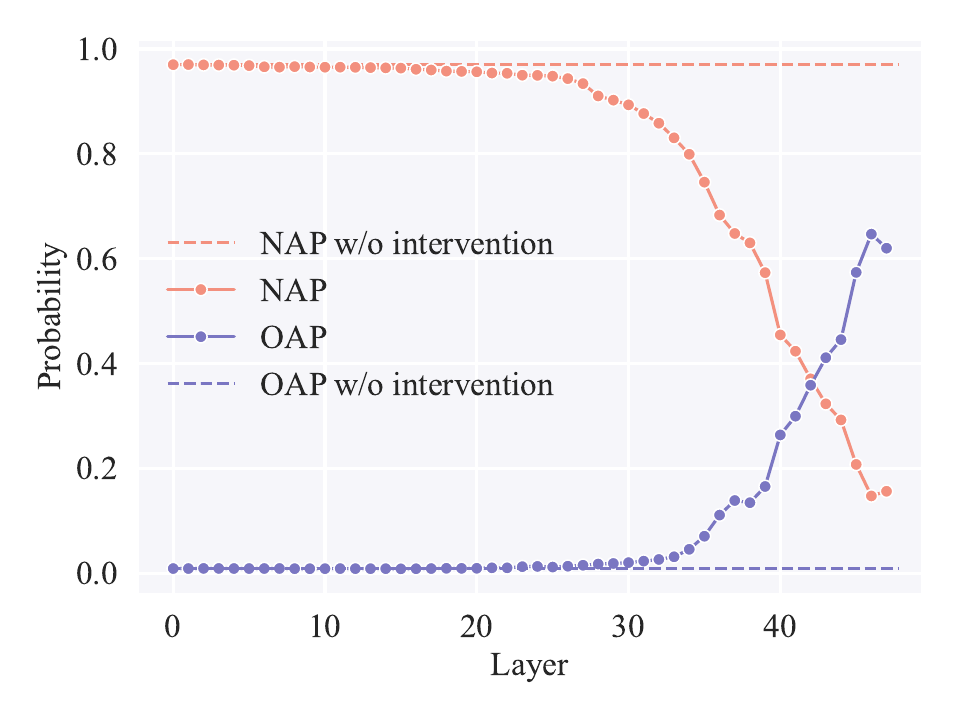}
      \caption{Intervention on the last token.}
      \label{subfig:last_14b_memit}
  \end{subfigure}
  \caption {Intervention results of Qwen2.5-14B-Instruct edited by MEMIT at different tokens.}
  \label{fig:resid_patch_qwen14b_memit}
\end{figure*}

The residual stream intervention results of other settings are illustrated in Figures \ref{fig:resid_patch_qwen7b_rome} to \ref{fig:resid_patch_qwen14b_memit}.

\begin{figure*}[t]
  \centering
  \begin{subfigure}[t]{0.48\textwidth}
      \centering
      \includegraphics[width=\linewidth]{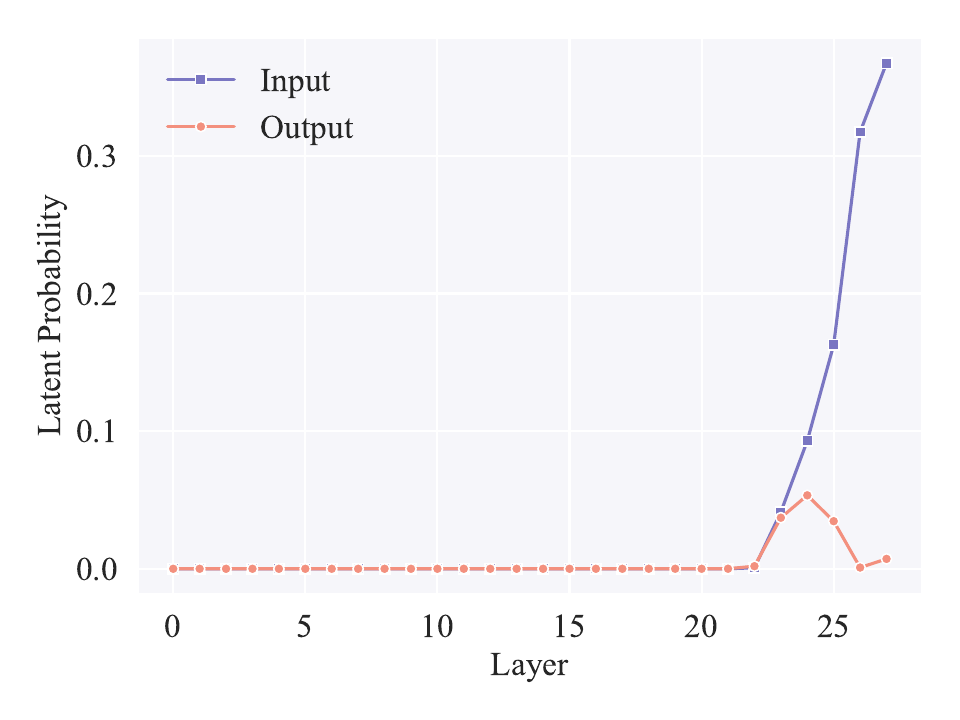}
      \caption{Results of MLP.}
      \label{subfig:qwen7b_mlp_rome}
  \end{subfigure}
  \hfill
  \begin{subfigure}[t]{0.48\textwidth}
      \centering
      \includegraphics[width=\linewidth]{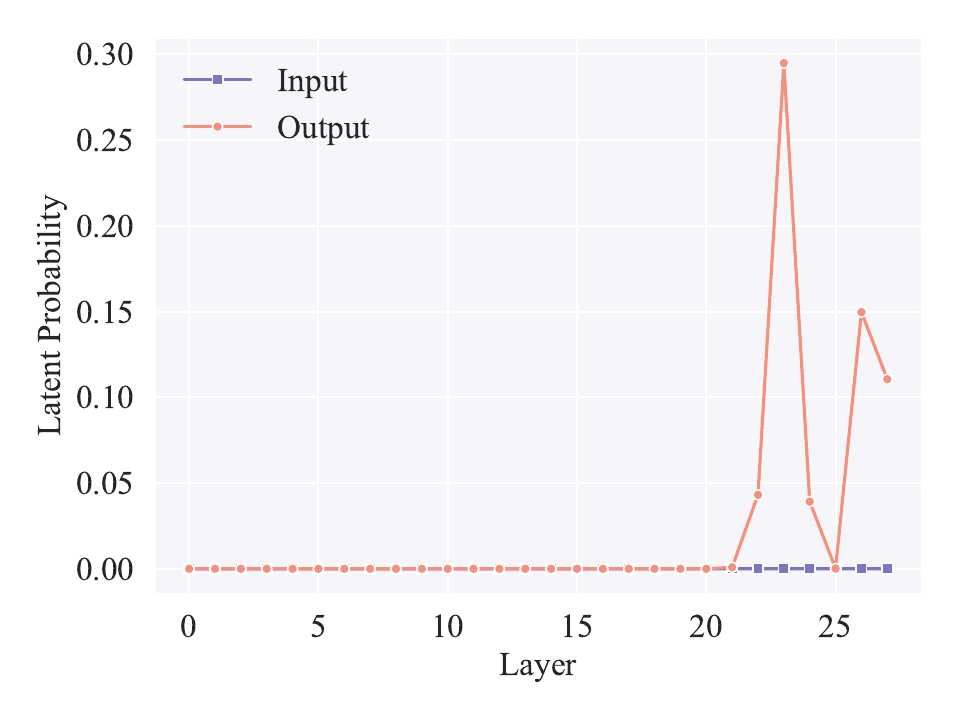}
      \caption{Results of Attention output matrix.}
      \label{subfig:qwen7b_attn_rome}
  \end{subfigure}
  \caption{The latent probabilities of $o$ for the input and output of MLP and Attention output matrix in Qwen2.5-7B-Instruct edited by ROME.}
  % \caption {Effects of MLP and Attention in Qwen2.5-7B-Instruct edited by ROME.}
  \label{fig:qwen7b_modio_rome}
\end{figure*}

\begin{figure*}[t]
  \centering
  \begin{subfigure}[t]{0.48\textwidth}
      \centering
      \includegraphics[width=\linewidth]{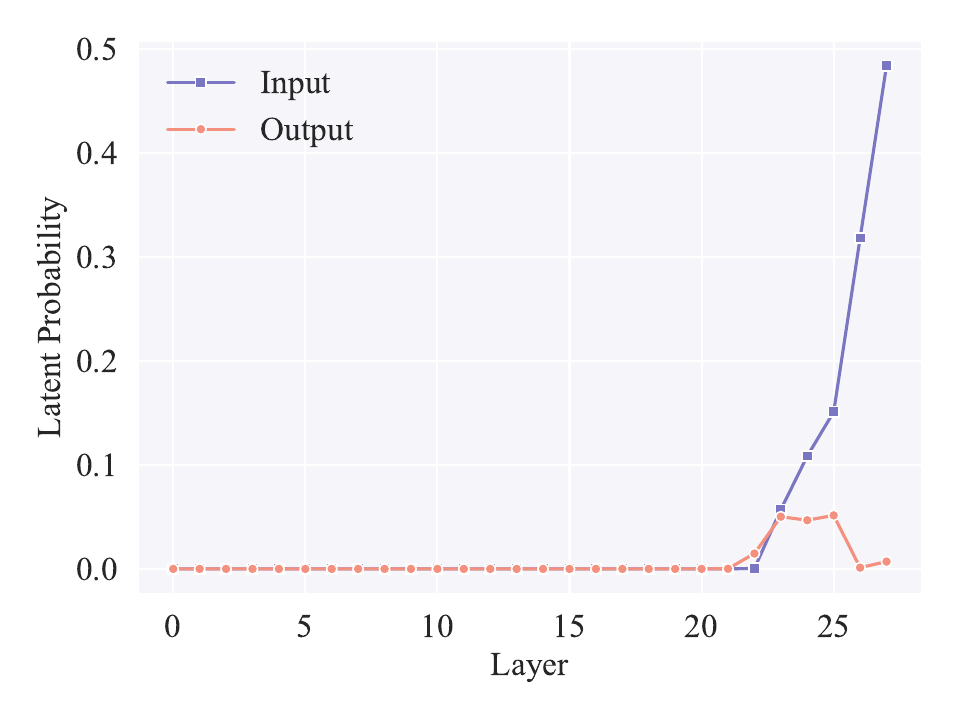}
      \caption{Results of MLP.}
      \label{subfig:qwen7b_mlp_memit}
  \end{subfigure}
  \hfill
  \begin{subfigure}[t]{0.48\textwidth}
      \centering
      \includegraphics[width=\linewidth]{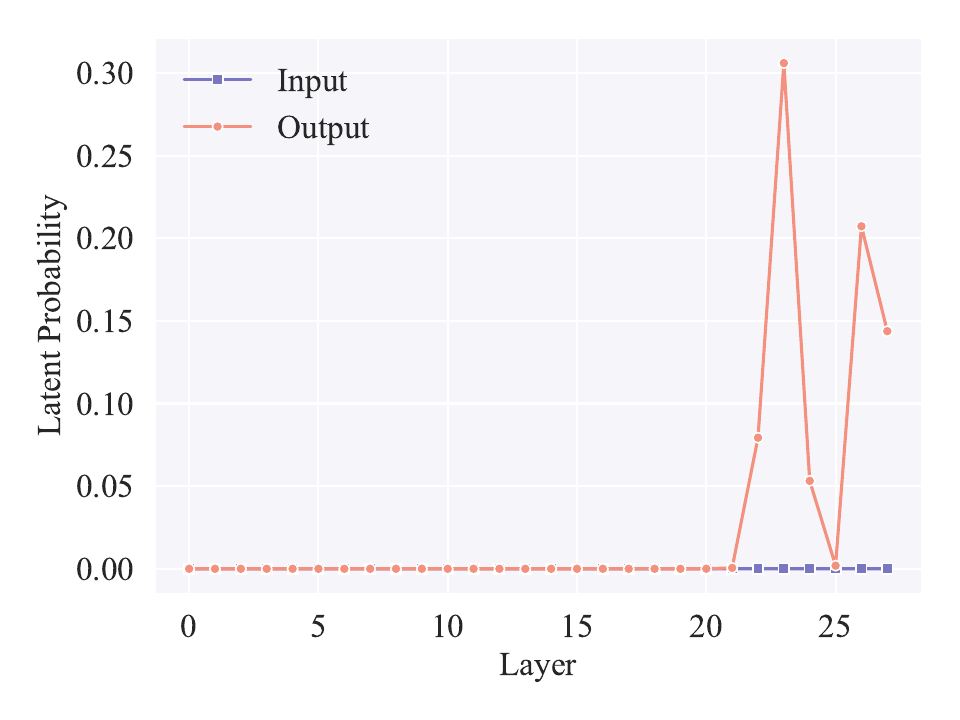}
      \caption{Results of Attention output matrix.}
      \label{subfig:qwen7b_attn_memit}
  \end{subfigure}
  \caption{The latent probabilities of $o$ for the input and output of MLP and Attention output matrix in Qwen2.5-7B-Instruct edited by MEMIT.}
  % \caption {Effects of MLP and Attention in Qwen2.5-7B-Instruct edited by MEMIT.}
  \label{fig:qwen7b_modio_memit}
\end{figure*}

\begin{figure*}[t]
  \centering
  \begin{subfigure}[t]{0.48\textwidth}
      \centering
      \includegraphics[width=\linewidth]{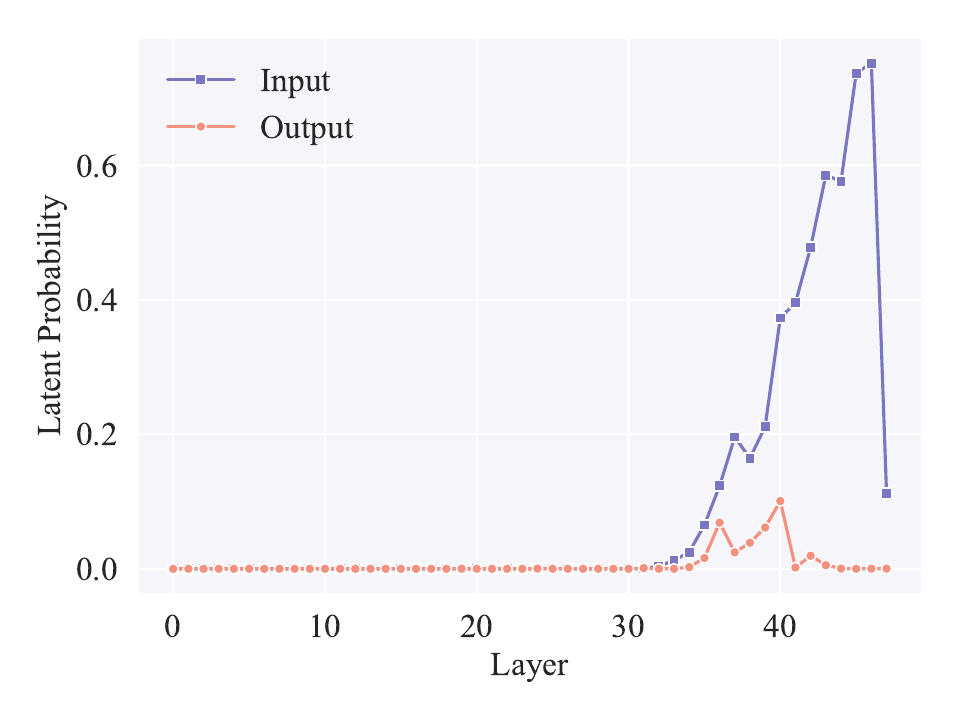}
      \caption{Results of MLP.}
      \label{subfig:qwen14b_mlp_rome}
  \end{subfigure}
  \hfill
  \begin{subfigure}[t]{0.48\textwidth}
      \centering
      \includegraphics[width=\linewidth]{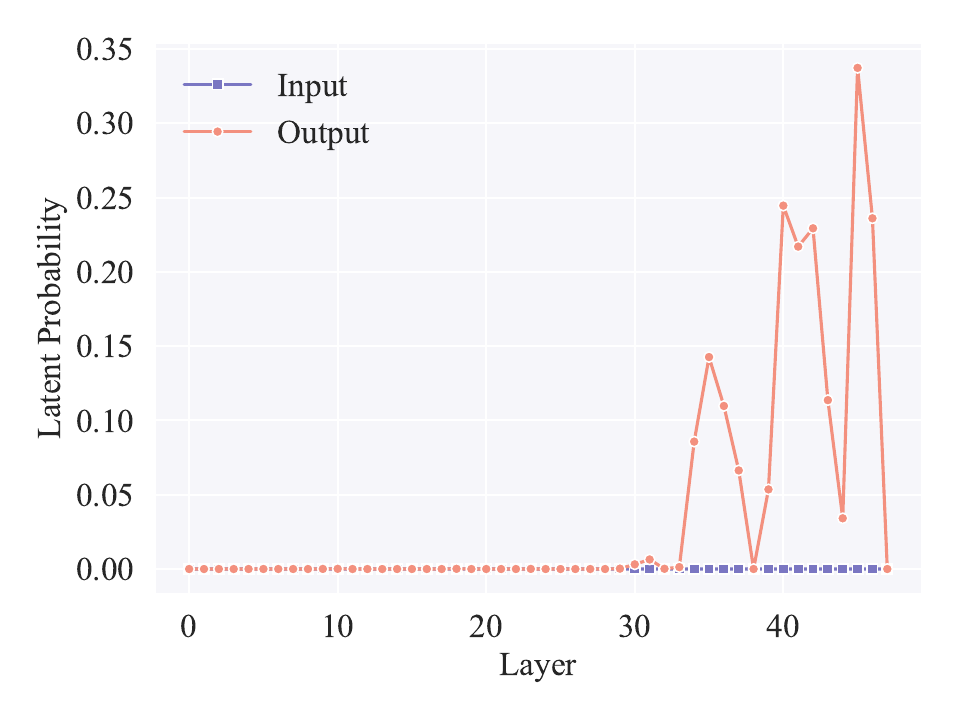}
      \caption{Results of Attention output matrix.}
      \label{subfig:qwen14b_attn_rome}
  \end{subfigure}
  \caption{The latent probabilities of $o$ for the input and output of MLP and Attention output matrix in Qwen2.5-14B-Instruct edited by ROME.}
  % \caption {Effects of MLP and Attention in Qwen2.5-14B-Instruct edited by ROME.}
  \label{fig:qwen14b_modio_rome}
\end{figure*}

\begin{figure*}[t]
  \centering
  \begin{subfigure}[t]{0.48\textwidth}
      \centering
      \includegraphics[width=\linewidth]{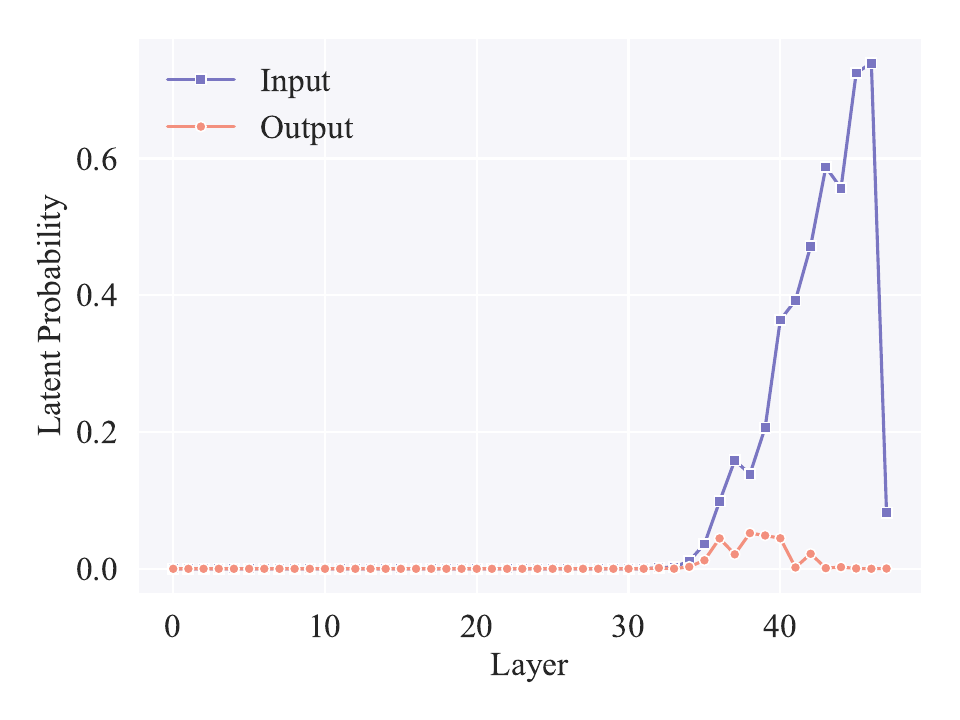}
      \caption{Results of MLP.}
      \label{subfig:qwen14b_mlp_memit}
  \end{subfigure}
  \hfill
  \begin{subfigure}[t]{0.48\textwidth}
      \centering
      \includegraphics[width=\linewidth]{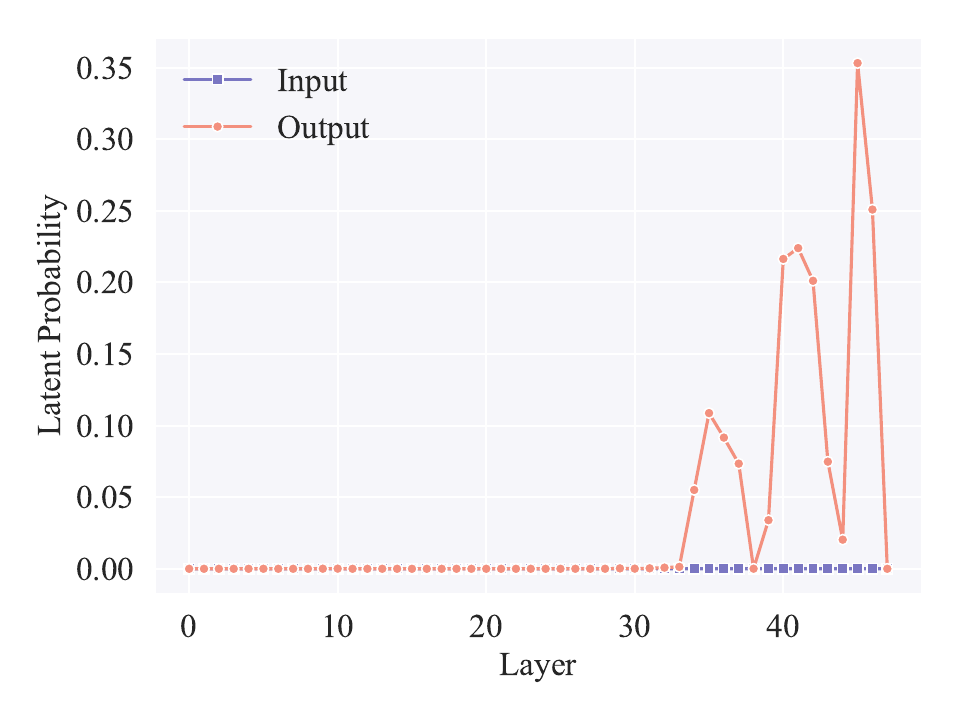}
      \caption{Results of Attention output matrix.}
      \label{subfig:qwen14b_attn_memit}
  \end{subfigure}
  \caption{The latent probabilities of $o$ for the input and output of MLP and Attention output matrix in Qwen2.5-14B-Instruct edited by MEMIT.}
  % \caption {Effects of MLP and Attention in Qwen2.5-14B-Instruct edited by MEMIT.}
  \label{fig:qwen14b_modio_memit}
\end{figure*}

The effects of the MLP and the attention mechanism for other settings are shown in Figures \ref{fig:qwen7b_modio_rome} to \ref{fig:qwen14b_modio_memit}.

\subsection{Investigation and Validation of \textbf{H1}}\label{subsec:appen_h1}
\begin{figure*}[t]
  \centering
  \begin{subfigure}[t]{0.48\textwidth}
      \centering
      \includegraphics[width=\linewidth]{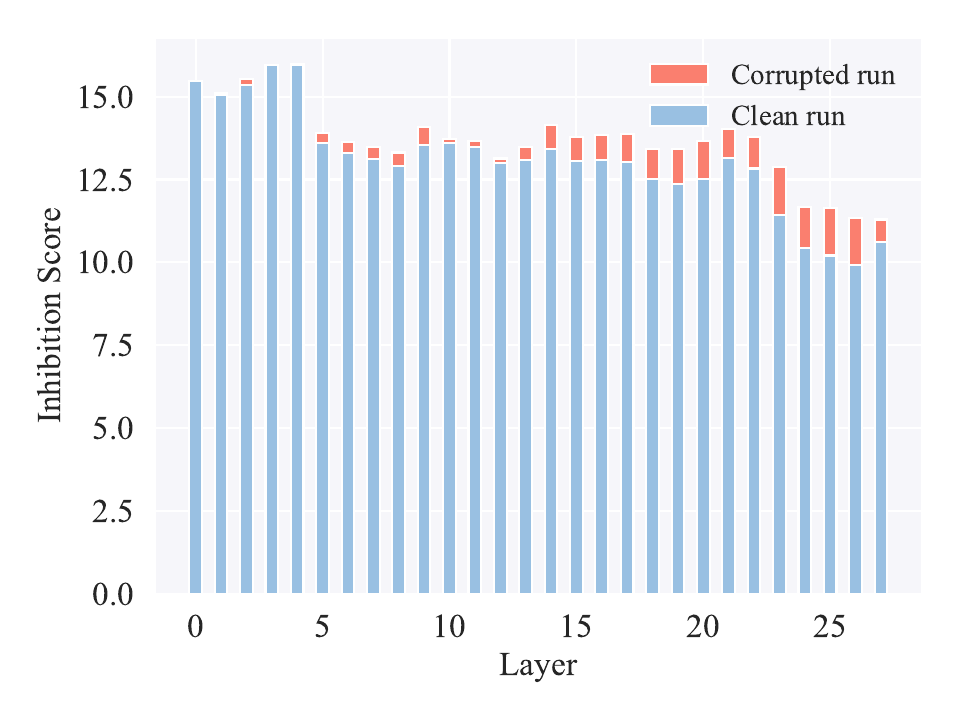}
      \caption{ROME.}
      \label{subfig:suppress_direct_qwen7b_rome}
  \end{subfigure}
  \hfill
  \begin{subfigure}[t]{0.48\textwidth}
      \centering
      \includegraphics[width=\linewidth]{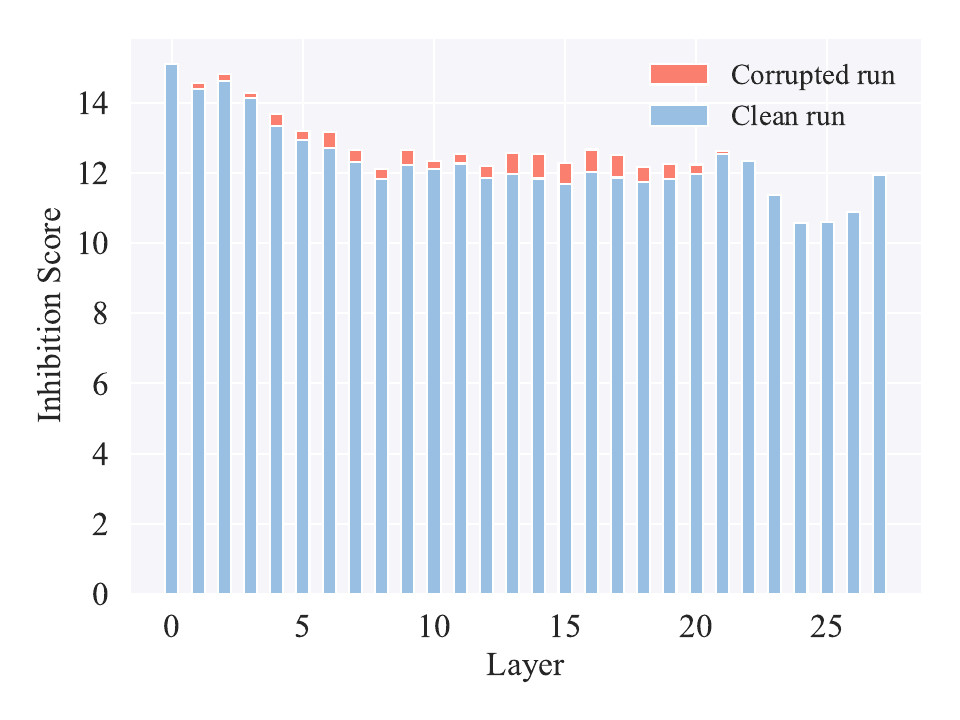}
      \caption{MEMIT.}
      \label{subfig:suppress_direct_qwen7b_memit}
  \end{subfigure}
  \caption {The suppression results for Qwen2.5-7B-Instruct edited by ROME and MEMIT.}
  \label{fig:suppress_direct_qwen7b}
\end{figure*}

\begin{figure*}[t]
  \centering
  \begin{subfigure}[t]{0.48\textwidth}
      \centering
      \includegraphics[width=\linewidth]{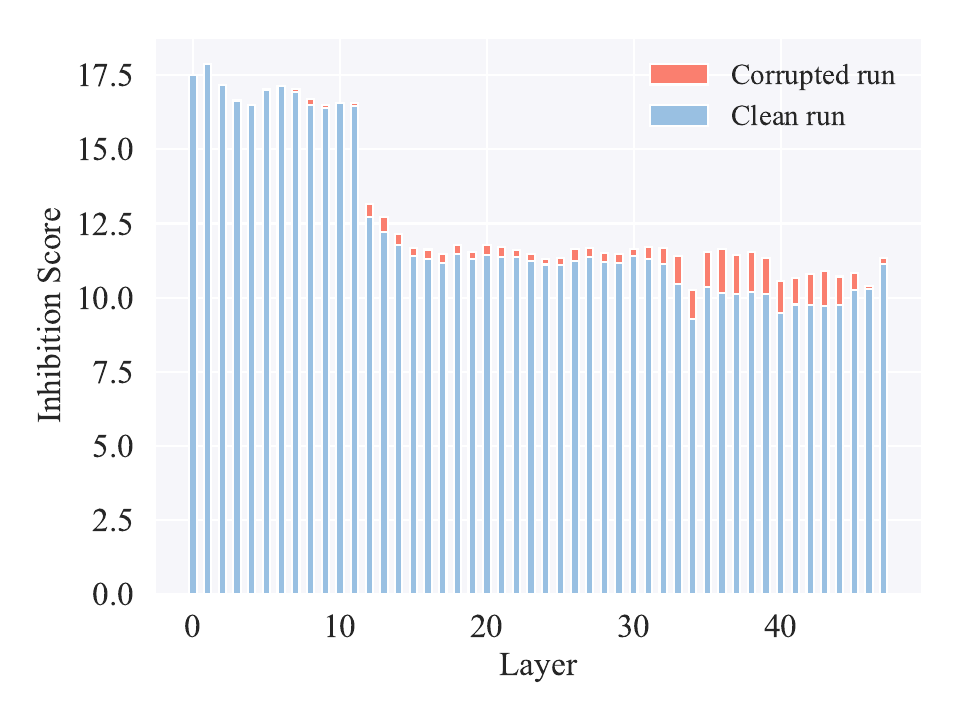}
      \caption{ROME.}
      \label{subfig:suppress_direct_qwen14b_rome}
  \end{subfigure}
  \hfill
  \begin{subfigure}[t]{0.48\textwidth}
      \centering
      \includegraphics[width=\linewidth]{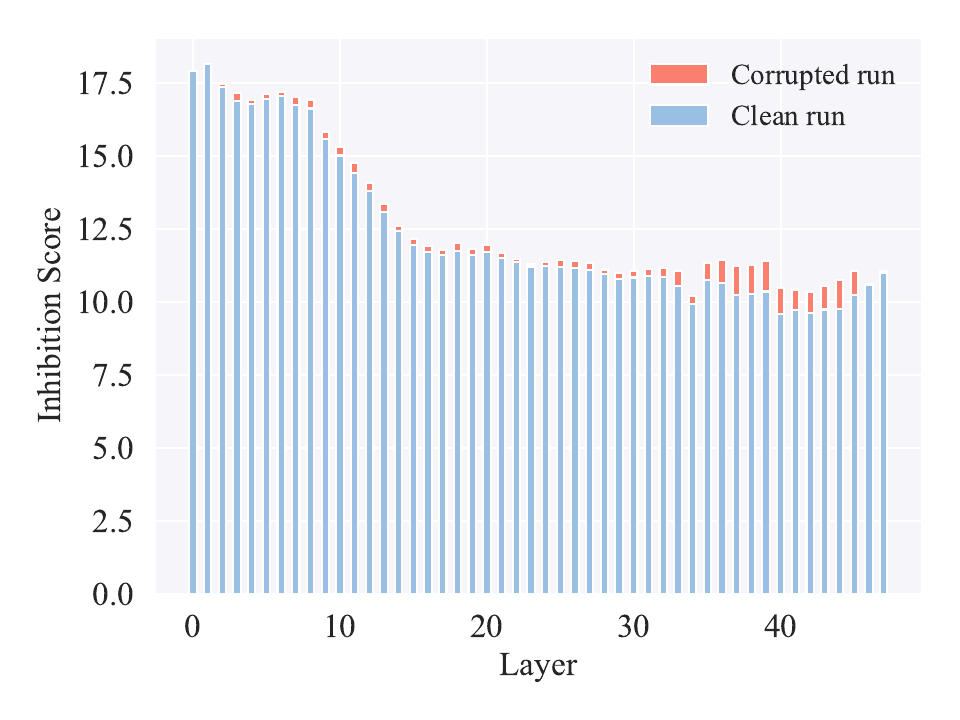}
      \caption{MEMIT.}
      \label{subfig:suppress_direct_qwen14b_memit}
  \end{subfigure}
  \caption {The suppression results for Qwen2.5-14B-Instruct edited by ROME and MEMIT.}
  \label{fig:suppress_direct_qwen14b}
\end{figure*}

The results of the Inhibition Score for other settings are depicted in Figures \ref{fig:suppress_direct_qwen7b} and \ref{fig:suppress_direct_qwen14b}.
% The suppression effect results for other settings are depicted in Figures \ref{fig:suppress_direct_qwen7b}, and \ref{fig:suppress_direct_qwen14b}.

\begin{figure*}[t]
  \centering
  \begin{subfigure}[t]{0.48\textwidth}
      \centering
      \includegraphics[width=\linewidth]{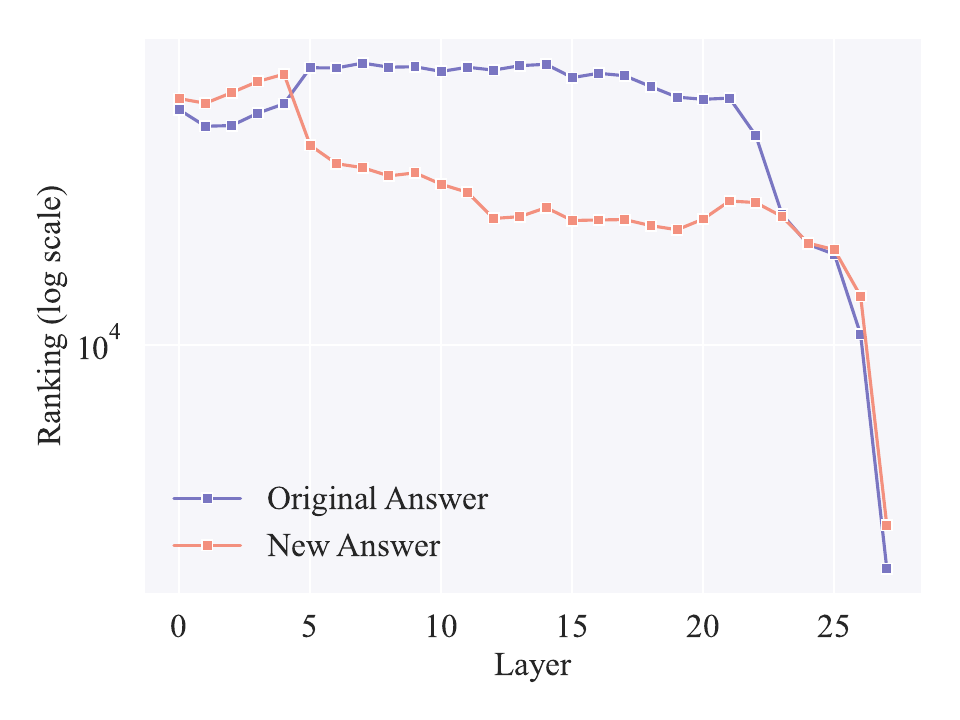}
      \caption{ROME.}
      \label{subfig:qwen7b_rome_rank}
  \end{subfigure}
  \hfill
  \begin{subfigure}[t]{0.48\textwidth}
      \centering
      \includegraphics[width=\linewidth]{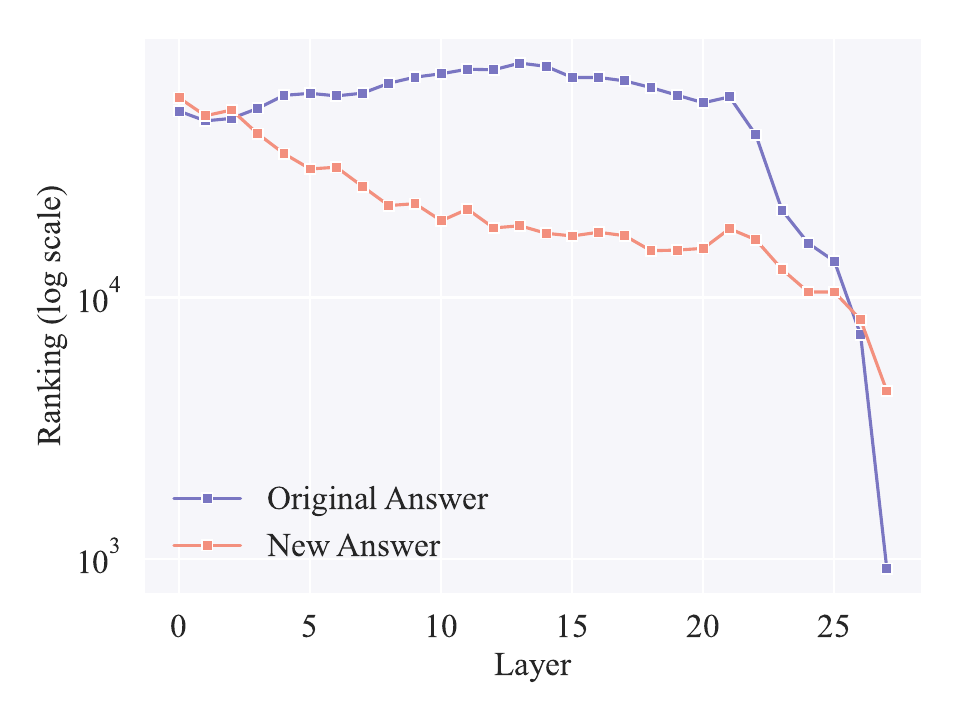}
      \caption{MEMIT.}
      \label{subfig:qwen7b_memit_rank}
  \end{subfigure}
  \caption {The ranking of $o$ and $o^*$ in the latent probability distribution at the last subject position for Qwen2.5-7B-Instruct edited by ROME and MEMIT.}
  \label{fig:noold_qwen7b}
\end{figure*}

\begin{figure*}[t]
  \centering
  \begin{subfigure}[t]{0.48\textwidth}
      \centering
      \includegraphics[width=\linewidth]{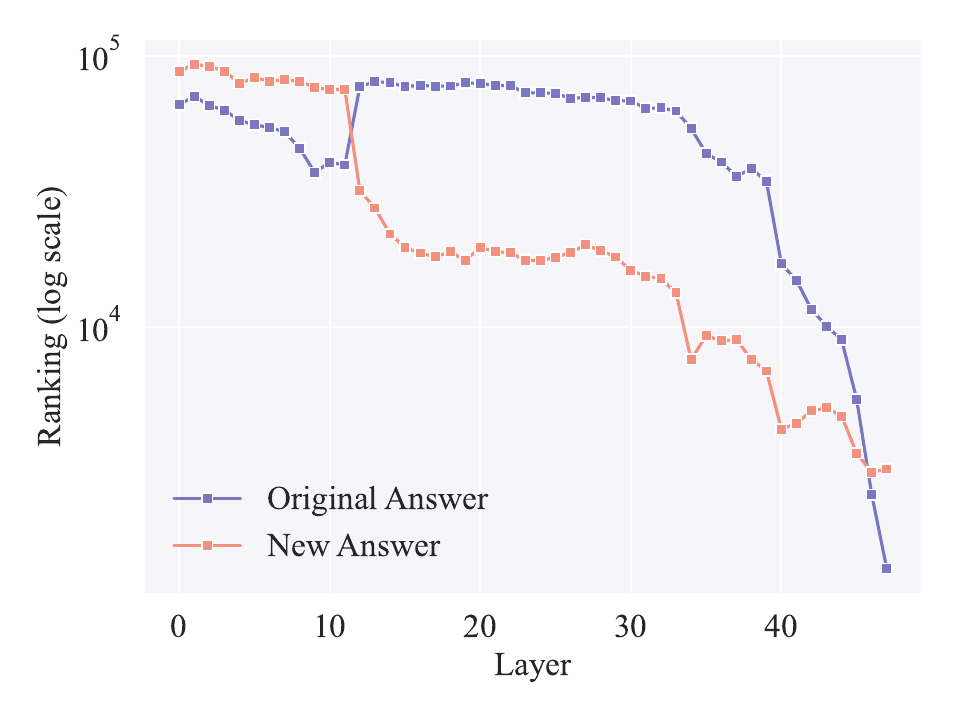}
      \caption{ROME.}
      \label{subfig:qwen14b_rome_rank}
  \end{subfigure}
  \hfill
  \begin{subfigure}[t]{0.48\textwidth}
      \centering
      \includegraphics[width=\linewidth]{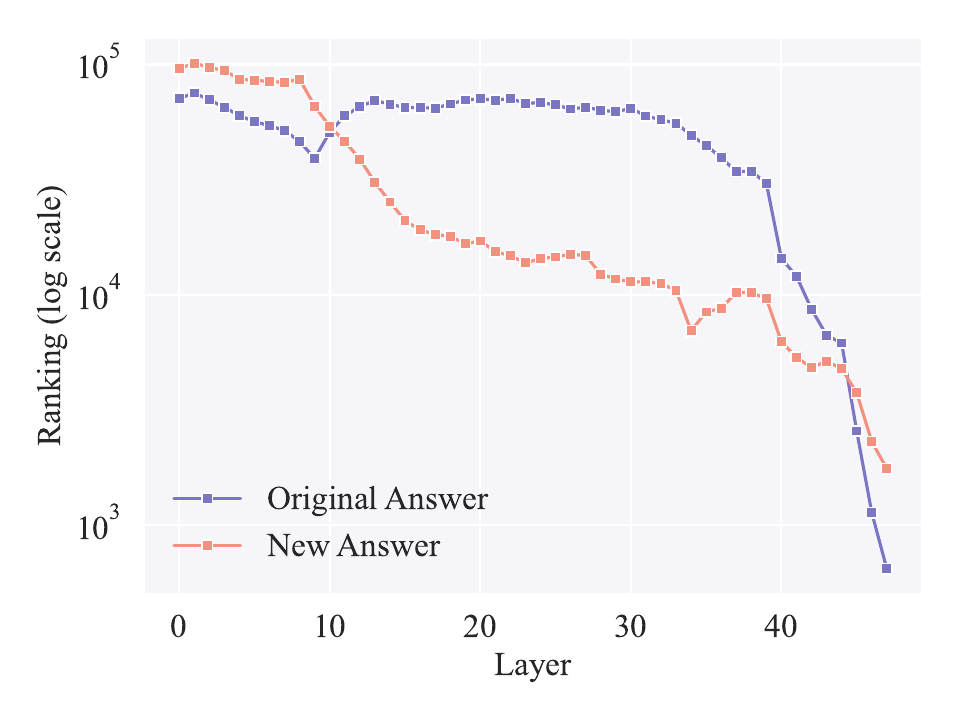}
      \caption{MEMIT.}
      \label{subfig:qwen14b_memit_rank}
  \end{subfigure}
  \caption {The ranking of $o$ and $o^*$ in the latent probability distribution at the last subject position for Qwen2.5-14B-Instruct edited by ROME and MEMIT.}
  \label{fig:noold_qwen14b}
\end{figure*}

The ranking results for Qwen2.5-7B-Instruct and Qwen2.5-14B-Instruct are presented in Figure \ref{fig:noold_qwen7b} and Figure \ref{fig:noold_qwen14b}, respectively.

\subsection{Investigation and Validation of \textbf{H2}}\label{subsec:appen_h2}
\begin{figure*}[t]
  \centering
  \begin{subfigure}[t]{0.48\textwidth}
      \centering
      \includegraphics[width=\linewidth]{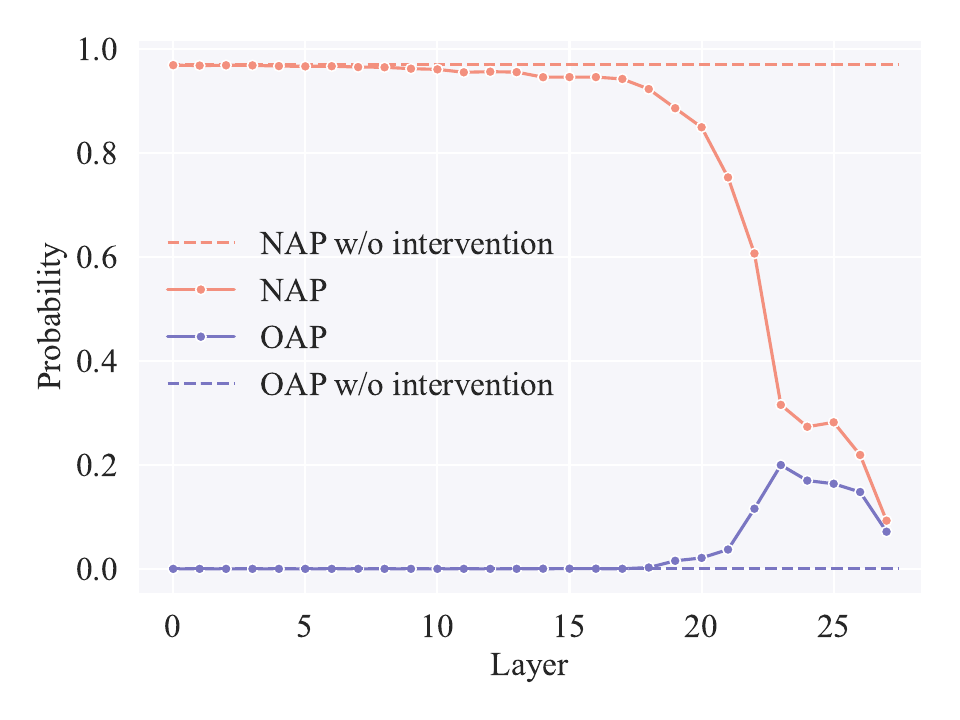}
      \caption{ROME.}
      \label{subfig:qwen7b_rome_patch_wo_attn}
  \end{subfigure}
  \hfill
  \begin{subfigure}[t]{0.48\textwidth}
      \centering
      \includegraphics[width=\linewidth]{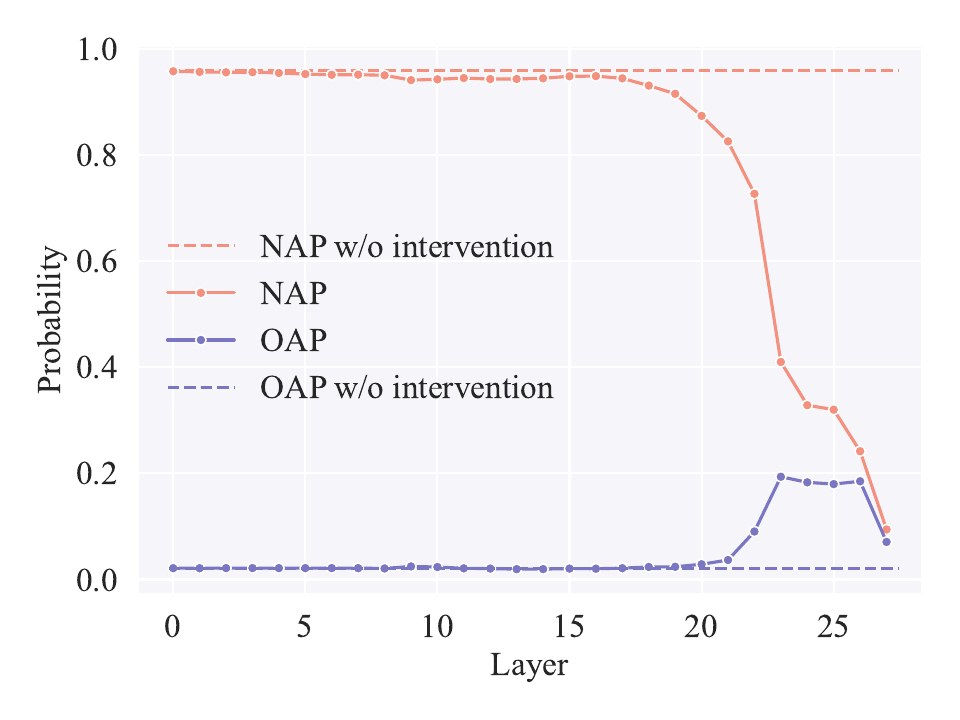}
      \caption{MEMIT.}
      \label{subfig:qwen7b_memit_patch_wo_attn}
  \end{subfigure}
  \caption {Intervention effects following critical attention module ablation in Qwen2.5-7B-Instruct.}
  \label{fig:patch_wo_attn_qwen7b}
\end{figure*}

\begin{figure*}[t]
  \centering
  \begin{subfigure}[t]{0.48\textwidth}
      \centering
      \includegraphics[width=\linewidth]{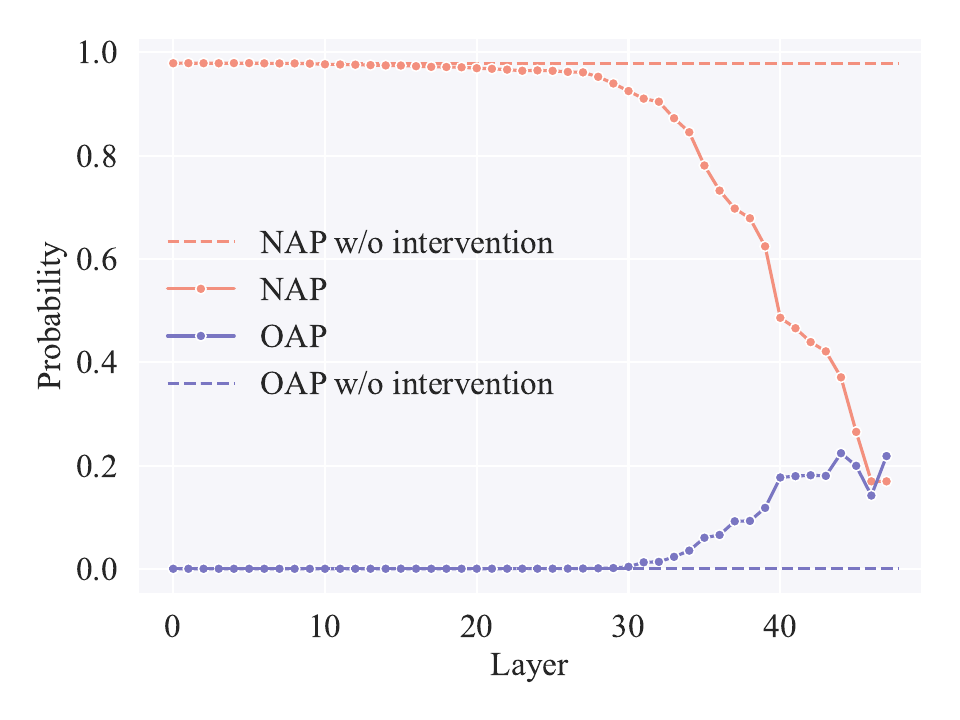}
      \caption{ROME.}
      \label{subfig:qwen14b_rome_patch_wo_attn}
  \end{subfigure}
  \hfill
  \begin{subfigure}[t]{0.48\textwidth}
      \centering
      \includegraphics[width=\linewidth]{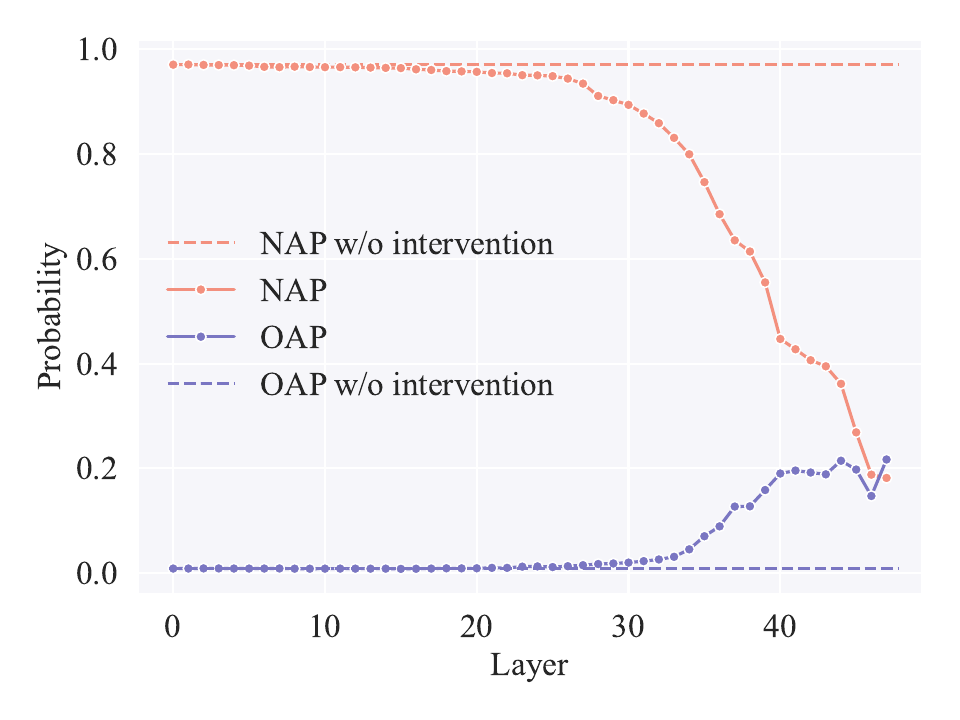}
      \caption{MEMIT.}
      \label{subfig:qwen14b_memit_patch_wo_attn}
  \end{subfigure}
  \caption {Intervention effects following critical attention module ablation in Qwen2.5-14B-Instruct.}
  \label{fig:patch_wo_attn_qwen14b}
\end{figure*}

The intervention results with specific attention modules ablated for other settings are shown in Figures \ref{fig:patch_wo_attn_qwen7b} and \ref{fig:patch_wo_attn_qwen14b}.

\begin{figure*}[t]
  \centering
  \begin{subfigure}[t]{0.48\textwidth}
      \centering
      \includegraphics[width=\linewidth]{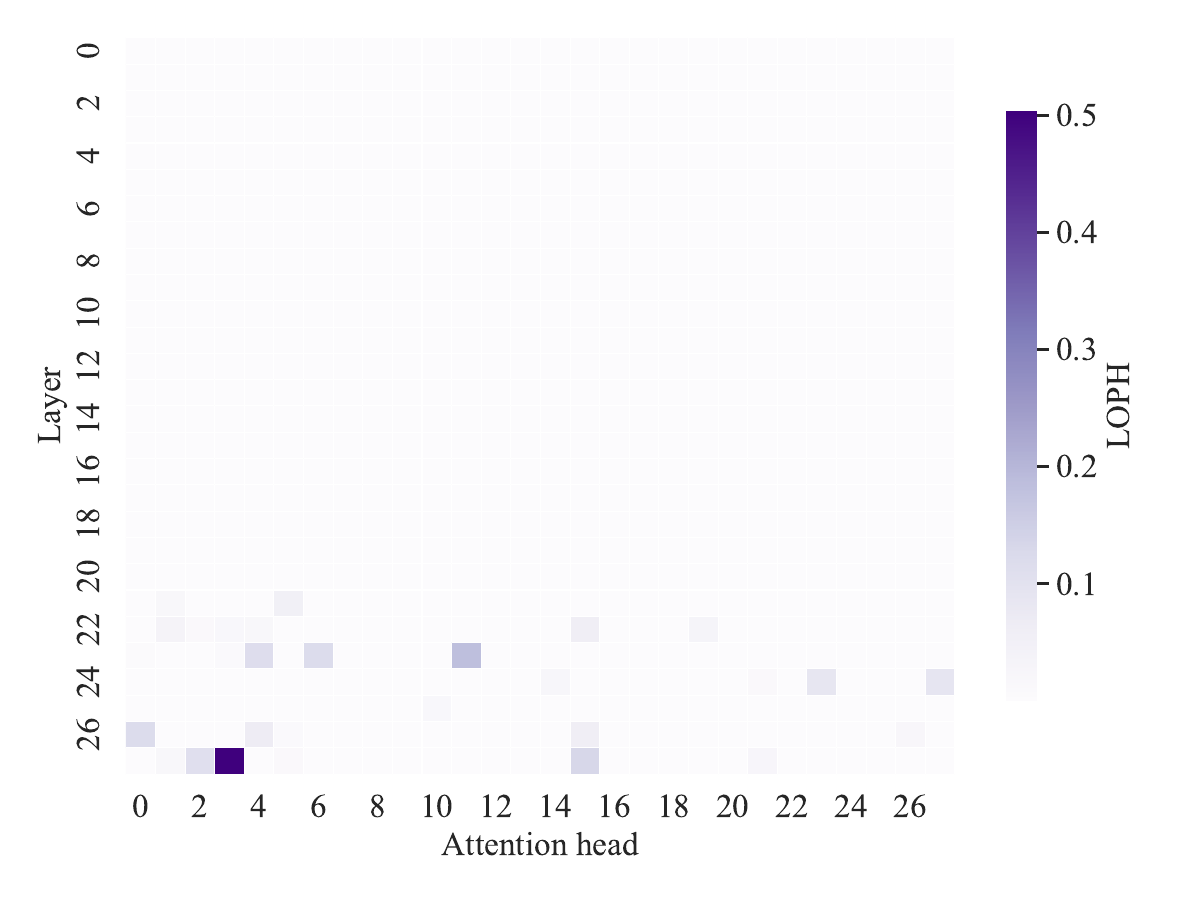}
      \caption{ROME.}
      \label{subfig:qwen7b_rome_heatmap}
  \end{subfigure}
  \hfill
  \begin{subfigure}[t]{0.48\textwidth}
      \centering
      \includegraphics[width=\linewidth]{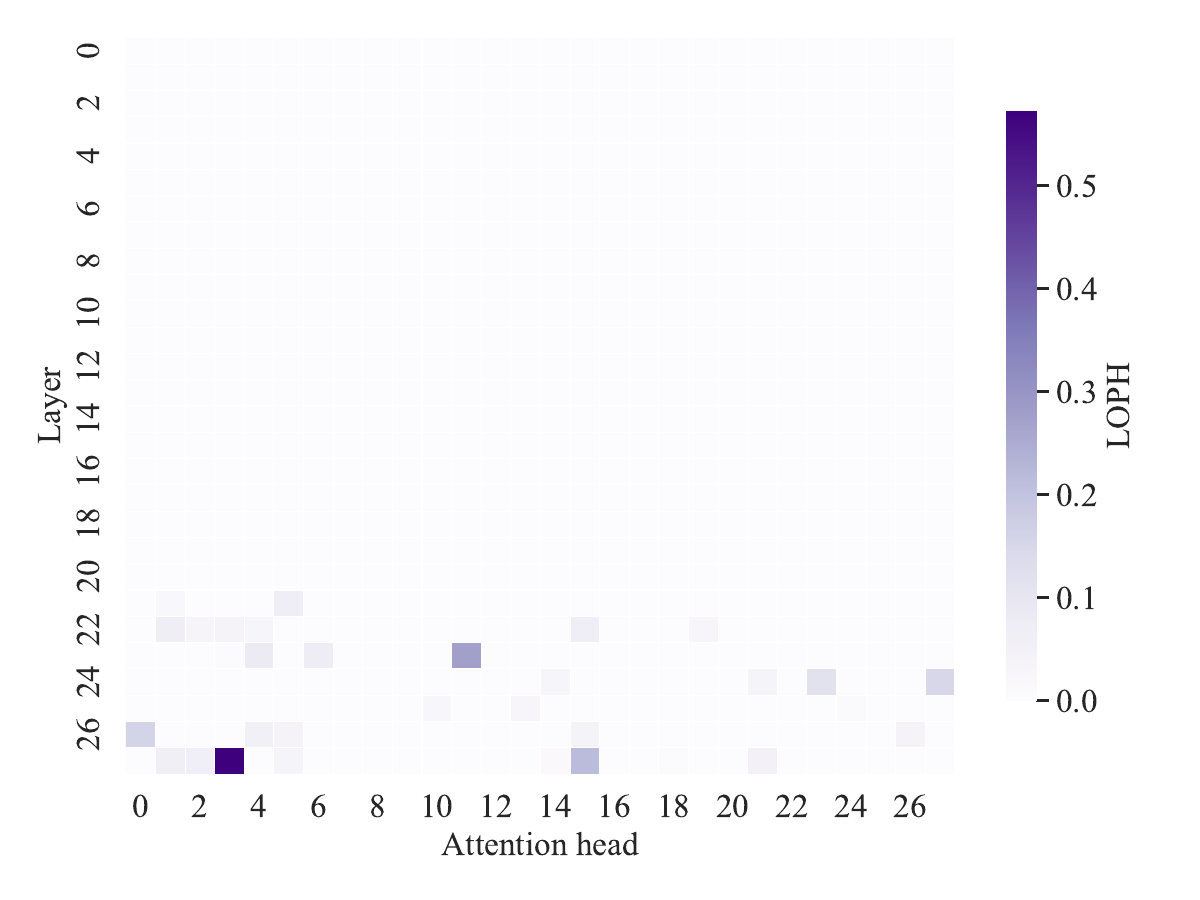}
      \caption{MEMIT.}
      \label{subfig:qwen7b_memit_heatmap}
  \end{subfigure}
  \caption { LOPH of Qwen2.5-7B-Instruct edited by ROME and MEMIT.}
  \label{fig:qwen7b_heatmap}
\end{figure*}

\begin{figure*}[t]
  \centering
  \begin{subfigure}[t]{0.48\textwidth}
      \centering
      \includegraphics[width=\linewidth]{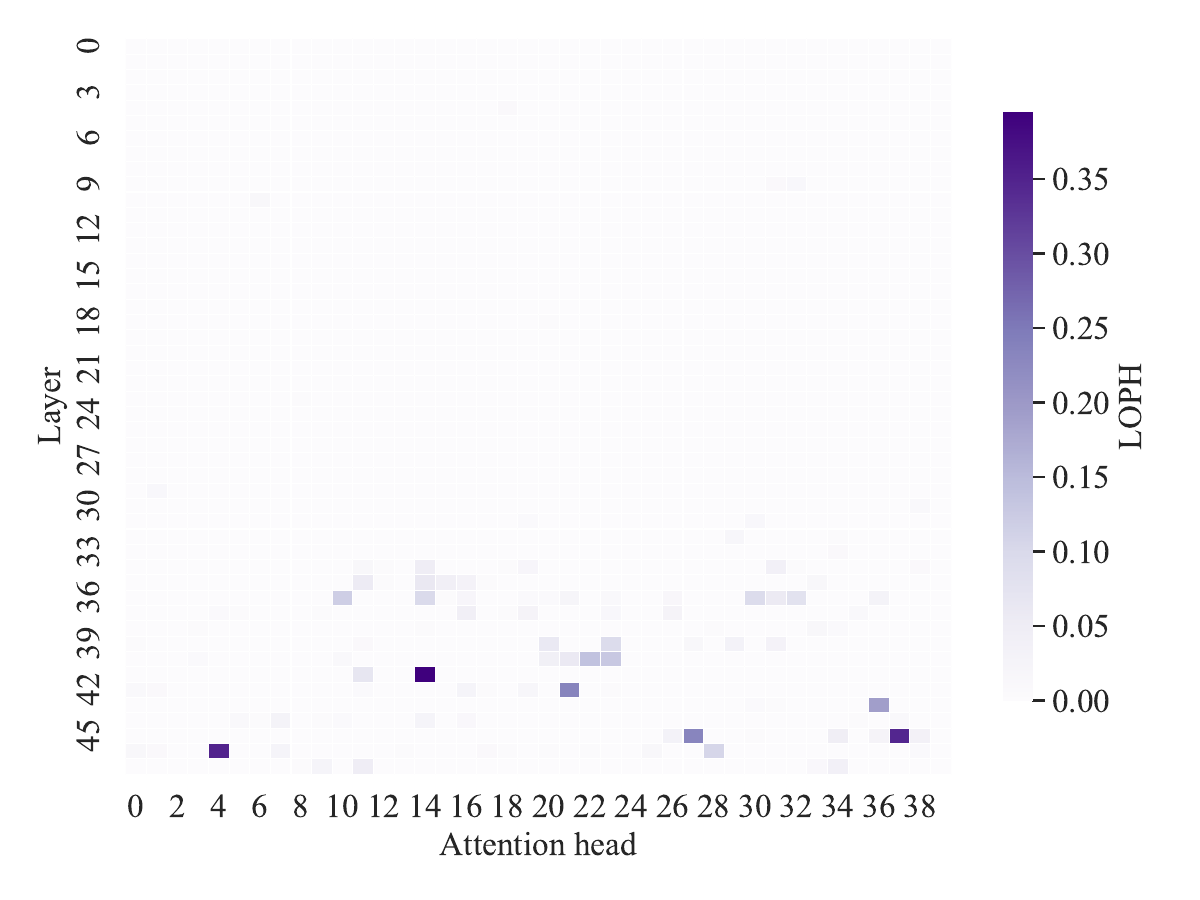}
      \caption{ROME.}
      \label{subfig:qwen14b_rome_heatmap}
  \end{subfigure}
  \hfill
  \begin{subfigure}[t]{0.48\textwidth}
      \centering
      \includegraphics[width=\linewidth]{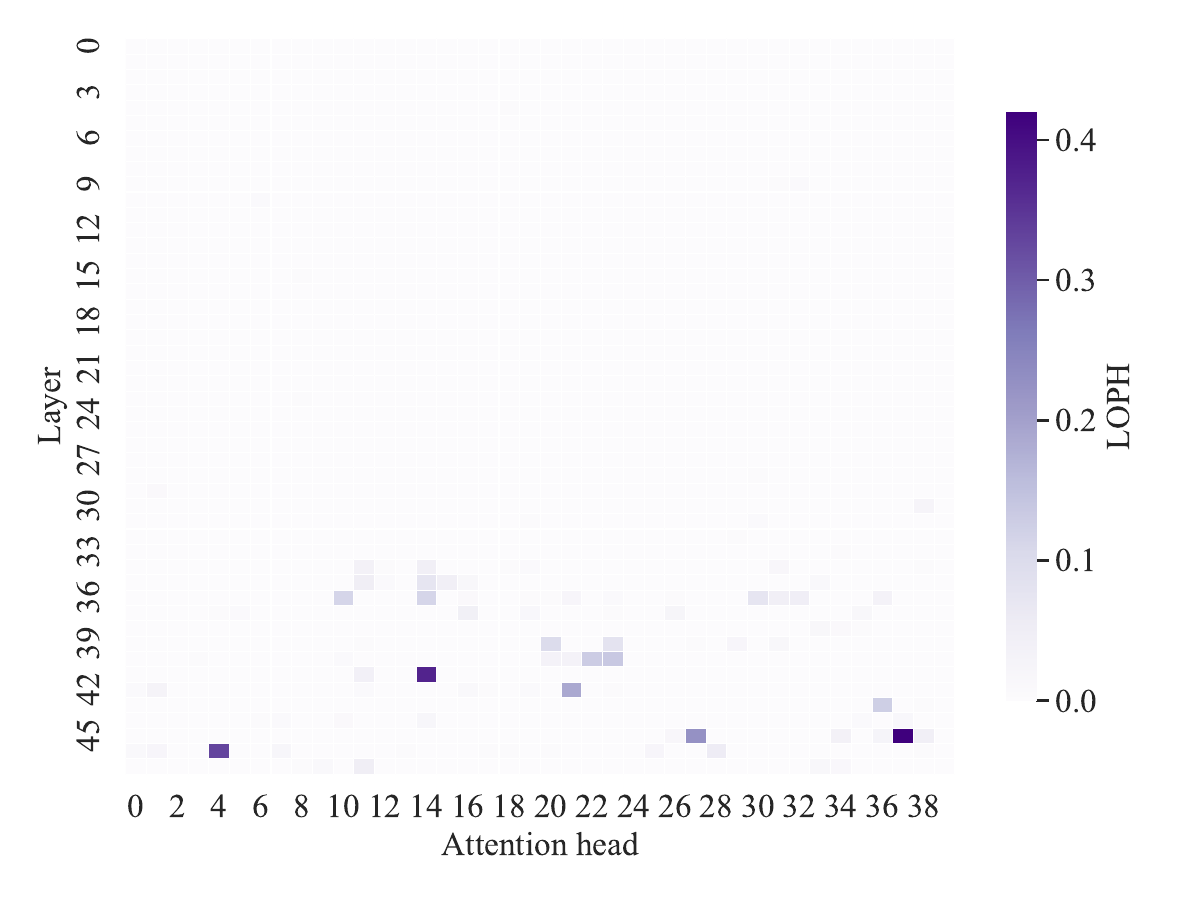}
      \caption{MEMIT.}
      \label{subfig:qwen14b_memit_heatmap}
  \end{subfigure}
  \caption { LOPH of Qwen2.5-14B-Instruct edited by ROME and MEMIT.}
  \label{fig:qwen14b_heatmap}
\end{figure*}

The LOPH results for other settings are illustrated in Figures \ref{fig:qwen7b_heatmap} and \ref{fig:qwen14b_heatmap}.

The DSR results for other settings are provided in Tables \ref{tab:llama3_sdr_memit} to \ref{tab:qwen14b_sdr_memit}.

% \begin{figure}[t]
%   \includegraphics[width=\columnwidth]{figures/noold/llama3_memit_rank.pdf}
%   \caption{The ranking results for LLaMA3-8B-Instruct edited by MEMIT.}
%   \label{fig:noold_llama3_memit}
% \end{figure}

% \begin{figure}[t]
%   \includegraphics[width=\columnwidth]{figures/patch_wo_attn/llama3_memit.pdf}
%   \caption{Probabilities of $o$ and $o^*$ of the clean run, which is intervened by the attention-ablated corrupted run. The model LLaMA3-8B-Instruct is edited by MEMIT. }
%   \label{fig:patch_wo_attn_llama3_memit}
% \end{figure}

% \begin{figure}[t]
%   \includegraphics[width=\columnwidth]{figures/heatmap/llama3_memit.pdf}
%   \caption{LOPH for LLaMA3-8B-Instruct edited by MEMIT.}
%   \label{fig:llama3_heatmap_memit}
% \end{figure}

\subsection{Superficial Unlearning}\label{appen:su}
We first collect data based on the RWKU dataset (\citealp{rwku}). Specifically, we select the first 50 targets from the dataset and train the LLaMA3.2-3B-Instruct\footnote{\url{https://huggingface.co/meta-llama/Llama-3.2-3B-Instruct}} model using gradient ascent (\citealp{su_gaalg}) for each target. Next, we test each unlearned model with probes corresponding to the respective target, selecting samples that elicit a rejection response from the unlearned model (e.g., ``I couldn't...'' or ``I do not have information...''). For each filtered query, we apply GCG (\citealp{gcg}; \citealp{su_yuan}) to train an attack suffix that enables the unlearned model to answer the original knowledge. Finally, we perform a secondary filtering to ensure that all final samples meet the following criteria: they prompt the unlearned model to produce a rejection response in the absence of the attack suffix, while simultaneously allowing the unlearned model to generate the original knowledge when presented with the attack suffix. Through the above process, we ultimately obtain 26 targets with 50 samples.

To explore the mechanisms underlying superficial unlearning, we project the output of each attention head into the vocabulary space and observe the latent probability of $o$ using the method outlined in Section \ref{subsubsection:attn2}. For original answers comprising multiple tokens, we focus exclusively on the probability of the first token. The results, presented in Figure \ref{fig:su_heatmap_document}, reveal that under the unlearning setting, specific attention heads remain active, with the majority concentrated in the later layers. This observation aligns with the conclusion drawn in Section \ref{subsubsection:attn2}.

% \begin{figure}[t]
%   \includegraphics[width=\columnwidth]{figures/unlearning/heatmap.pdf}
%   \caption{Average LOPH of the unlearned models.}
%   \label{fig:su_heatmap}
% \end{figure}

We select the heads with LOPH greater than 0.02 and perform SVD on them. Following this, we ablate the identified left singular vectors using the method described in Section \ref{subsubsection:attn_svd} and observe the resulting variations in the model's output probability of $o$. The results in Table \ref{tab:unlearning_abl_svd_document} show that, following the identification and ablation of the top 5\% and top 10\% left singular vectors, the probability of the unlearned model generating the original answer decreases when confronted with adversarial inputs. 
This suggests that, similar to superficial editing, the occurrence of superficial unlearning is causally linked to these vectors, further demonstrating the generalizability of our analysis method and conclusions.

% \begin{figure}[t]
%   \includegraphics[width=\columnwidth]{figures/unlearning/abl_svd2.pdf}
%   \caption{Output probabilities of $o$ under different ablation settings of left singular vectors.}
%   \label{fig:su_ablsvd}
% \end{figure}

\section{Logit Lens}\label{sec:appen_related_work}
The logit lens (\citealp{logitlens0}; \citealp{geva-logitlens}; \citealp{dar-logitlens}; \citealp{halawi-logitlens}) technique has emerged as a powerful tool for understanding the internal mechanisms of language models. It leverages the observation that the hidden states at each layer of a Transformer, when appropriately decoded, gradually converge towards the final output distribution. The core idea is to project an internal representation into the vocabulary space:
\begin{equation}
    P_{LL}\left(t\mid \boldsymbol{x} \right) = \text{softmax}\left( \boldsymbol{W}_U \boldsymbol{x} \right),
\end{equation}
where $t$ is the next token, $\boldsymbol{x}$ is an internal representation, $\boldsymbol{W}_U$ is the unembedding matrix, $P_{LL}\left(t\mid\boldsymbol{x}\right)$ denotes the probability of obtaining $t$ after decoding $\boldsymbol{x}$. In this study, we refer to $P_{LL}$ as \textbf{latent probability}.

\newpage

\begin{table*}
  \centering
\scalebox{0.95}{\begin{tabular}{c|c|cc|cc|cc|cc|cc}
\toprule
\multirow{2}{*}{} & \multirow{2}{*}{\textbf{Top-K}} & \multicolumn{2}{c|}{\textbf{L23H27}} & \multicolumn{2}{c|}{\textbf{L24H3}} & \multicolumn{2}{c|}{\textbf{L27H20}} & \multicolumn{2}{c}{\textbf{L31H6}} & \multicolumn{2}{c}{\textbf{L31H7}} \\ 
% \cline{3-6}
% \midrule
&   & 5\% & 10\% & 5\%  & 10\% & 5\%  & 10\% & 5\%  & 10\% & 5\%  & 10\%  \\ 
\midrule
\multirow{3}{*}{Original}  
&5  & 5.45 & 8.18 & 30.00 & 35.45 & 36.36 & 47.27 & 37.27 & 47.27 & 37.27 & 42.73 \\
 & 10 & 8.18 & 10.00 & 33.64 & 42.73 & 39.09 & 47.27 & 41.82 & 50.00 & 39.09 & 44.55   \\
& 15  & 9.09 & 10.91 & 38.18 & 42.73 & 40.00 & 49.09 & 43.64 & 51.82 & 40.91 & 44.55  \\ 
\midrule
\multirow{3}{*}{New}   
& 5   & 0.91 & 0.91 & 0.00 & 0.00 & 1.82 & 0.91 & 2.73 & 3.64 & 3.64 & 2.73  \\
& 10  &  0.91 & 0.91 & 0.00 & 0.00 & 3.64 & 4.55 & 3.64 & 4.55 & 3.64 & 4.55  \\
& 15   & 0.91 & 0.91 & 0.00 & 0.00 & 5.45 & 6.36 & 3.64 & 7.27 & 3.64 & 6.36   \\ 
\bottomrule
\end{tabular}}
  \caption{Decoding Success Rate (DSR) of different heads in LLaMA3-8B-Instruct edited by MEMIT.}
  \label{tab:llama3_sdr_memit}
\end{table*}

\begin{table*}
  \centering
\scalebox{0.75}{\begin{tabular}{c|c|cc|cc|cc|cc|cc|cc|cc}
\toprule
\multirow{2}{*}{} & \multirow{2}{*}{\textbf{Top-K}} & \multicolumn{2}{c|}{\textbf{L23H4}} & \multicolumn{2}{c|}{\textbf{L23H6}} & \multicolumn{2}{c|}{\textbf{L23H11}} & \multicolumn{2}{c|}{\textbf{L26H0}} & \multicolumn{2}{c|}{\textbf{L27H2}} & \multicolumn{2}{c|}{\textbf{L27H3}} & \multicolumn{2}{c}{\textbf{L27H15}} \\ 
% \cline{3-6}
% \midrule
&   & 5\% & 10\% & 5\%  & 10\% & 5\%  & 10\% & 5\%  & 10\% & 5\%  & 10\% & 5\%  & 10\% & 5\%  & 10\%  \\ 
\midrule
\multirow{3}{*}{Original}  
&5  &  25.17 & 34.69 & 34.01 & 49.66 & 28.57 & 59.86 & 8.16 & 17.01 & 21.77 & 43.54 & 33.33 & 41.50 & 3.40 & 17.01  \\
 & 10 & 31.29 & 42.18 & 41.50 & 54.42 & 37.41 & 61.90 & 10.88 & 25.17 & 27.89 & 46.94 & 38.10 & 44.90 & 7.48 & 19.73  \\
& 15  & 36.05 & 43.54 & 43.54 & 55.78 & 40.82 & 62.59 & 12.93 & 27.21 & 34.69 & 47.62 & 38.78 & 46.94 & 10.20 & 21.77  \\ 
\midrule
\multirow{3}{*}{New}   
& 5   &  0.68   &  1.36 &  1.36  &  1.36 & 0.00 & 0.00 & 0.00 & 0.00 & 0.00 & 0.00 & 1.36 & 0.00 & 0.00 & 0.00 \\
& 10  & 0.68 & 2.04 & 2.72 & 2.72 & 0.00 & 0.00 & 0.00 & 0.00 & 0.00 & 0.00 & 3.40 & 2.04 & 0.00 & 0.00  \\
& 15   & 0.68 & 3.40 & 4.76 & 4.76 & 0.00 & 0.00 & 0.00 & 0.00 & 0.00 & 0.00 & 4.08 & 2.04 & 0.00 & 0.68  \\ 
\bottomrule
\end{tabular}}
  \caption{Decoding Success Rate (DSR) of different heads in Qwen2.5-7B-Instruct edited by ROME.}
  \label{tab:qwen7b_sdr_rome}
\end{table*}

\begin{table*}
  \centering
\scalebox{0.85}{\begin{tabular}{c|c|cc|cc|cc|cc|cc|cc}
\toprule
\multirow{2}{*}{} & \multirow{2}{*}{\textbf{Top-K}} & \multicolumn{2}{c|}{\textbf{L23H11}} & \multicolumn{2}{c|}{\textbf{L24H23}} & \multicolumn{2}{c|}{\textbf{L24H27}} & \multicolumn{2}{c|}{\textbf{L26H0}} & \multicolumn{2}{c|}{\textbf{L27H3}} & \multicolumn{2}{c}{\textbf{L27H15}} \\ 
% \cline{3-6}
% \midrule
&   & 5\% & 10\% & 5\%  & 10\% & 5\%  & 10\% & 5\%  & 10\% & 5\%  & 10\% & 5\%  & 10\% \\ 
\midrule
\multirow{3}{*}{Original}  
&5  & 33.93 & 63.39 & 30.36 & 56.25 & 22.32 & 49.11 & 9.82 & 22.32 & 37.50 & 48.21 & 7.14 & 28.57 \\
 & 10 & 42.86 & 67.86 & 40.18 & 63.39 & 30.36 & 58.93 & 14.29 & 25.00 & 41.96 & 50.00 & 16.96 & 31.25  \\
& 15  & 48.21 & 68.75 & 41.96 & 63.39 & 33.93 & 59.82 & 16.07 & 27.68 & 49.11 & 51.79 & 23.21 & 33.04  \\ 
\midrule
\multirow{3}{*}{New}   
& 5   & 0.00 & 0.00 & 0.00 & 0.00 & 0.00 & 0.00 & 0.00 & 0.00 & 1.79 & 1.79 & 0.00 & 0.00  \\
& 10  & 0.00 & 0.00 & 0.00 & 0.00 & 0.00 & 0.00 & 0.00 & 0.00 & 3.57 & 4.46 & 0.89 & 0.89 \\
& 15  & 0.00 & 0.00 & 0.00 & 0.00 & 0.00 & 0.00 & 0.89 & 0.00 & 5.36 & 6.25 & 0.89 & 0.89 \\ 
\bottomrule
\end{tabular}}
  \caption{Decoding Success Rate (DSR) of different heads in Qwen2.5-7B-Instruct edited by MEMIT.}
  \label{tab:qwen7b_sdr_memit}
\end{table*}

\begin{table*}
  \centering
\scalebox{0.95}{\begin{tabular}{c|c|cc|cc|cc|cc|cc}
\toprule
\multirow{2}{*}{} & \multirow{2}{*}{\textbf{Top-K}} & \multicolumn{2}{c|}{\textbf{L36H10}} & \multicolumn{2}{c|}{\textbf{L40H22}} & \multicolumn{2}{c|}{\textbf{L40H23}} & \multicolumn{2}{c}{\textbf{L41H14}} & \multicolumn{2}{c}{\textbf{L42H21}} \\ 
% \cline{3-6}
% \midrule
&   & 5\% & 10\% & 5\%  & 10\% & 5\%  & 10\% & 5\%  & 10\% & 5\%  & 10\%  \\ 
\midrule
\multirow{3}{*}{Original}  
&5  & 29.06 & 52.22 & 29.56 & 40.39 & 29.06 & 46.80 & 26.11 & 37.44 & 10.34 & 22.17 \\
 & 10 & 35.96 & 60.10 & 36.45 & 45.32 & 34.97 & 50.74 & 33.00 & 45.81 & 15.27 & 28.57  \\
& 15  & 40.89 & 64.04 & 39.41 & 45.81 & 39.41 & 56.16 & 36.95 & 51.23 & 18.72 & 32.02  \\ 
\midrule
\multirow{3}{*}{New}   
& 5 & 0.00 & 0.00 & 0.99 & 1.48 & 0.49 & 0.49 & 1.48 & 2.46 & 0.00 & 0.00 \\
& 10 & 0.00 & 0.00 & 2.46 & 2.96 & 0.49 & 0.49 & 2.46 & 4.43 & 0.00 & 0.99  \\
& 15 & 0.00 & 0.00 & 3.94 & 3.94 & 0.49 & 0.99 & 4.43 & 8.87 & 0.00 & 0.99 \\ 
\bottomrule
\end{tabular}}
\scalebox{0.95}{\begin{tabular}{c|c|cc|cc|cc|cc|cc}
\toprule
\multirow{2}{*}{} & \multirow{2}{*}{\textbf{Top-K}} & \multicolumn{2}{c|}{\textbf{L43H36}} & \multicolumn{2}{c|}{\textbf{L45H27}} & \multicolumn{2}{c|}{\textbf{L45H37}} & \multicolumn{2}{c}{\textbf{L46H4}} & \multicolumn{2}{c}{\textbf{L46H28}} \\ 
% \cline{3-6}
% \midrule
&   & 5\% & 10\% & 5\%  & 10\% & 5\%  & 10\% & 5\%  & 10\% & 5\%  & 10\%  \\ 
\midrule
\multirow{3}{*}{Original}  
&5  & 15.27 & 21.67 & 11.33 & 26.11 & 19.70 & 33.99 & 20.20 & 36.95 & 13.30 & 22.66 \\
 & 10 & 17.24 & 25.12 & 17.24 & 29.56 & 24.63 & 37.44 & 26.11 & 40.89 & 14.78 & 29.56  \\
& 15 & 20.20 & 28.08 & 18.23 & 32.02 & 27.59 & 38.42 & 29.56 & 43.84 & 15.76 & 34.48  \\ 
\midrule
\multirow{3}{*}{New}   
& 5   & 1.48 & 1.48 & 0.00 & 0.00 & 0.00 & 0.49 & 0.00 & 0.49 & 0.00 & 0.00  \\
& 10  & 1.97 & 1.48 & 0.00 & 0.49 & 0.49 & 0.49 & 0.49 & 0.49 & 0.00 & 0.49   \\
& 15  & 2.46 & 1.97 & 0.00 & 0.49 & 0.49 & 0.49 & 0.99 & 0.49 & 0.00 & 0.99 \\ 
\bottomrule
\end{tabular}}
  \caption{Decoding Success Rate (DSR) of different heads in Qwen2.5-14B-Instruct edited by ROME.}
  \label{tab:qwen14b_sdr_rome}
\end{table*}

\begin{table*}
  \centering
\scalebox{0.95}{\begin{tabular}{c|c|cc|cc|cc|cc|cc}
\toprule
\multirow{2}{*}{} & \multirow{2}{*}{\textbf{Top-K}} & \multicolumn{2}{c|}{\textbf{L36H14}} & \multicolumn{2}{c|}{\textbf{L39H20}} & \multicolumn{2}{c|}{\textbf{L40H22}} & \multicolumn{2}{c|}{\textbf{L40H23}} & \multicolumn{2}{c}{\textbf{L41H14}} \\ 
% \cline{3-6}
% \midrule
&   & 5\% & 10\% & 5\%  & 10\% & 5\%  & 10\% & 5\%  & 10\% & 5\%  & 10\%  \\ 
\midrule
\multirow{3}{*}{Original}  
& 5 & 30.61 & 55.78 & 32.31 & 55.78 & 29.59 & 37.07 & 28.91 & 41.50 & 26.53 & 36.05 \\
 & 10 & 37.07 & 58.84 & 42.52 & 58.16 & 34.35 & 41.50 & 34.69 & 46.60 & 34.69 & 42.86 \\
& 15 & 41.16 & 61.22 & 45.58 & 58.84 & 36.73 & 43.20 & 36.05 & 48.64 & 38.10 & 46.94  \\ 
\midrule
\multirow{3}{*}{New}   
& 5 & 0.00 & 0.00 & 0.34 & 1.02 & 0.34 & 0.68 & 0.34 & 1.02 & 1.02 & 1.36  \\
& 10 & 0.00 & 0.00 & 1.70 & 1.36 & 1.02 & 1.70 & 0.34 & 1.36 & 2.72 & 3.74  \\
& 15 & 0.00 & 0.00 & 1.70 & 2.38 & 1.70 & 2.38 & 0.68 & 1.36 & 4.76 & 6.46 \\ 
\bottomrule
\end{tabular}}
\scalebox{0.95}{\begin{tabular}{c|c|cc|cc|cc|cc|cc}
\toprule
\multirow{2}{*}{} & \multirow{2}{*}{\textbf{Top-K}} & \multicolumn{2}{c|}{\textbf{L42H21}} & \multicolumn{2}{c|}{\textbf{L43H36}} & \multicolumn{2}{c|}{\textbf{L45H27}} & \multicolumn{2}{c|}{\textbf{L45H37}} & \multicolumn{2}{c}{\textbf{L46H4}} \\ 
&   & 5\% & 10\% & 5\%  & 10\% & 5\%  & 10\% & 5\%  & 10\% & 5\%  & 10\%  \\ 
\midrule
\multirow{3}{*}{Original}  
& 5 & 8.84 & 19.73 & 10.20 & 15.99 & 7.82 & 19.05 & 22.11 & 39.80 & 18.71 & 37.41  \\
 & 10 & 11.22 & 26.19 & 11.90 & 18.03 & 10.88 & 22.79 & 26.87 & 43.20 & 24.49 & 40.48  \\
& 15 & 12.93 & 29.93 & 12.93 & 19.39 & 13.61 & 26.87 & 30.27 & 44.56 & 27.89 & 42.52 \\ 
\midrule
\multirow{3}{*}{New}   
& 5   & 0.00 & 0.00 & 0.68 & 1.70 & 0.34 & 0.34 & 0.00 & 0.00 & 0.68 & 0.68 \\
& 10  & 0.00 & 0.00 & 1.36 & 2.04 & 0.34 & 0.34 & 0.00 & 0.00 & 1.02 & 1.36  \\
& 15  & 0.00 & 0.68 & 1.70 & 2.72 & 0.34 & 0.34 & 0.00 & 0.34 & 1.02 & 2.04 \\ 
\bottomrule
\end{tabular}}
  \caption{Decoding Success Rate (DSR) of different heads in Qwen2.5-14B-Instruct edited by MEMIT.}
  \label{tab:qwen14b_sdr_memit}
\end{table*}

% \begin{table*}
% \centering
% \scalebox{0.75}{\begin{tabular}{c|c|cccccc|cccccc}
% \toprule
% \multirow{3}{*}{\textbf{Models}} & \multirow{3}{*}{\textbf{Methods}} & \multicolumn{6}{c|}{15\%} & \multicolumn{6}{c}{20\%}\\ 

% & & \multicolumn{3}{c}{\textbf{OAP}} & \multicolumn{3}{c|}{\textbf{NAP}} & \multicolumn{3}{c}{\textbf{OAP}} & \multicolumn{3}{c}{\textbf{NAP}} \\ 
%  & & w/o abl. & abl. & $\downarrow\Delta P$ & w/o abl.  & abl. & $\uparrow\Delta P$ & w/o abl. & abl. & $\downarrow\Delta P$ & w/o abl.  & abl. & $\uparrow\Delta P$  \\ 
% \midrule
% \multirow{2}{*}{\shortstack{LLaMA3-8B-\\ Instruct}} & ROME & 61.41 & 45.45 & 15.96 & 16.12 & 24.21 & 8.09 & 61.41 & 43.41 & 18.00 & 16.12 & 25.11 & 8.99   \\
%  & MEMIT & 57.42 & 41.37 & 16.05 & 17.05 & 24.68 & 7.63 & 57.42 & 39.30 & 18.12 & 17.05 & 25.84 & 8.79 \\ 
% \midrule
% \multirow{2}{*}{\shortstack{Qwen2.5-7B-\\ Instruct}} & ROME & 64.33 & 47.52 & 16.81 & 11.93 & 21.26 & 9.33 & 64.33 & 45.46 & 18.87 & 11.93 & 22.26 & 10.33 \\
% & MEMIT & 66.41 & 54.89 & 11.52 & 15.72 & 23.41 & 7.69 & 66.41 & 52.98 & 13.43 & 15.72 & 24.62 & 8.90 \\
%   \midrule
% \multirow{2}{*}{\shortstack{Qwen2.5-14B-\\ Instruct}} & ROME  &  \\
%  &  MEMIT &  \\
% \bottomrule
% \end{tabular}}
%   \caption{Answer probabilities before and after singular vector ablation. 
%   % ``w/o abl.'' and ``abl.'' represents ``without ablation'' and ``ablation'', respectively. $\downarrow \Delta P$ and $\uparrow \Delta P$ denote the probability decrease and increase, respectively.
%   }
%   \label{tab:abl_svd_expan}
% \end{table*}

\end{document}